\documentclass{svjour3}                     
\smartqed  
\usepackage{graphicx}
\usepackage{hyperref}
\usepackage{url}
\usepackage{smile}
\usepackage{graphicx} 
\usepackage{algorithm}
\usepackage{algorithmic}
\usepackage{epstopdf}
\usepackage[margin=1.0in]{geometry}
\usepackage[export]{adjustbox}
\usepackage{bbm}
\usepackage{wrapfig}
\usepackage{subcaption}
\usepackage{caption}
\usepackage{enumitem}
\usepackage{tabularx}
\usepackage{smile}
\usepackage{tcolorbox}
\usepackage{enumitem}
\usepackage{mathtools, cuted}
\usepackage{caption}
\usepackage{subcaption}

\numberwithin{equation}{section}

\usepackage[nocompress]{cite}

\renewcommand{\frac}{\tfrac}

\hypersetup{
    colorlinks=true,
    linkcolor=blue,
    anchorcolor=blue, 
    citecolor=blue,
}

\allowdisplaybreaks

\begin{document}

\title{Homotopic Policy Mirror Descent
}
\subtitle{Policy Convergence, Implicit Regularization, and Improved Sample Complexity}

\titlerunning{Homotopic Policy Mirror Descent}        

\author{
  Yan Li       
  \and
  Guanghui Lan
    \and 
    Tuo Zhao 
    }

\authorrunning{Yan Li, Guanghui Lan, and Tuo Zhao} 

\institute{Yan Li \at
              H. Milton Stewart School of Industrial and Systems Engineering, Georgia Institute of Technology, Atlanta, GA, 30332. \\
              \email{yli939@gatech.edu}           
            \and
            Guanghui Lan \at
            H. Milton Stewart School of Industrial and Systems Engineering, Georgia Institute of Technology, Atlanta, GA, 30332. \\
            \email{george.lan@isye.gatech.edu}   
           \and
           Tuo Zhao \at
            H. Milton Stewart School of Industrial and Systems Engineering, Georgia Institute of Technology, Atlanta, GA, 30332. \\
            \email{tourzhao.gatech.edu}   
}




\maketitle

\begin{abstract}
We propose a new policy gradient method, named homotopic policy mirror descent (HPMD),
for solving discounted, infinite horizon MDPs with finite state and action spaces.
HPMD performs a mirror descent type policy update with an additional diminishing regularization term, and possesses several computational properties that seem to be new in the literature.
We first establish the global linear convergence of HPMD instantiated with Kullback-Leibler divergence, 
for both the optimality gap, and a weighted distance to the set of optimal policies. Then local superlinear convergence is obtained for both quantities without any assumption.
With local acceleration and diminishing regularization, 
we establish the first result among policy gradient methods on certifying and characterizing  the limiting policy, 
by showing, with a non-asymptotic characterization,  that the last-iterate policy converges to the unique optimal policy with the maximal entropy.
We then extend all the aforementioned results to HPMD instantiated with a broad class of decomposable Bregman divergences, demonstrating the generality of the these computational properties. 
As a by product, we discover the finite-time exact convergence for some commonly used Bregman divergences,
implying the continuing convergence of HPMD to the limiting policy even if the current policy is already optimal.
Finally, we develop a stochastic version of HPMD and establish similar convergence properties. 
By exploiting the local acceleration, we show that for small optimality gap, a better than $\tilde{\cO}(\abs{\cS} \abs{\cA} / \epsilon^2)$ sample complexity holds with high probability, when assuming a generative model for policy evaluation.

\end{abstract}

\section{Introduction}\label{sec:introduction}

We consider a discrete time Markov decision process (MDP)  denoted by the quintuple $\cM = (\cS, \cA, \cP, C, \gamma)$, where $\cS$ denotes the finite state space, $\cA$ denotes the finite action space, $\cP: \cS \times \cA \to \Delta_{\cS}$ denotes the transition kernel, with $\Delta_{\cS}$ being the probability simplex over $\cS$, $c: \cS \times \cA \to \RR$ denotes the cost function with $\abs{c(s,a)} \leq C$ for all $(s, a)\in \cS \times \cA$, and $\gamma \in (0,1)$ denotes the discount factor.

A randomized, stationary policy $\pi : \cS \to \Delta_{\cA}$ maps a given state $s\in \cS$ into $\pi(\cdot|s) \in \Delta_{\cA}$, and we denote the set of all such policies by $\Pi$.
At any timestep $t$,  the policy explicitly governs what action to be made given the current state $s_t$, by $a_t \sim \pi(\cdot| s_t)$. 
Then a cost $c(s_t, a_t)$ is incurred, followed by the transition to the next state $s_{t+1} \sim \cP(\cdot|s_t, a_t)$. 
The decision process is then repeated iteratively at future timesteps.

For a given policy $\pi$, we measure its performance by its value function $V^{\pi}: \cS \to \RR$, defined as
\begin{align*}
V^{\pi} (s) = \EE \sbr{\tsum_{t=0}^\infty \gamma^t c(s_t, a_t)  \big| s_0 = s, a_t \sim \pi(\cdot|s_t), s_{t+1} \sim \cP(\cdot|s_t,a_t)  }.
\end{align*}
Accordingly, we also define its state-action value function (Q-function) $Q^\pi: \cS \times \cA \to \RR$, as 
\begin{align*}
Q^{\pi} (s, a) = \EE \sbr{\tsum_{t=0}^\infty \gamma^t  c(s_t, a_t) \big| s_0 = s, a_0 = a, a_t \sim \pi(\cdot|s_t), s_{t+1} \sim \cP(\cdot|s_t,a_t)  }.
\end{align*}

The  planning objective of the MDP is to find an optimal policy $\pi^*$ that minimizes the value for every state $s\in \cS$ simultaneously,  
\begin{align}\label{obj:state_wise}
V^{\pi^*}(s) \leq V^{\pi} (s),  ~~ \forall s \in \cS, ~~ \forall  \pi \in \Pi.
\end{align}
The optimal value functions are thus defined as 
\begin{align*}
V^* (s) = \min_{\pi \in \Pi} V^{\pi}(s), ~ Q^*(s,a) = \min_{\pi \in \Pi} Q^{\pi} (s,a), ~ \forall s \in \cS, a \in \cA. 
\end{align*}
The existence of an optimal policy $\pi^*$ for \eqref{obj:state_wise} is well known in the literature of dynamic programming \cite{puterman2014markov}. Hence we can succinctly reformulate \eqref{obj:state_wise} into a single-objective optimization problem
\begin{align}
\textstyle
\label{eq:mdp_single_obj_raw}
\min_{\pi} \cbr{f_\rho (\pi) \coloneqq \EE_{\rho} \sbr{V^{\pi}(s)}}, ~~ \mathrm{s.t.} ~~ \pi(\cdot|s) \in \Delta_{\cA}, \forall s \in \cS,
\end{align}
where $\rho$ can be an arbitrary distribution defined over $\cS$. 
It has been recently discussed that (e.g. \cite{liu2019neural, lan2022policy}) setting $\rho$ as the stationary state distribution induced by an optimal policy $\pi^*$, denoted by $\nu^*$, can simplify the analyses of various algorithms. 
In this case, \eqref{eq:mdp_single_obj_raw} becomes 
\begin{align}
\textstyle
\label{eq:mdp_single_obj}
\min_{\pi} \cbr{f(\pi) \coloneqq \EE_{\nu^*} \sbr{V^{\pi}(s)}}, ~~ \mathrm{s.t.} ~~ \pi(\cdot|s) \in \Delta_{\cA}, \forall s \in \cS.
\end{align}
It is worth stressing here that the convergence result we obtain for \eqref{eq:mdp_single_obj} can be analogously established for the general weighted objective \eqref{eq:mdp_single_obj_raw}. 
We provide detailed discussion of this in Section \ref{subsec_discuss_assump}.

There has been a surge of interests in designing efficient first-order methods for directly searching the optimal policy \cite{agarwal2020optimality, cen2021fast, lillicrap2015continuous, schulman2015trust, shani2020adaptive}, despite the objective \eqref{eq:mdp_single_obj} being non-convex w.r.t. the policy   $\pi$ \cite{agarwal2020optimality}.
These methods utilize the  gradient information of objective \eqref{eq:mdp_single_obj} for policy improvement, and are hence termed policy gradient (PG) methods.
Basic policy gradient method, which corresponds to the project gradient descent applied to \eqref{eq:mdp_single_obj_raw},
 converges sublinearly  with exact gradients \cite{agarwal2020optimality}.
Natural policy gradient method \cite{kakade2001natural} further applies pre-conditioning with Fisher-information matrix, and yields better dimensional dependence  on top of the sublinear convergence \cite{agarwal2020optimality}.
Linearly converging PG methods have been discussed in \cite{bhandari2020note,cen2021fast,khodadadian2021linear}.
The analyses therein heavily exploit the contraction properties of the Bellman optimality condition and thus being less applicable to the stochastic setting \cite{cen2021fast}.
Connections between PG methods and the classical mirror descent method \cite{beck2003mirror, nemirovski2009robust, nemirovskij1983problem} have also been studied in \cite{schulman2015trust, shani2020adaptive, neu2017unified, peters2010relative}. 
Until recently, \cite{lan2022policy} proposes the policy mirror descent method and its stochastic variants, and establishes linear convergence in both deterministic and stochastic settings, as well as optimal sampling complexity bounds w.r.t. target accuracy (optimality gap). 
\cite{lan2022block} further proposes a partial update rule for solving MDPs with large state spaces, which evaluates and updates the policy for a subset of randomly selected states, and establishes  computational and sample complexities for different sampling distributions.

Despite the aforementioned progresses in algorithmic design and global convergence, 
these developments have left some important facets of PG methods either insufficiently addressed, or completely untouched. 

The first  facet concerns with the gap between the current theories and empirical performances of PG methods.
Empirically, it is believed that PG methods achieves local acceleration toward the end of the optimization process. 
 Yet  existing results establish local superlinear convergence, either with algorithm-dependent assumptions that are difficult to verify \cite{puterman2014markov, khodadadian2021linear}, or for a restricted MDP class (e.g., entropy-regularized MDPs \cite{cen2021fast}).
It is unclear whether there exists a general argument for establishing the superlinear convergence of PG methods.
More importantly, 
prior developments have established an $\tilde{\cO}(\abs{\cS}\abs{\cA} / \epsilon^2)$ sample complexity for certain PG variant \cite{lan2022policy}.
Superlinear convergence, if holds true, naturally begs the question whether an improved sample complexity can be attained as an implication of the faster convergence.

The second  facet, and surprisingly untouched by the current literature, concerns with the convergence behavior of the policy.
Indeed, except for the special class of regularized MDPs (e.g., entropy-regularized MDPs \cite{cen2021fast}) for which the optimal policy is unique, it is completely unclear whether PG methods exhibit the last-iterate convergence of the policy at all.
This is due to the fact that for un-regularized MDPs, 
there can be infinitely many optimal policies.
In this case,  only  asymptotic subsequence convergence to an optimal policy can be claimed based on  the value convergence \cite{lan2022policy, zhan2021policy, agarwal2020optimality}.
Even when PG methods exhibit the last-iterate convergence, it is also unclear whether one can characterize the limiting policy, as the number of optimal policies can be infinite.
This aspect also bears practical consequences, as different policies can exhibit completely different behavior despite sharing the same value, notably their robustness.

\vspace{0.05in}
{\bf Summary of Contributions.}
This paper is dedicated to address  the aforementioned under-explored facets of PG methods. 
In summary, our contributions mainly exist in the following aspects. 


First, 
we develop the homotopic policy mirror descent (HPMD) method, which performs the mirror descent type policy update with a diminishing regularization term. 
HPMD can be viewed as a simplification, both algorithmically and analytically,  to the approximate policy mirror descent  \cite{lan2022policy}.
The inclusion of diminishing regularization also shares similar spirit with the homotopic method in the statistics literature \cite{zhao2007stagewise, hastie2004entire, park2007l1}.
We unveil a phase transition in the convergence of HPMD.
Specifically, we first establish the global linear convergence of the optimality gap, and the weighted distance to the set of optimal policies, where the weights are defined via the gap values of the underlying MDP through the optimal state-action value function. 
We further validate the necessity of the gap-dependent weights by constructing a special class of MDPs with a gap-dependent lower bound for best-policy identification.
 We then discover the local superlinear convergence for both quantities,  in an assumption-free manner.

Second, we show that HPMD, with the negative entropy as the distance-generating function, exhibits the last-iterate convergence of the policy, despite the existence of potentially infinitely many optimal policies.
 In addition, we characterize the limiting policy as the optimal policy $\pi^*_U$ with the maximum entropy.
That is, $\pi_U^* (a|s) = 1/ \abs{\cA^*(s)}$ for any $a \in \cA^*(s)$, and $\pi_U^* (a|s) = 0$ otherwise,
where  $\cA^*(s) = \Argmin_{a \in \cA} Q^*(s,a)$ denotes the set of optimal actions at a given state.
This appears to be surprising since any Dirac measure $\delta_a$  with support $a \in \cA^*(s)$ suffices to become the optimal strategy (see Lemma \ref{lemma:optimal_policy_set}).
Accordingly, we term this phenomenon as the algorithmic (implicit) regularization of HPMD.

Third, we establish that all the aforementioned computational properties of HPMD hold for a general class of decomposable Bregman divergences.
In addition,  the generalized HPMD variants with this class of Bregman divergences  converge to the same limiting policy defined above.
As a by product of our analysis, we discover the finite-time exact convergence of HPMD with many common distance-generating functions, including $p$-th power of the $\ell_p$-norm, and the negative Tsallis entropy.

Finally, we develop the stochastic homotopic policy mirror descent (SHPMD) method and establish noise conditions that ensure similar convergence properties as deterministic HPMD.
Accordingly, we establish an $\tilde{\cO}(\abs{\cS} \abs{\cA} / \epsilon^2)$ sample complexity for finding an $\epsilon$-optimal policy.
We further show that by exploiting superlinear convergence of SHPMD, an improved sample complexity can be achieved with high probability,  when searching for a policy with a small optimality gap $\epsilon$. 
Informally, for $\epsilon_0 $ small enough, with probability $1 - \cO(\epsilon_0^{\scriptscriptstyle 1/3})$, 
 an $\epsilon$-optimal policy can be found with 
 $\tilde{\cO}(\abs{\cS} \abs{\cA} / \epsilon_0^2)$ samples,  for any $\epsilon < \epsilon_0$.
SHPMD also attains the last-iterate policy convergence almost surely,  with a slightly increased sample complexity.

To the best of our knowledge,  all the above findings appear to be new in the literature of PG methods. 
In particular,
the global convergence of the weighted distance to the set of optimal policies seem to be the first non-asymptotic characterization of policy convergence among PG methods.
 The superlinear convergence  is established without any assumption,
 dropping strict algorithmic assumptions required by existing results \cite{puterman2014markov, khodadadian2021linear}, 
 and holds for  both the policy and  optimality gap.
The last-iterate convergence of the policy has also not been established, in any form, prior to our development, neither does an exact characterization of  the limiting policy.
Finally, the improved sample complexity by exploiting superlinear convergence of stochastic HPMD also appears to be the first of its kind.


\begin{figure}[t!]
\makebox[\linewidth][c]{%
     \centering
     \begin{subfigure}[b]{0.25\textwidth}
         \centering
         \includegraphics[width=\textwidth]{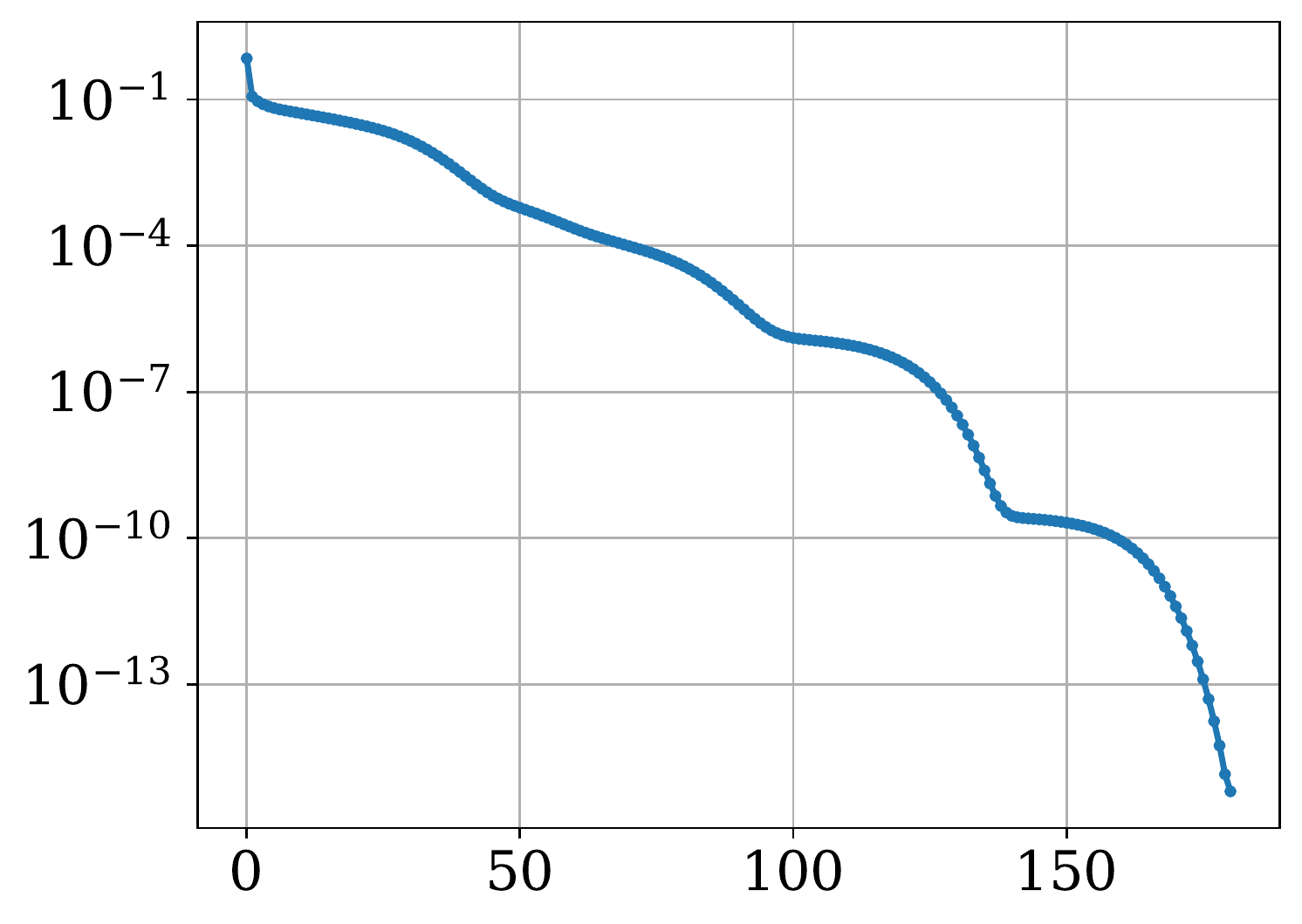}
             \vspace{-0.15in} 
             \caption{$f_\rho(\pi_k) - f_\rho(\pi^*)$.}
             \label{subfig:gap}
     \end{subfigure}
               \begin{subfigure}[b]{0.25\textwidth}
         \centering
         \includegraphics[width=\textwidth]{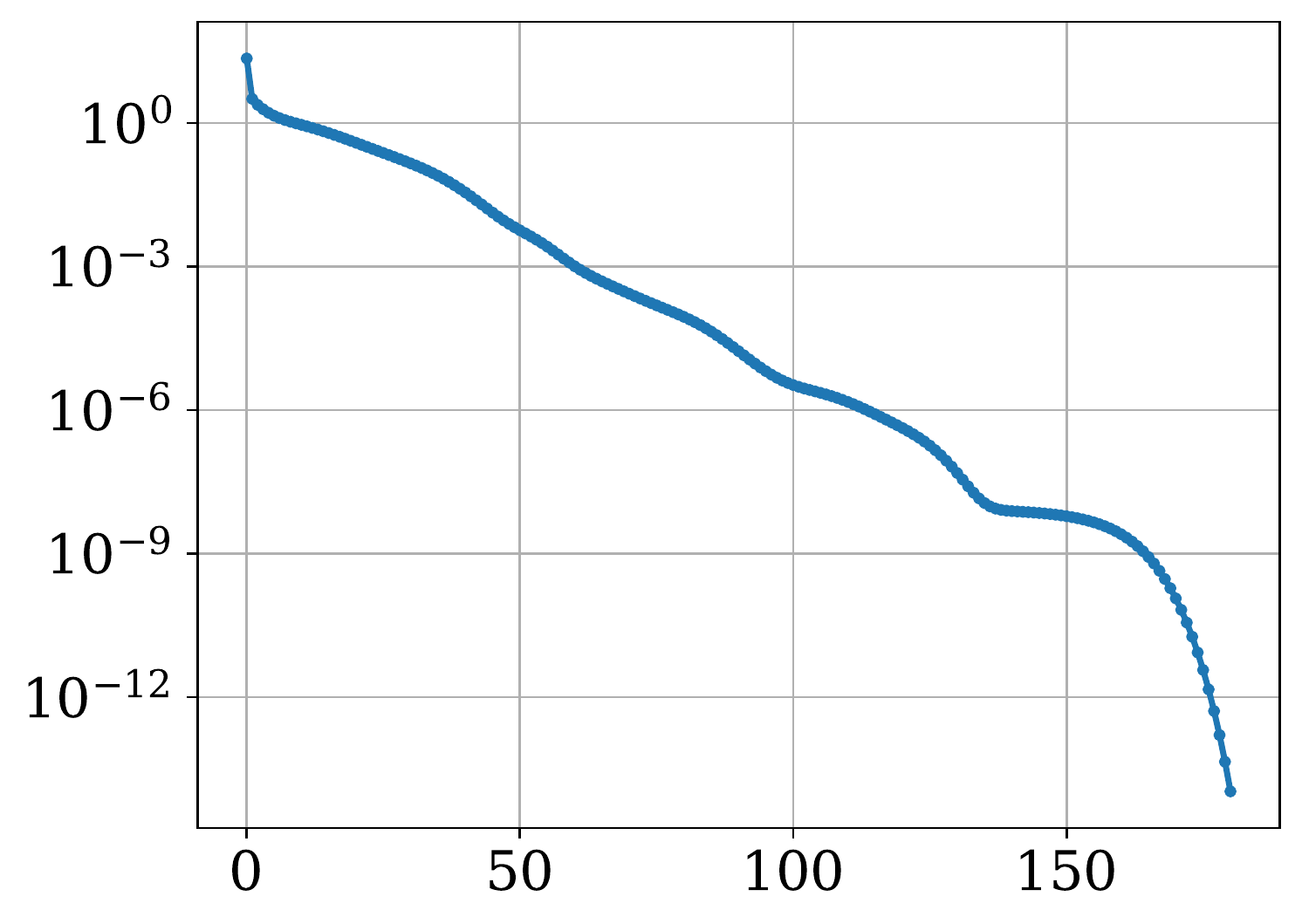}
             \vspace{-0.15in}
           \caption{ $\mathrm{dist}_{\rho, \cM}(\pi_k, \Pi^*)$. }
            \label{subfig:non_support_wsum}
     \end{subfigure}  
     \begin{subfigure}[b]{0.255\textwidth}
         \centering
         \includegraphics[width=\textwidth]{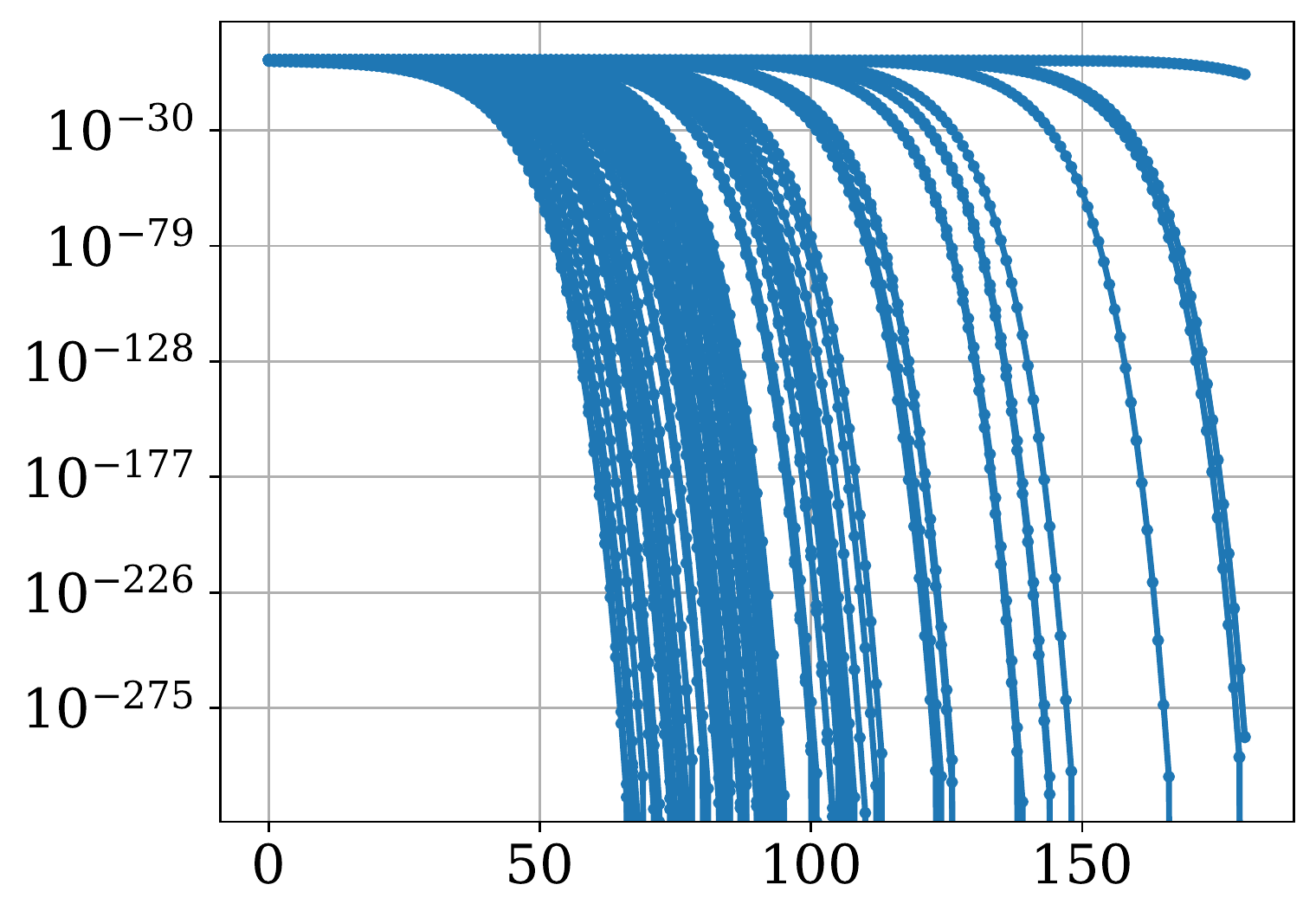}
             \vspace{-0.17in}
            \caption{ $\tsum_{a \notin \cA^*(s)} \pi_k(a|s)$. }
            \label{subfig:non_support_sum}
     \end{subfigure}  
             \begin{subfigure}[b]{0.24\textwidth}
         \centering
         \includegraphics[width=\textwidth]{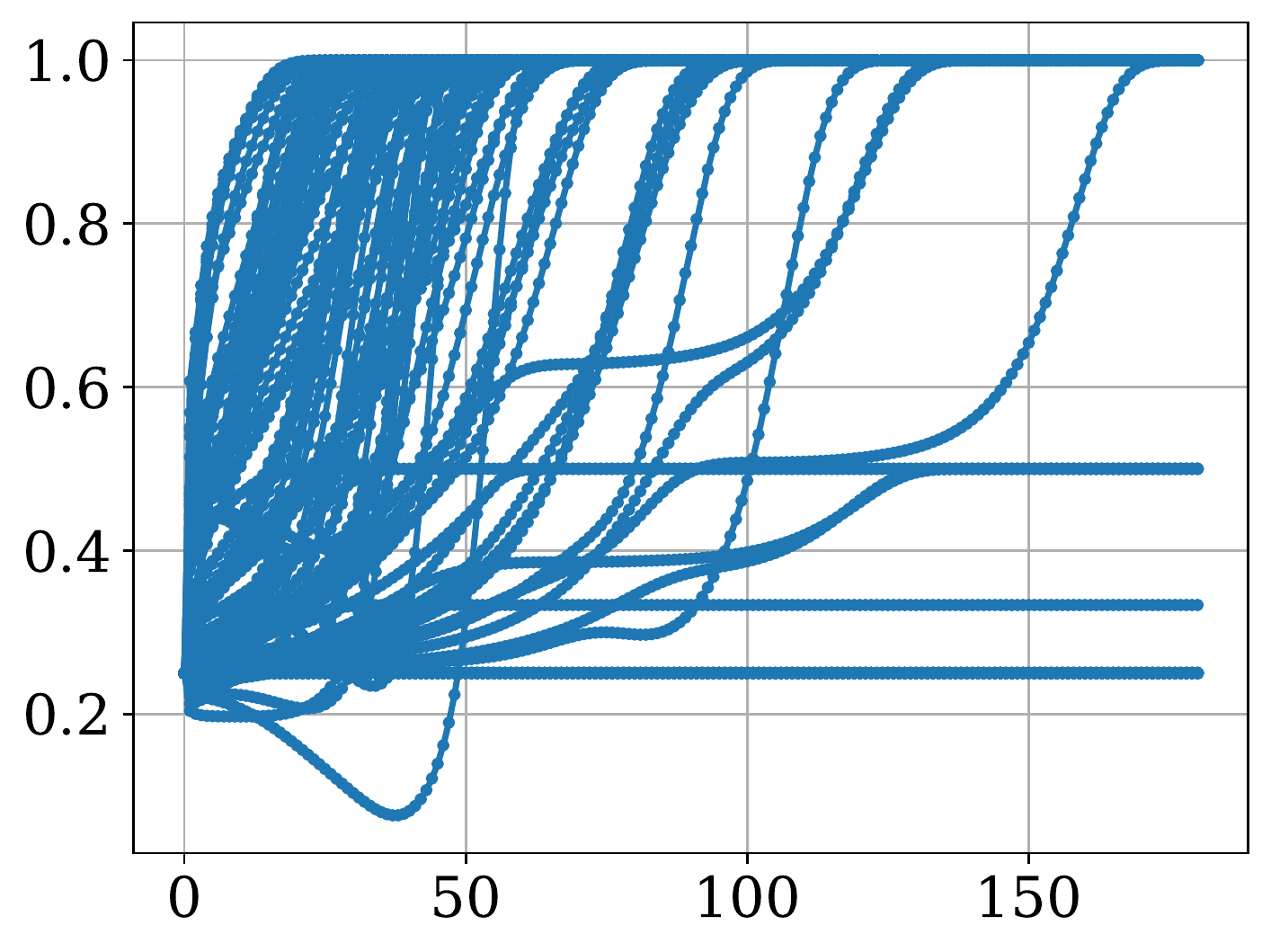}
             \vspace{-0.16in}
            \caption{ $ \min_{a \in \cA^*(s)} \pi_k(a|s)$. }
            \label{subfig:min_policy}
     \end{subfigure} 
     }    
    \vspace{-0.15in}
  \caption{ 
 Optimality gap and the policy of the linearly converging HPMD  on a randomly generated GridWorld MDP. 
Each line in Fig.  \ref{subfig:non_support_sum} and Fig. \ref{subfig:min_policy} corresponds to a single state.
Here we take $\rho = \mathrm{Unif}(\cS)$.
  }
  \vspace{-0.05in}
  \label{fig:motivation}
\end{figure}

\vspace{0.05in}
{\bf An Empirical Preview.}
To illustrate the applicability of the aforementioned results, we apply HPMD on the benchmark GridWorld environment ($\abs{\cS} = 100$)\footnote{For more environment details, we refer readers to  \cite{lan2022block}, which adopts  the same experiment setup. 
}, and report findings in Figure \ref{fig:motivation}.

Figure \ref{subfig:gap} and \ref{subfig:non_support_wsum} confirm the global convergence of the optimality gap, and the weighted distance (see Definition \ref{def_weighted_norm}) to the set of optimal policies. 
Both quantities exhibit phase transition from linear convergence to superlinear convergence. 
Moreover, Figure \ref{subfig:non_support_sum} also demonstrates a state-dependent acceleration effect for the policy convergence. 
Finally,  Figure \ref{subfig:min_policy}  implies that the limiting policy is sufficiently exploring every action within $ \cA^*(s)$. 
All these empirical observations will be covered by our ensuing technical discussions.


\vspace{0.05in}
{\bf Organization of the Paper.}
The rest of the paper is organized as follows. 
Section \ref{sec_deterministic} introduces the deterministic HPMD method, and establishes its global linear convergence for the optimality gap and the weighted distance to the set of optimal policies. 
Section \ref{sec_local_convergence_and_last_iterate} discusses the local superlinear convergence of both the policy and optimality gap, and the last-iterate convergence of the policy.
Section \ref{subsec_policy_convergence_decomposable} generalizes results in Section \ref{sec_deterministic} and \ref{sec_local_convergence_and_last_iterate} to a much more general class of separable Bregman divergences.
In Section \ref{sec:stochastic}, we extend our findings to the stochastic setting, and demonstrate that an improved sample complexity can be achieved by exploiting the local superlinear convergence.
Concluding remarks are made   in Section \ref{sec:discussion}.

\subsection{Notation and Terminology}
We use $\Pi^*$ to denote the set of optimal stationary randomized policies of \eqref{obj:state_wise}. 
We denote the stationary state distribution of a policy $\pi$ by $\nu^{\pi}$.
In addition, we define the discounted state visitation measure $d_s^{\pi}$ induced by policy $\pi$ as 
$
\textstyle
d_s^{\pi}(s' ) = (1-\gamma) \tsum_{t = 0}^{\infty} \gamma^t \PP^{\pi} (s_t = s' | s_0 = s),
$
where $\PP^{\pi}(s_t = s'|s_0 = s)$ denotes the probability of reaching state $s'$ at timestep t when starting at state $s$ and following policy $\pi$.
Accordingly, for any $\rho \in \cS$, we define $d_{\rho}^{\pi} (s) = \EE_{s' \sim \rho} d_{s'}^{\pi}(s)$.

For a pair of policies $\pi, \pi' \in \Pi$,  we define the Bregman divergence between $\pi(\cdot|s)$ and $\pi'(\cdot|s)$ as 
\begin{align}\label{eq_bregman_def}
D^{\pi}_{\pi'}(s) =
w(\pi(\cdot|s)) - w(\pi'(\cdot|s)) - \inner{\nabla w(\pi'(\cdot|s))}{\pi(\cdot|s) - \pi'(\cdot|s)},
\end{align}
where $w: \RR^{\abs{\cA}} \to \RR$ is a strictly convex,  also known as the distance-generating function. $\nabla w(p) \in \partial w(p)$ denotes a subgradient of $w$ at $p \in \RR^{\abs{\cA}}$.
Common distance-generating functions include $w(p) = \norm{p}_2^2$, which induces $D^{\pi}_{\pi'}(s) = \norm{\pi(\cdot|s) - \pi'(\cdot|s)}_2^2$;  
and the negative entropy function $w(p) = \tsum_{a \in \cA} p_a \log p_a \coloneqq - \cH(p)$, which induces  the Kullback-Leibler divergences:
\begin{align*}
D^{\pi}_{\pi'}(s) = \tsum_{a \in \cA} \pi(a|s) \log\rbr{\pi(a|s)/ \pi'(a|s)} \coloneqq \mathrm{KL}(\pi(\cdot|s) \Vert \pi'(\cdot|s)).
\end{align*}

For any distribution $\mu$ defined over set $\cX$, we denote its support as $\mathrm{supp}(\mu) = \{x \in \cX: \mu(x) > 0 \} $.
If $\mu, \nu \in \Delta_\cS$ and $\mathrm{supp}(\nu) = \cS$,
 we let $\norm{\mu / \nu}_\infty = \max_{s \in \cS} \mu(s) / \nu(s)$. 
We use $\mathrm{ReInt}(\cX)$ to denote the relative interior of $\cX$.
For any pair of policies $(\pi, \pi')$, we define their $\ell_1$ distance as
$\norm{\pi - \pi'}_{1} = \max_{s \in \cS} \norm{\pi(\cdot|s) - \pi'(\cdot|s)}_1$.
Similarly, we  also define $\norm{\pi - \pi'}_{\infty} = \max_{s \in \cS} \norm{\pi(\cdot|s) - \pi'(\cdot|s)}_\infty$.
We define the $\ell_1$ distance between a policy $\pi$ and a policy set $\Pi^*$ at state $s \in \cS$ as 
$\mathrm{dist}_{\ell_1}(\pi(\cdot|s), \Pi^*(\cdot|s) ) = \inf_{\pi' \in \Pi^*} \norm{\pi(\cdot|s)  - \pi' (\cdot|s)}_1$,
and accordingly let $\mathrm{dist}_{\ell_1}(\pi, \Pi^*) =  \max_{s \in \cS} \mathrm{dist}_{\ell_1}(\pi(\cdot|s), \Pi^*(\cdot|s) ) $.

\section{Homotopic Policy Mirror Descent}\label{sec_deterministic}
In this section, we introduce the deterministic homotopic policy mirror descent  (HPMD) method, and study its convergence properties in terms of both the optimality gap and the policy.
The HPMD method (Algorithm \ref{alg:fpmd_main}) minimizes the objective \eqref{eq:mdp_single_obj} by performing the following update at every iteration $k \geq 0$, 
\begin{align}\label{fpmd:update}
	\pi_{k+1} (\cdot| s) =
	\argmin_{p(\cdot|s) \in \Delta_{\cA}} \eta_k\sbr{ \inner{Q^{\pi_k}(s, \cdot)}{ p(\cdot| s ) }  + \tau_k D^{p}_{\pi_0} (s) } + D^{p}_{\pi_k}(s),  ~ \forall s \in \cS,
\end{align}
where $\pi_0$ denotes the uniform policy.
For the ease of determining constant terms, we  focus on using the negative entropy as the distance-generating function, that is,  $D^{\pi}_{\pi'}(s) = \mathrm{KL}(\pi(\cdot|s) \Vert \pi'(\cdot|s))$. 
As will be shown in Section \ref{subsec_policy_convergence_decomposable}, 
all the analyses and results in this section can be extended to general Bregman divergences without any essential change.

\begin{algorithm}[b!]
    \caption{The homotopic  policy mirror descent (HPMD) method}
    \label{alg:fpmd_main}
    \begin{algorithmic}
    \STATE{\textbf{Input:} Initial policy $\pi_0$, nonnegative parameters $\cbr{\tau_k} $, and stepsizes $\{\eta_k\}$.}
    \FOR{$k=0, 1, \ldots$}
	\STATE{Update policy:
			\vspace{-0.1in}
	\begin{align*}
	\pi_{k+1} (\cdot| s) =
	\argmin_{p(\cdot|s) \in \Delta_{\cA}} \eta_k\sbr{ \inner{Q^{\pi_k}(s, \cdot)}{ p(\cdot| s ) }  + \tau_k D^{p}_{\pi_0} (s) } + D^{p}_{\pi_k}(s),  ~ \forall s \in \cS
	\end{align*}}
	\vspace{-0.15in}
    \ENDFOR
    \end{algorithmic}
\end{algorithm}

In a nutshell, the HPMD method can be considered as a simplification of the approximate policy mirror descent (APMD) method proposed in \cite{lan2022policy}, and later extended in \cite{zhan2021policy}, by dropping the need for evaluating the perturbed state-action value function $Q^{\pi_k}_{\tau_k}$, defined as 
\begin{align*}
Q^{\pi}_{\tau} (s,a) & = \EE \sbr{\tsum_{t=0}^\infty \gamma^t \rbr{r(s_t, a_t) + \tau D^{\pi}_{\pi_0} (s_t) } \big|  (s_0, a_0) = (s,a), a_t \sim \pi(\cdot|s_t), s_{t+1} \sim \cP(\cdot|s_{t}, a_{t})},  \tau \geq 0.
\end{align*}
Thus aside from the conceptual simplicity, HPMD also allows an easier implementation without the requirement to incorporate the regularization term when evaluating the policy.

\subsection{Global Convergence of the Optimality Gap}\label{sec:algorithm}

In this subsection, we establish the global convergence of the HPMD method in terms of the optimality gap.
We begin by establishing the following lemma characterizing each update of HPMD.

\begin{lemma}\label{lemma_three_point}
For any $p\in \Pi$ and any $s\in\cS$,  we have 
\begin{align}\label{eq_three_point}
& \eta_k \inner{Q^{\pi_k}(s, \cdot)}{\pi_{k+1}(\cdot|s) - p(\cdot|s) } + \eta_k \tau_k \rbr{D^{\pi_{k+1}}_{\pi_0}(s) - D^{p}_{\pi_0}(s)  }
+ D^{\pi_{k+1}}_{\pi_k}(s) \nonumber \\
\leq &  D^{p}_{\pi_k}(s) - (\tau_k \eta_k + 1) D^{p}_{\pi_{k+1}}(s).
\end{align}
\end{lemma}

\begin{proof}
From the optimality condition of the HPMD update \eqref{fpmd:update}, we have for any $p \in \Pi$,
\begin{align}\label{fpmd_update_opt_condition_raw}
\eta_k \inner{Q^{\pi_k}(s, \cdot)}{p(\cdot|s) - \pi_{k+1}(\cdot|s)} + \eta_k \tau_k \inner{\nabla D^{\pi_{k+1}}_{\pi_0}(s)}{ p(\cdot|s) - \pi_{k+1}(\cdot|s)}
\nonumber \\ + \inner{\nabla D^{\pi_{k+1}}_{\pi_k}(s)}{ p(\cdot|s)  -\pi_{k+1}(\cdot|s)} \geq 0,
\end{align}
where $\nabla D^{\pi_{k+1}}_{\pi}(s)$ denotes a subgradient of $D^{\pi_{k+1}}_{\pi}(s)$ with respect to $\pi_{k+1}(\cdot|s)$.
Given the definition of Bregman divergence \eqref{eq_bregman_def}, we have the following identity
\begin{align*}
\inner{\nabla D^{\pi_{k+1}}_{\pi}(s)}{ p(\cdot|s) - \pi_{k+1}(\cdot|s)} &= D^{p}_{\pi}(s) - D^{\pi_{k+1}}_{\pi}(s) - D^{p}_{\pi_{k+1}}(s), ~\forall \pi \in \Pi,
\end{align*}
Combining the previous  observation with \eqref{fpmd_update_opt_condition_raw}, we immediately obtain the result.
\qed \end{proof}

Our development also makes use of  the following lemma, also known as the performance difference lemma in the literature (Lemma 2, \cite{lan2022policy}; see also \cite{kakade2002approximately}).

\begin{lemma}\label{lemma_perf_diff}
For any pair of policies $\pi, \pi'$, we have 
\begin{align}\label{ineq_perf_diff}
V^{\pi'}(s) - V^{\pi}(s)
= \frac{1}{1-\gamma} \EE_{s' \sim d^{\pi‘}_s} 
\inner{Q^{\pi}(s', \cdot)}{\pi'(\cdot|s') - \pi(\cdot|s')}.
\end{align}
\end{lemma}

Let $\nu^* = \nu^{\pi^*}$ be the stationary state distribution induced by the optimal policy $\pi^*$, and define $\phi(\pi, \pi^*) = \EE_{s \sim \nu^*} D^{\pi^*}_{\pi}(s)$.
We proceed to establish the generic convergence property of the HPMD method.

\begin{lemma}\label{deterministic_generic}
Suppose there exists $\cbr{\alpha_k}$ with $\alpha_k > 0$, which together with $\cbr{(\eta_k, \tau_k)}$, satisfy
\begin{align}\label{alpha_choice}
\alpha_k \geq \alpha_{k+1} \gamma, ~
\alpha_k \rbr{\frac{1}{\eta_k} + \tau_k}  \geq \frac{\alpha_{k+1}}{\eta_{k+1}},
\end{align}
then at any iteration $k \geq 0$, HPMD satisfies 
\begin{align}\label{deterministic_generic_recursion}
& \alpha_k \rbr{f(\pi_{k+1}) - f(\pi^*)}
 + \alpha_k \rbr{\frac{1}{\eta_k} + \tau_k} \phi(\pi_{k+1}, \pi^*) \nonumber \\
 \leq & 
\alpha_0 \gamma \rbr{f(\pi_{0}) - f(\pi^*) } + \frac{\alpha_0}{\eta_0} \phi(\pi_0, \pi^*)
+ \tsum_{t=0}^{k} \frac{3 \alpha_t \tau_t }{ 1-\gamma } \log \abs{\cA}.
\end{align}
\end{lemma}

\begin{proof}
For each $s \in \cS$, by plugging  $p = \pi_k$ into \eqref{eq_three_point}, we have 
\begin{align}\label{three_point_negative}
\eta_k \inner{Q^{\pi_k}(s, \cdot)}{\pi_{k+1}(\cdot|s) - \pi_k(\cdot|s) } 
+ \eta_k \tau_k \rbr{D^{\pi_{k+1}}_{\pi_0}(s) - D^{\pi_k}_{\pi_0}(s)  } \leq -(\tau_k \eta_k + 1) D^{\pi_k}_{\pi_{k+1}} \leq 0.
\end{align}
On the other hand, choosing $p = \pi^*$ in \eqref{eq_three_point}, we obtain 
\begin{align}
\eta_k \inner{Q^{\pi_k}(s, \cdot)}{\pi_{k}(\cdot|s) - \pi^*(\cdot|s) } + & \eta_k \inner{Q^{\pi_k}(s, \cdot)}{\pi_{k+1}(\cdot|s) - \pi_{k}(\cdot|s) } + \eta_k \tau_k \rbr{D^{\pi_{k+1}}_{\pi_0}(s) - D^{\pi^*}_{\pi_0}(s)  } \nonumber  \\
& \leq D^{\pi^*}_{\pi_k}(s) - (\tau_k \eta_k + 1) D^{\pi^*}_{\pi_{k+1}}(s). \label{three_point_decomposition}
\end{align}
We will make use of the following observations,
\begin{align}
\EE_{s \sim \nu^*} \inner{Q^{\pi_k}(s, \cdot)}{\pi_{k}(\cdot|s) - \pi^*(\cdot|s) }
& = \EE_{s \sim \nu^*} \EE_{s' \sim d_s^{\pi^*}}  \inner{Q^{\pi_k}(s', \cdot)}{\pi_{k}(\cdot|s') - \pi^*(\cdot|s') }  \nonumber \\
& = - \EE_{s \sim \nu^*} \EE_{s' \sim d_s^{\pi^*}}  \inner{Q^{\pi_k}(s', \cdot)}{\pi^*(\cdot|s') - \pi_k(\cdot|s') } \nonumber \\
& \overset{(a)}{=}  -(1-\gamma) \EE_{s \sim \nu^*} \rbr{V^{*}(s) - V^{\pi_k}(s) } \nonumber\\
& =(1-\gamma) \rbr{ f(\pi_k) - f(\pi^*)}, \label{diff_old_pi}
\end{align}
where $(a)$ uses Lemma \ref{lemma_perf_diff}. 
In addition, we also have 
\begin{align}
& (1-\gamma) \rbr{V^{\pi_{k+1}}(s) - V^{\pi_k}(s)} + \EE_{s' \sim d_{s}^{\pi_{k+1}}} \sbr{\tau_k \rbr{D^{\pi_{k+1}}_{\pi_0}(s')  - D^{\pi_{k}}_{\pi_0} (s')  } } \nonumber \\
\overset{(b)}{=} & \EE_{s' \sim d_{s}^{\pi_{k+1}}} \sbr{
\inner{Q^{\pi_k}(s', \cdot)}{ \pi_{k+1}(\cdot|s') - \pi_k(\cdot|s')} + \tau_k \rbr{D^{\pi_{k+1}}_{\pi_0}(s')  - D^{\pi_{k}}_{\pi_0} (s')  }
}  \nonumber \\
\overset{(c)}{\leq} & (1-\gamma) \sbr{
\inner{Q^{\pi_k}(s, \cdot)}{ \pi_{k+1}(\cdot|s) - \pi_k(\cdot|s)} + \tau_k \rbr{D^{\pi_{k+1}}_{\pi_0}(s)  - D^{\pi_{k}}_{\pi_0} (s)  }
}  \leq 0  \label{deterministic_non_monotone} ,
\end{align}
where $(b)$ uses again Lemma \ref{lemma_perf_diff}, and $(c)$ uses \eqref{three_point_negative}, and the fact that $d_s^{\pi_{k+1}}(s) \geq 1- \gamma$.
Thus from the previous relation, we obtain
\begin{align}\label{diff_new_pi}
\inner{Q^{\pi_k}(s, \cdot)}{ \pi_{k+1}(\cdot|s) - \pi_k(\cdot|s)} \geq 
V^{\pi_{k+1}}(s) - V^{\pi_k}(s) - \frac{2\tau_k}{1-\gamma} \max_{s\in \cS} \abs{ D^{\pi_{k+1}}_{\pi_0}(s')  - D^{\pi_{k}}_{\pi_0} (s') } .
\end{align}
By taking expectation w.r.t. $s \sim \nu^*$ in \eqref{three_point_decomposition},  and combining \eqref{diff_old_pi}, \eqref{diff_new_pi}, we obtain
\begin{align*}
& (1-\gamma) \sbr{f(\pi_k) - f(\pi^*)} + f(\pi_{k+1}) - f(\pi_k)  \\
\leq & 
\frac{1}{\eta_k} \phi(\pi_k, \pi^*) - \rbr{\frac{1}{\eta_k} + \tau_k}  \phi(\pi_{k+1}, \pi^*)
+ \frac{2 \tau_k}{1-\gamma} \max_{s\in \cS} \abs{ D^{\pi_{k+1}}_{\pi_0}(s')  - D^{\pi_{k}}_{\pi_0} (s') }  + \tau_k \max_{s \in \cS} \abs{D^{\pi_{k+1}}_{\pi_0}(s) - D^{\pi^*}_{\pi_0}(s)  } .
\end{align*}
After simple rearrangement, 
and using the fact that $ 0 \leq D^{\pi}_{\pi_0}(s) \leq \log \abs{\cA}$ given $\pi_0$ being the uniform policy, 
the previous relation becomes 
\begin{align}
&f(\pi_{k+1}) - f(\pi^*) + \rbr{\frac{1}{\eta_k} + \tau_k}   \phi(\pi_{k+1}, \pi^*) \leq
\gamma \rbr{f(\pi_{k}) - f(\pi^*)} + \frac{1}{\eta_k}  \phi(\pi_k, \pi^*)
+ \frac{3 \tau_k}{1-\gamma}  \log \abs{\cA}.
  \label{ineq_recursion_before_spec_parameters}
\end{align}
Now multiplying both sides of the  above inequality  by positive constant $\alpha_k$ satisfying 
$
\alpha_k \geq \alpha_{k+1} \gamma$ and 
$
\alpha_k (1/\eta_k + \tau_k) \geq {\alpha_{k+1}}/{\eta_{k+1}},
$
and then summing up from $t=0$ to $k$,  we obtain the desired result.
\qed \end{proof}

It should be noted that in view of \eqref{deterministic_non_monotone}, the optimality gap in HPMD is not necessarily monotonically decreasing.
This creates a separation between HPMD and other PG methods, as the monotonicity is heavily exploited in the existing analyses of PG methods \cite{lan2022policy, agarwal2020optimality}. 
Nevertheless, we proceed to establish both sublinear and linear convergence of HPMD, by exploiting Lemma \ref{deterministic_generic} with properly chosen parameters.

\begin{theorem}\label{thrm_sublinear_convergence}
Let $\eta_k = k + k_0$, $\tau_k = {1}/{(k+k_0)^2}$, where $k_0 = \ceil{{\gamma}/\rbr{1-\gamma}}$, then HPMD satisfies 
\begin{align*}
   f(\pi_{k+1}) - f(\pi^*)   \leq 
\frac{1}{k + k_0} 
\rbr{
k_0 \rbr{f(\pi_0) - f(\pi^*)} 
+ \frac{4 \log \rbr{3(k+k_0)} \log \abs{\cA}  }{1-\gamma}  
}.
\end{align*}
\end{theorem}

\begin{proof}
By taking  $\alpha_k = k + k_0$,
it can be verified that with the choice of $\cbr{(\eta_k, \tau_k ,\alpha_k)}$, condition \eqref{alpha_choice} holds, and hence one can apply Lemma \ref{deterministic_generic} and obtain 
\begin{align*}
   f(\pi_{k+1}) - f(\pi^*)   & \leq \alpha_k^{-1} \rbr{ 
\alpha_0 \gamma \rbr{f(\pi_{0}) - f(\pi^*) } + \frac{\alpha_0}{\eta_0} \phi(\pi_0, \pi^*)
+ \tsum_{t=0}^{k} \frac{3 \alpha_t \tau_t }{1-\gamma } \log \abs{\cA}
} \\
& \leq 
\frac{1}{k + k_0} 
\rbr{
k_0 \rbr{f(\pi_0) - f(\pi^*)} 
+ \frac{4 \log \rbr{3(k+k_0)} \log \abs{\cA}  }{1-\gamma}  
}.
\end{align*}
The proof is then completed.
\qed \end{proof}

Next, we show that
 proper specification of $\cbr{\tau_k} $ and $\cbr{\eta_k} $,  HPMD converges linearly to the minimum of the policy optimization objective \eqref{eq:mdp_single_obj}.

\begin{theorem}\label{thrm_convergence_fpmd}
By choosing $1+ \eta_k \tau_k = 1/\gamma$ and $\eta_k = \gamma^{-2(k+1)}$ in the HPMD method, then at any iteration $k \geq 1$, HPMD produces policy $\pi_k$ satisfying 
\begin{align*}
f(\pi_{k}) - f(\pi^*) 
\leq 
\gamma^k \rbr{
 f(\pi_{0}) - f(\pi^*)  + \frac{4  \log \abs{\cA}}{1-\gamma}
}.
\end{align*} 
\end{theorem}

\begin{proof}
By choosing $\alpha_k = \gamma^{-(k+1)}$, $\eta_k = \gamma^{-2(k+1)}$, one can readily verify that condition \eqref{alpha_choice} holds.
Thus from the recursion \eqref{deterministic_generic_recursion}, we obtain
\begin{align*}
\gamma^{-k} \rbr{f(\pi_{k}) - f(\pi^*)} + \gamma^{k-1} \phi(\pi_{k}, \pi^*) & \leq 
  f(\pi_{0}) - f(\pi^*)  + \gamma \phi(\pi_0, \pi^*)
+ \tsum_{t=0}^{k-1} \frac{3 \alpha_t}{\gamma \eta_t } \log \abs{\cA}  \\
& \leq  f(\pi_{0}) - f(\pi^*)  + \gamma \phi(\pi_0, \pi^*)
+  \tsum_{t=0}^{k-1}  3 \gamma^t \log \abs{\cA} \\
& \leq  f(\pi_{0}) - f(\pi^*)  + \frac{4  \log \abs{\cA}}{1-\gamma},
\end{align*}
from which we immediately obtain the desired result.
\qed \end{proof}

Note that the obtained linear convergence of HPMD comes with a much simplified analysis compared to existing linearly converging PG methods, namely the APMD method \cite{lan2022policy, zhan2021policy}.
See also \cite{xiao2022convergence} for another algorithmic simplification of APMD, which further drops the vanishing regularization term.
As will be clarified in our ensuing discussions, this vanishing regularization term in HPMD is the foremost factor in obtaining a precise characterization of the limiting policy.

\subsection{Global Convergence of the Policy}\label{subsec_global_policy}

Before stating the formal results, 
the following assumption is posed for the remainder of our discussions. 
It is worth stressing here that Assumption \ref{assump_support} is posed only for the purpose of presentation simplicity,
and can be removed with a slightly modified analysis.
We provide detailed discussions in  Section \ref{subsec_discuss_assump}. 

\begin{assumption}\label{assump_support}
There exists an optimal policy $\pi^* \in \Pi^*$ such that $\nu^{\pi^*}$ has full support on $\cS$.
Accordingly, we define the discounted distribution mismatch ratio 
$\varrho \coloneqq \gamma \max_{s \in \cS, a \in \cA} \lVert{ \cP(\cdot  |s, a) / \nu^{\pi^*}}\rVert_\infty$.
\end{assumption}

Next, we  define the gap function of an MDP instance, a recurring quantity in our ensuing discussions.

\begin{definition}[Gap Value]\label{def_gap}
Fix an MDP instance $\cM$.
For each state-action pair $(s,a) \in \cZ \coloneqq \cS \times \cA$, we define its gap value $\delta_{\cZ, \cM} (s,a) = Q^*(s,a) - \min_{a \in \cA} Q^*(s,a)$. 
In addition,  let $\delta_{\cS, \cM}^*(s) = \min_{a  \notin \cA^*(s)} Q^*(s,a) - \min_{a \in \cA} Q^*(s,a) $ if $\cA^*(s) \neq \cA$, and $ \delta_{\cS, \cM}^*(s) = \infty$ otherwise, then
the gap function of the MDP $\cM$ is defined as $\Delta^*(\cM) = \min_{s \in \cS} \delta_{\cS, \cM}^*(s)$.
\end{definition}

Before proceeding, let us  recall a folklore characterization of optimal stationary policies.
\begin{lemma}[Characterization of Optimal Policies]\label{lemma:optimal_policy_set}
Let $Q^* \in \RR^{\abs{\cS} \times \abs{\cA}}$ denote the optimal $Q$-function.
Then the set of optimal stationary policies $\Pi^*$ is given by
\begin{align*}
\textstyle
\Pi^* = \cbr{
\pi \in \Pi: \mathrm{supp}(\pi(\cdot|s)) \subseteq \Argmin_{a \in \cA} Q^*(s, a), ~ \forall s \in \cS
}.
\end{align*}
\end{lemma}
\begin{proof}
The claim is a direct consequence of Lemma \ref{lemma_perf_diff} by taking $\pi = \pi^*$ therein.
\qed \end{proof}

By definition, $\Delta^*(\cM) > 0$. 
In addition, in view of Lemma \ref{lemma:optimal_policy_set}, whenever $\Delta^*(\cM) = \infty$, then any policy is optimal, and consequently there is not need for planning.
Hence going forward, we only consider the scenario when 
$\Delta^*(\cM) < \infty$.

\begin{definition}\label{def_weighted_norm}
For any $\rho \in \Delta_{\cS}$ with $\mathrm{supp}(\rho) = \cS$, 
and an MDP instance $\cM$,
we define the seminorm $\norm{\cdot}_{\cM}: \RR^{\abs{\cS} \times \abs{\cA}} \mapsto \RR_+$,  as  
$
\norm{x}_{\rho, \cM}  = \tsum_{s \in \cS}  \tsum_{a \in \cA}  \rho(s) \delta_{\cZ, \cM} (s,a)  \abs{ x(s,a) },
$
and its induced weighted-distance to $\Pi^*$ as  $\mathrm{dist}_{\rho, \cM} (\pi, \Pi^*) = \inf_{\pi^* \in \Pi^*} \lVert \pi - \pi^* \rVert_{\rho, \cM}$,
where we identify $\Pi$ as a subset in $\RR^{\abs{\cS} \times \abs{\cA}}$.
\end{definition}

From Definition  \ref{def_gap}, \ref{def_weighted_norm}, and Lemma \ref{lemma:optimal_policy_set},  it is clear that $\mathrm{dist}_{\rho, \cM} (\pi, \Pi^*) = 0$ if and only $\pi \in \Pi^*$.
Our next result shows that for $\cbr{\pi_k}$ generated by HPMD, $\cbr{\mathrm{dist}_{\rho, \cM} (\pi_k, \Pi^*) }$ converges to zero at the same rate as the optimality gap, thus establishing the policy convergence to the set of optimal policies.

\begin{proposition}\label{prop_weighted_policy_convergence_linear}
Under the same settings in Theorem \ref{thrm_convergence_fpmd}, 
for any $\rho \in \Delta_{\cS}$ with $\mathrm{supp}(\rho) = \cS$, 
it holds that 
\begin{align*}
\mathrm{dist}_{\rho, \cM} (\pi_k, \Pi^*)  
\leq 
   \norm{\frac{\rho}{\nu^*}}_\infty^2  \norm{\tfrac{d_\rho^{\pi^*}}{\rho}}_\infty 
 \frac{\gamma^k }{1-\gamma}
 \rbr{
\mathrm{dist}_{\rho, \cM} (\pi_0, \Pi^*)
+ 4 \log \abs{\cA}
 } .
\end{align*}
\end{proposition}

\begin{proof}
For any policy $\pi$ and any $\pi^* \in \Pi^*$, we obtain from Lemma \ref{lemma_perf_diff} that 
\begin{align*}
 (1- \gamma) \rbr{V^{\pi}(s) - V^*(s)} = \EE_{s' \in d_s^{\pi}} \inner{Q^*(s',\cdot)}{\pi(\cdot|s') - \pi^*(\cdot|s')}   \overset{(a)}{\geq} (1-\gamma)  \inner{Q^*(s, \cdot)}{\pi(\cdot|s) - \pi^*(\cdot|s)} ,
\end{align*}
where $(a)$ follows from Lemma \ref{lemma:optimal_policy_set}, and the fact that $d_{s}^{\pi}(s) \geq (1-\gamma)$. 
Setting $\pi = \pi_k$, and further taking expectation with respect to $s \sim \rho$ in the above inequality,  it holds that
\begin{align}
\mathrm{dist}_{\rho, \cM} (\pi_k, \Pi^*)  & \overset{(b)}{=} 
\EE_{s \sim \rho} \inner{Q^*(s, \cdot)}{\pi_k(\cdot|s) - \pi^*(\cdot|s)} \nonumber \\
& \leq \EE_{s \sim \rho}
\rbr{ V^{\pi_k}(s) - V^*(s)} \nonumber\\
& \overset{(c)}{\leq} 
 \norm{\frac{\rho}{\nu^*}}_\infty  
 \gamma^k 
 \sbr{
 \EE_{s \sim \nu^*}
\rbr{ V^{\pi_k}(s) - V^*(s)}
+ \frac{4 \log \abs{\cA}}{1-\gamma} 
 } \nonumber\\
 & \leq 
  \norm{\frac{\rho}{\nu^*}}_\infty^2  
 \gamma^k 
 \sbr{
 \EE_{s \sim \rho}
\rbr{ V^{\pi_k}(s) - V^*(s)}
+ \frac{4 \log \abs{\cA}}{1-\gamma} 
 } \nonumber\\
 &\overset{(d)}{ = }
  \norm{\frac{\rho}{\pi^*}}_\infty^2  
 \frac{\gamma^k }{1-\gamma}
 \rbr{
\EE_{s \sim d_\rho^{\pi^*}} \inner{Q^*(s,\cdot)}{\pi_0(\cdot|s) - \pi^*(\cdot|s)}
+ 4 \log \abs{\cA}
 } \label{mismatch_ratio_in_weighted_policy}\\
 & \leq 
  \norm{\frac{\rho}{\nu^*}}_\infty^2  \norm{\tfrac{d_\rho^{\pi^*}}{\rho}}_\infty 
 \frac{\gamma^k }{1-\gamma}
 \rbr{
\EE_{s \sim \rho} \inner{Q^*(s,\cdot)}{\pi_0(\cdot|s) - \pi^*(\cdot|s)}
+ 4 \log \abs{\cA}
 } \nonumber \\
 & = 
   \norm{\frac{\rho}{\nu^*}}_\infty^2  \norm{\tfrac{d_\rho^{\pi^*}}{\rho}}_\infty 
 \frac{\gamma^k }{1-\gamma}
 \rbr{
\mathrm{dist}_{\rho, \cM} (\pi_0, \Pi^*)
+ 4 \log \abs{\cA}
 } , \nonumber
\end{align} 
where $(b)$ follows from Definition \ref{def_gap},  \ref{def_weighted_norm} and Lemma \ref{lemma:optimal_policy_set};
$(c)$ follows from Theorem \ref{thrm_convergence_fpmd};
and $(d)$ follows from Lemma \ref{lemma_perf_diff}.
The proof is then completed.
\qed \end{proof}

With the same arguments, we can also establish the policy convergence of sublinearly converging HPMD.

\begin{proposition}\label{prop_weighted_policy_convergence_sublinear}
Under the same settings in Theorem \ref{thrm_sublinear_convergence}, 
for any $\rho \in \Delta_{\cS}$ with $\mathrm{supp}(\rho) = \cS$, 
it holds that 
\begin{align*}
\mathrm{dist}_{\rho, \cM} (\pi_k, \Pi^*)  
\leq 
   \norm{\frac{\rho}{\nu^*}}_\infty^2  \norm{\tfrac{d_\rho^{\pi^*}}{\rho}}_\infty 
 \frac{\gamma^k \rbr{
k_0 \cdot \mathrm{dist}_{\rho, \cM} (\pi_0, \Pi^*)
+ 4 \log \rbr{3(k+k_0)} \log \abs{\cA}
} }{(1-\gamma) (k + k_0)}
 .
\end{align*}
\end{proposition}

In view of Proposition \ref{prop_weighted_policy_convergence_linear} and \ref{prop_weighted_policy_convergence_sublinear}, 
the policy converges to the set of optimal policies at a rate similar to that of the optimality gap, measured in the weighted distance $\mathrm{dist}_{\rho, \cM}$. 
It should be noted that the policy convergence is also non-monotonic. In particular,  $\cbr{\pi_k(a|s)}$ is not monotonically increasing for $a \in \cA^*(s)$, as illustrated by Figure \ref{subfig:min_policy}.
We will also construct a provable example of this phenomenon in Section \ref{subsec_tight_gap_dependence}.

We now briefly conclude our discussions in this section. 
Theorem \ref{thrm_convergence_fpmd} and Proposition \ref{prop_weighted_policy_convergence_linear} imply that to find an $\epsilon$-optimal policy,
in terms of both the optimality gap and the weighted distance to $\Pi^*$, 
 HPMD requires at most $\cO (\log(1/\epsilon))$  iterations, which seems to prescribe the initial convergence in Figure \ref{subfig:gap} and \ref{subfig:non_support_wsum} faithfully.
On the other hand, the linear convergence seems pessimistic when describing the convergence in the second stage therein.
Moreover, Figure \ref{subfig:non_support_sum} suggests that the local acceleration of policy convergence takes effect in a state-dependent manner, 
a phenomenon not captured by Proposition \ref{prop_weighted_policy_convergence_linear} and \ref{prop_weighted_policy_convergence_sublinear}.
Theses aforementioned remarks then serve as the motivation for our ensuing discussions in Section \ref{sec_local_convergence_and_last_iterate}.

%

\section{Local Acceleration and the Last-iterate Policy Convergence}\label{sec_local_convergence_and_last_iterate}
Our discussions in this section start by first focusing on the linearly converging HPMD variant,  specified in Theorem \ref{thrm_convergence_fpmd}.
Specifically, we establish in Section \ref{subsec_hpmd_suplinear} that HPMD   exhibits local superlinear convergence, 
and the acceleration applies to both the optimality gap and the policy convergence.
In particular, the convergence of the set of optimal policies comes with an analysis that can be readily extended to provide a state-dependent characterization.
More importantly, Section \ref{subsec_last_iterate} establishes the last-iterate convergence of the policy, thus validating that the nontrivial probability lower bound of the limiting policy, observed in Figure \ref{subfig:min_policy}, holds in general scenarios.

We then discuss, in Section \ref{subsec_results_sublinear_hpmd}, similar computational behavior for the sublinearly converging HPMD, specified in Theorem \ref{thrm_sublinear_convergence},  thus  demonstrating the generality of the local acceleration and the last-iterate policy convergence as algorithmic properties of HPMD.

\subsection{Local Superlinear Convergence}\label{subsec_hpmd_suplinear}
The following theorem establishes the local superlinear convergence of the policy to $\Pi^*$.

\begin{theorem}[Local Superlinear Convergence]\label{thrm:convergence_to_optimal}
With $1+ \eta_k \tau_k = 1/\gamma$ and $\eta_k = \gamma^{-2(k+1)}$,  HPMD satisfies
\begin{align}\label{eq_local_policy_convegence_hpmd_linear}
\mathrm{dist}_{\ell_{1}} (\pi_{k+1} ,\Pi^*)  \leq   2 C_\gamma \abs{\cA}
\exp \rbr{ - {\Delta^*(\cM)}  \gamma^{-2k - 1} /2  }, 
\end{align} 
for any iteration $k \geq K_1 \coloneqq   { 3 \log_\gamma \rbr{ \frac{ \Delta^*(\cM) (1-\gamma) }{ 2\varrho( 4 \log \abs{\cA} + C) }} }$, and $C_\gamma =  \exp \rbr{ \frac{2 C  }{(1-\gamma^3)(1-\gamma) \gamma}  }$.
\end{theorem}

\begin{proof}
Note that  
$
Q^{\pi_k}(s,a) - Q^*(s,a) = \gamma \tsum_{s' \in \cS} \cP (s'| s,a) \sbr{V^{\pi_k}(s') - V^*(s')}
$
from the definition of state-action value function.
Combining  this observation with Assumption \ref{assump_support}  and Theorem \ref{thrm_convergence_fpmd}, we obtain
\begin{align}\label{q_convergence}
Q^{\pi_k}(s,a) - Q^*(s,a) \leq   \gamma^k 
\varrho(4  \log \abs{\cA} + C )/ (1-\gamma)
,
\end{align}
where $C$ denotes the upper bound of the cost.
We proceed by inspecting the update rule of HPMD. Directly examining the optimality condition of the update \eqref{fpmd:update}, we obtain
\begin{align}
\label{update_recursion_crude}
\log \pi_{k+1}(a|s) = \frac{\log \pi_k(a|s) - \eta_k Q^{\pi_k}(s,a) }{ \eta_k \tau_k + 1} + \zeta_k(s), ~~ \forall (s,a) \in \cS \times \cA,
\end{align}
where $\zeta_k(s)$ denotes the normalization constant to make the right hand side of \eqref{update_recursion_crude}  the entry-wise logarithm of a valid probability vector in $\mathrm{ReInt} ( \Delta_{\cA})$. 
Now since $1+\tau_k \eta_k = 1/\gamma$, we have 
\begin{align}
\label{update:recursion}
\log \pi_{k+1}(a|s) = \gamma \sbr{\log \pi_k(a|s) - \eta_k Q^{\pi_k}(s,a) } + \zeta_k(s), ~~ \forall (s,a) \in \cS \times \cA.
\end{align}

We define $z_i^k(s) = \log \pi_{k}(i |s)$ for any pair $(s, i) \in \cS \times \cA$. 
For the ease of presentation, let us fix a state $s \in \cS$ and write $z_i^k$ in short for $z_i^k(s)$,  $Q_i^k$ in short for $Q^{\pi_k}(s,i)$, 
and $\zeta_k$ in short for $\zeta_k(s)$, 
when the context is clear. 
By applying \eqref{update:recursion} recursively, we obtain that for any $i \in \cA$,
$
z_i^{k+1} = \gamma^{k+1} z_i^0 - \tsum_{t=0}^{k} \gamma^{k+1 - t} \eta_t Q_i^t + \tsum_{t=0}^k \gamma^{k-t} \zeta_t.
$
Hence for any pair of actions $i, j \in \cA$, we have 
\begin{align}\label{difference_logit}
z_i^{k+1} - z_j^{k+1} = \gamma^{k+1} (z_i^0 - z_j^0) - \tsum_{t=0}^{k} \gamma^{k+1 - t} \eta_t ( Q_i^t - Q_j^t) = - \tsum_{t=0}^{k} \gamma^{k+1 - t} \eta_t ( Q_i^t - Q_j^t),
\end{align}
where the last equality uses the fact that $\pi_0$ is the uniform policy.

From Lemma \ref{lemma:optimal_policy_set}, 
to show $\pi_k \to \Pi^*$, 
it suffices to establish that $\pi_k(a|s) \to 0$ for $a \notin \cA^*(s)$.
To this end, we consider any pair of action $i, j$ with $j \notin \cA^*(s)$ and $ i \in \cA^*(s)$. 
Note that from Lemma \ref{lemma:optimal_policy_set} we must have $Q^*(s, i) < Q^*(s,j)$.
Combining this  observation with \eqref{q_convergence}, then there exists $\underline{K}_1(s)$ such that 
\begin{align}\label{diff_q_convergence}
Q_i^t - Q_j^t \leq (Q^*(s, i) - Q^*(s,j))/2 < 0, ~ \forall t \geq \underline{K}_1(s).
\end{align}
Recall that we choose $\eta_k = \gamma^{-2(k+1)}$ in HPMD.
Thus for any $k > \underline{K}_1(s)$, given \eqref{difference_logit}, we have 
\begin{align*}
z_i^{k+1} - z_j^{k+1}
& = - \tsum_{t=0}^{ \underline{K}_1(s) } \gamma^{k+1 - t} \eta_t ( Q_i^t - Q_j^t)
-  \tsum_{t=\underline{K}_1(s) + 1}^{ k } \gamma^{k+1 - t} \eta_t ( Q_i^t - Q_j^t) \\
& = - \tsum_{t=0}^{ \underline{K}_1(s) } \gamma^{k-3t-1} ( Q_i^t - Q_j^t)
-  \tsum_{t=\underline{K}_1(s) + 1 }^{ k } \gamma^{k-3t-1}( Q_i^t - Q_j^t) \\
& \overset{(a)}{\geq} - \tsum_{t=0}^{ \underline{K}_1(s) } \gamma^{k-3t-1} \cdot \frac{2C}{1-\gamma}
+ \frac{ \gamma^{-2k - 1}}{2} \sbr{Q^*(s, j) - Q^*(s,i)}  \\
& \geq - \frac{2 C \gamma^{k-3\underline{K}_1(s) }}{(1-\gamma^3)(1-\gamma) \gamma} 
+ \frac{\gamma^{-2k - 1}}{2} \sbr{Q^*(s, j) - Q^*(s,i)} ,
\end{align*}
where $(a)$ uses \eqref{diff_q_convergence}.
Hence
\begin{align}
z_i^{k+1} - z_j^{k+1}
 \geq - \frac{2 C }{(1-\gamma^3)(1-\gamma) \gamma} 
+ \frac{\gamma^{-2k - 1}}{2} \sbr{Q^*(s, j) - Q^*(s,i)} , ~ \forall k \geq 3 \underline{K}_1(s). \label{logit_diff_exp}
\end{align}

Given  \eqref{q_convergence}, it suffices to choose 
$
\underline{K}_1(s) \geq  { \log_\gamma \rbr{ \frac{ \sbr{Q^*(s,j) - Q^*(s,i)} (1-\gamma) }{ 2\varrho\rbr{ 4 \log \abs{\cA} + C} }}},
$
so that  \eqref{diff_q_convergence} is satisfied.
Combining this observation with \eqref{logit_diff_exp}, we know that for $k \geq K_1 =   { 3 \log_\gamma \rbr{ \frac{ \Delta^*(\cM) (1-\gamma) }{ 2\varrho\rbr{ 4 \log \abs{\cA} + C} }} }$,  
\begin{align*}
z_i^{k+1}(s) - z_j^{k+1}(s) \geq 
 - \frac{2 C }{(1-\gamma^3)(1-\gamma) \gamma} 
+  \frac{\gamma^{-2k - 1}}{2} \Delta^*(\cM)  , ~ \forall s \in \cS.
\end{align*}
Thus  from the definition of $(z_i^k, z_j^k)$, we obtain that for any $j \notin \cA^*(s)$ and any $k \geq K_1$, 
\begin{align}
\pi_{k+1}(j | s) & \leq \pi_{k+1}(i|s) \exp \rbr{ \frac{2 C  }{(1-\gamma^3)(1-\gamma) \gamma} 
-  \frac{\Delta^*(\cM)}{2} \gamma^{-2k - 1} } \leq  C_\gamma
\exp \rbr{- {\Delta^*(\cM)} \gamma^{-2k - 1} /2 }, \label{state_wise_support_recovery}
\end{align}
where $C_\gamma =  \exp \rbr{ \frac{2 C  }{(1-\gamma^3)(1-\gamma) \gamma}  }$. 
Hence from the characterization of the optimal policies in Lemma \ref{lemma:optimal_policy_set}, we conclude that 
$
\mathrm{dist}_{\ell_{1}} (\pi_{k+1} ,\Pi^*)  \leq    2 C_\gamma \abs{\cA}
\exp \rbr{ - {\Delta^*(\cM)} \gamma^{-2k - 1} /2  }$,
 for any $ k \geq K_1
$.
\qed \end{proof}

In view of Theorem \ref{thrm:convergence_to_optimal},  $\cbr{\pi_k}$ begins to converge superlinearly to $\Pi^*$ within $\cO(\log_\gamma ((1-\gamma) \Delta^*(\cM)) ) $ iterations. 
Local acceleration of  $\mathrm{dist}_{\rho, \cM} (\pi_k, \Pi^*)$ then follows immediately by noting that $\mathrm{dist}_{\rho, \cM} (\pi_k, \Pi^*) \leq C \mathrm{dist}_{\ell_1}(\pi_k ,\Pi^*) / (1-\gamma) $.
One can also easily adapt the analysis to show that for each state $s \in \cS$, the superlinear convergence of
$\mathrm{dist}_{\ell_1} (\pi_k(\cdot|s), \Pi^*(\cdot|s))$ 
takes effect
within $\cO(\log_\gamma ( (1-\gamma) \delta_{\cS,\cM}^*(s) )) $ iterations, where $\delta_{\cS, \cM}(\cdot)$ is defined in Definition \ref{def_gap}.
This state-dependent acceleration of policy convergence is also consistent with Figure \ref{subfig:non_support_sum}.
In addition, by having the local convergence of the policy, the local convergence of the optimality gap  follows by  invoking Lemma \ref{lemma_perf_diff}. 
Specifically, we have the following corollary.

\begin{corollary}\label{corollary:value_convergence}
Assume the same settings in Theorem \ref{thrm:convergence_to_optimal}.
For  any policy $\pi$ and $\pi^* \in \Pi^*$,  
$
V^{\pi}(s) - V^{\pi^*}(s)  \leq \frac{C }{(1-\gamma)^2} \norm{\pi - \pi^*}_{1} .
$
In particular, for $k \geq K_1$,
\begin{align*}
V^{\pi_{k+1}}(s) - V^{*}(s) \leq \frac{ 2 C \abs{\cA}  C_\gamma}{(1-\gamma)^2} \exp \rbr{- \frac{{\Delta^*(\cM)} \gamma^{-2k - 1}}{2} } , ~\forall s\in \cS.
\end{align*}
Consequently, we have $f(\pi_{k+1}) - f(\pi^*) \leq \frac{ 2 C \abs{\cA}  C_\gamma}{(1-\gamma)^2} \exp \rbr{ - \frac{{\Delta^*(\cM)} \gamma^{-2k - 1}}{2} }$ for all $k \geq K_1$.
\end{corollary}

\begin{proof}
In view of Lemma \ref{lemma_perf_diff}, we have
\begin{align}
V^{\pi_k}(s) - V^{\pi^*}(s) &=\frac{1}{1-\gamma} \EE_{s' \sim d_s^{\pi_k}} \inner{Q^*(s', \cdot)}{ \pi_k(\cdot|s') - \pi^*(\cdot|s')} \nonumber \\
& \leq \frac{1}{1-\gamma} \EE_{s' \sim d_s^{\pi_k}}  \norm{Q^*(s', \cdot)}_{\infty} \norm{ \pi_k(\cdot|s') - \pi^*(\cdot|s')}_{1} \nonumber \\
& \leq \frac{C }{(1-\gamma)^2} \norm{\pi_k - \pi^*}_{1}   ,
 \label{ineq_from_dist_to_value}
\end{align} 
where the last inequality uses $\norm{Q^*(\cdot, \cdot)}_{\infty} \leq C/(1-\gamma)$.
Letting $\pi^* \in \Argmin_{\pi' \in \Pi^*} \norm{\pi'  - \pi_k }_1$ in \eqref{ineq_from_dist_to_value},  and using the definition of $ \mathrm{dist}_{\ell_{1}} (\pi_k ,\Pi^*)$,
the claim then follows immediately after applying Theorem \ref{thrm:convergence_to_optimal}.
\qed \end{proof}

We compare our results with related literature.
Existing superlinear convergence for PG methods when solving non-regularized MDPs has only been established for  the optimality gap,  by exploiting its connection with policy iteration \cite{khodadadian2021linear, bhandari2020note}, and the superlinear convergence of the latter method \cite{puterman2014markov}.
However, such approaches assume strong  assumptions on bounding the difference between the current and the optimal policy in their induced transition kernels,  by the optimality gap (see, e.g., Theorem 6.4.8, \cite{puterman2014markov}), which seems difficult to verify given its algorithm-dependent nature.

In contrast to existing literature, our obtained result holds in an assumption-free manner, with Assumption \ref{assump_support} posed only for presentation simplicity.
The result establishes local superlinear convergence of the policy after a finite number of iterations, which also closely parallels the algorithmic behavior of Newton's method \cite{nocedal2006numerical}. 
It is also worth mentioning that by extending similar arguments, the local superlinear convergence can also be established for the approximate policy mirror descent method \cite{lan2022policy, zhan2021policy}.

\subsection{The Necessity of Dependence on Gap Values}\label{subsec_tight_gap_dependence}
The  linearly converging HPMD instantiated in Theorem \ref{thrm_sublinear_convergence} has an update  approaching  that of the Howard's policy iteration (PI) method, and both methods converge linearly.
It is also well known that PI converges in finite number iterations \cite{puterman2014markov}, and indeed strongly polynomial in $(\abs{\cS}, \abs{\cA}, 1/(1-\gamma))$ \cite{ye2011simplex, scherrer2013improved},  
to the exact optimal policy and the optimal value function.
In comparison, the results we established for the weighted policy convergence (Proposition \ref{prop_weighted_policy_convergence_linear}), local acceleration of the policy (Theorem \ref{thrm:convergence_to_optimal}) and optimality gap (Corollary \ref{corollary:value_convergence}), all clearly depend on the gap values of the underlying MDP instance $\cM$.

The contrast between the finite time, gap-free convergence of PI, and the gap-dependent policy convergence characterization of HPMD, brings forward the question on the necessity of gap values' role in previously established results. 
Below, we construct a class of MDPs confirming the necessity of this dependence. 
It is also important to note there that this result also applies to other popular PG methods, including the natural policy gradient method (NPG, \cite{kakade2001natural}).

\begin{theorem}\label{thrm_dependence_gap_tight}
There exists a class of MDPs $\cbr{\cM_{\epsilon}: \epsilon \in (0,1)}$,  where each $\cM_\epsilon$ has cost function bounded in $[-2, 2]$.  $\cM_\epsilon$ and $\cM_{\epsilon'}$ differ only in their cost functions, while sharing the same unique deterministic optimal  policy $\pi^*$.

For this class of MDPs, there exists a fixed state $s \in \cS$, and a sub-optimal action $a \notin \cA^*(s)$, such that for any
$\epsilon \in (0,1)$, 
we have $\delta_{\cZ, \cM}(s,a) = \delta_{\cS, \cM_\epsilon}(s) = \Delta^*(\cM_\epsilon) = \epsilon \gamma^2 /2$. 
Moreover, running HPMD  starting from the uniform policy gives 
\begin{align*}
\pi_{k+1}(a|s) > \pi_{k}(a|s), ~
~ \forall 
0 \leq k \leq \overline{k} \coloneqq  \cbr{ 0, \log_{1/\gamma} \rbr{\log \rbr{ \tfrac{ 3 \gamma^2  }{4 \delta_{\cS, \cM_\epsilon}(s)}} (1-\gamma^3) } /2} .
\end{align*}
In other words, 
$\cbr{\pi_k(\cdot|s)}$ is moving away from $\pi^*(\cdot|s)$ for $k \leq \overline{k}$, where $\overline{k}$ inversely depends on the gap values.
\end{theorem}

\begin{figure}[t!]
\centering 
    \includegraphics[width=0.45\textwidth]{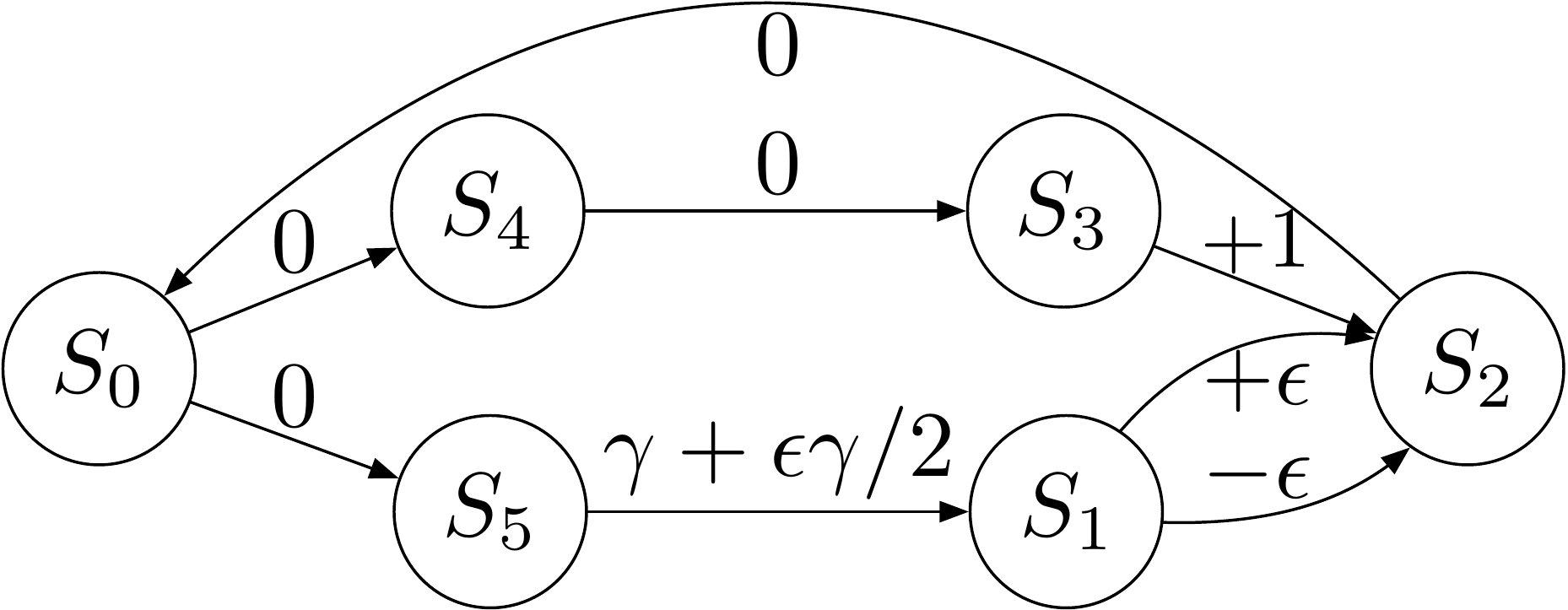}
  \caption{ 
 An MDP class for which HPMD and NPG have tight dependence on the gap values.
   }
  \label{fig_mdp_dependence}
\end{figure}

\begin{proof}

Consider the MDP with 6 states and 2 actions,  illustrated in Figure \ref{fig_mdp_dependence}, with deterministic transition. Each arc denotes the action with the associated cost on its edge. 
Only $S_0$ and $S_1$ have nontrivial action sets, consisting of choosing the upper arc (shorthanded with $\cU$) and the lower arc (shorthanded by $\cD$).
For other states, $\cU$ and $\cD$ correspond to the same arc.
It is clear that the optimal policy satisfies $\pi^*(D | S_0)  = \pi^*(D|S_1 ) = 1$. 
In addition, $\delta_{\cZ, \cM_\epsilon}(S_0, U) = \delta_{\cS, \cM_\epsilon}(S_0) = \epsilon \gamma^2 /2 $, $\delta_{\cS, \cM_\epsilon} (S_1) = 2 \epsilon$, and  $\Delta^*(\cM_\epsilon) = \epsilon \gamma^2 /2 $.

Now suppose we start HPMD at the uniform policy.
Since at state $S_1$, we transit to state $S_2$ regardless of the action, hence
$Q^t(S_1, \cU)  - Q^t(S_1, \cD) = 2 \epsilon$ for any $t \geq 0$. 
Consequently, applying \eqref{difference_logit}, we obtain 
\begin{align*}
\pi_k(\cU|S_1) / \pi_k(\cD | S_1) 
> \exp \rbr{ -2  \gamma^{-2k} \epsilon / (1 - \gamma^3 ) } ,
\end{align*}
which in turn implies that for any $k \leq \overline{k} $, it holds 
$\pi_k(\cD|S_1) - \pi_k(\cU|S_1) < 1/2$. 
In this case, it is immediate to see that  $Q^k(S_0, \cU)  - Q^k(S_0, \cD) < 0$.
Applying \eqref{difference_logit} again, we conclude that
${\pi_{k+1}(\cU|S_0) }/{ \pi_{k+1}(\cD | S_0)}  > {\pi_{k} (\cU|S_0) }/{ \pi_{k}(\cD | S_0)}  $ for any $k \leq \overline{k}$.
The proof is then completed.
\qed \end{proof}


A few remarks are in order for interpreting Theorem \ref{thrm_dependence_gap_tight}. 
First, the proof shows that the policy convergence is non-monotone. In particular, at state $S_0$ the probability for optimal action $\cD$ decreases for at least $\overline{k}$ iterations.  
Second, the duration of the phase in which the policy moves away from $\pi^*$ is completely controlled by the gap value of the MDP instance $\cM$, with smaller gap leading to longer duration.
This observation thus verifies the necessity on the dependence of gap values in the weighted norm policy convergence, and the local acceleration for both the policy and optimality gap. 
The observed dependence also demonstrates a major difference between the HPMD (and NPG) and PI. 
Finally, note that $\pi_0$ is indeed $(2\epsilon)$-optimal, yet learning the optimal policy takes infinite iterations as $\epsilon$ approaches 0,
which shows the different level of difficulties in value minimization and best policy identification for HPMD (and NPG).

\subsection{Last-iterate Convergence of the Policy}\label{subsec_last_iterate}
In this subsection, we proceed to show that the limiting policy value lower bound for every $a \in \cA^*(s)$, illustrated in Figure \ref{subfig:min_policy}, is a general property governed by the last-iterate policy convergence.
Note that as an immediate consequence of Lemma \ref{lemma:optimal_policy_set}, the optimal policy with the maximal entropy for every state, denoted by $\pi^*_U$,  
takes the form of 
\begin{align}
 \pi_U^* (a|s) \coloneqq 
\begin{cases}
1/ \abs{\cA^*(s)}, ~ &a \in \cA^*(s), \\
0, ~ & a \notin \cA^*(s).
\end{cases}
\label{eq_def_limit_policy}
\end{align}

Our main  result in this subsection shows that HPMD converges exactly to this particular optimal policy. 
Before we proceed, we first introduce a technical lemma that would prove useful in our ensuing discussions.

%
%

\begin{lemma}\label{lemma_prob_ub_and_lb}
Fix state $s \in \cS$, suppose for some $\cA^*(s) \subseteq \cA$,  $\epsilon > 0$ and $\rho \geq 1$, 
we have 
$
\tsum_{a \in \cA^*(s)} \pi(a|s) \geq 1- \epsilon$, and   
$\frac{\pi(i|s)}{\pi(j|s)} \leq \rho$, for any  $i, j \in \cA^*(s)$.
Then if $\abs{\cA^*(s)} > 1$.
\begin{align}\label{lemma_eq_prob_ub_and_lb}
\frac{1-\epsilon}{\abs{\cA^*(s)}} - (\rho - 1)
\leq \pi(i|s) 
\leq 
\frac{1}{\abs{\cA^*(s)}} + \rho - 1, ~ \forall i \in \cA^*(s).
\end{align}
In particular, if  $\rho$ satisfies 
$\rho \leq 1+ \rbr{1- 1/ \abs{\cA^*(s)}} \epsilon$, then 
\begin{align*}
\abr{ \pi(i|s) - 1/ \abs{\cA^*(s)} } \leq  \epsilon   ,~ \forall i \in \cA^*(s).
\end{align*}
On the other hand, if $\abs{\cA^*(s)} =1$, then the above inequality holds for any $\rho \geq 1$.
\end{lemma}

\begin{proof}
Given ${\pi(i|s)}/{\pi(j|s)} \leq \rho$ for any  $i, j \in \cA^*(s)$, and $\pi_i, \pi_j \in [0,1]$, 
it is immediate that 
$\abs{\pi_i - \pi_j} \leq \rho -1$ for any $i, j \in \cA^*(s)$.
Combining this observation with $\tsum_{a \in \cA^*(s)} \pi(a|s) \geq 1- \epsilon$, 
we then obtain $\min_{i \in \cA^*(s)} \pi_i \geq (1-\epsilon) /\abs{\cA^*(s)} - (\rho - 1)$.
Combining the same observation with 
 $\tsum_{a \in \cA^*(s)} \pi(a|s) \leq 1 $, 
 we obtain $\max_{i \in \cA^*(s)} \pi_i \leq 1/ \abs{\cA^*(s)}+ (\rho - 1)$.
 Thus \eqref{lemma_eq_prob_ub_and_lb} is proved.
 The rest of the claim follows from direct calculations.
\qed \end{proof}

With prior discussions in place, we are now ready to establish the last-iterate convergence of the policy and characterize the limiting policy.
\begin{theorem}[Last-iterate Policy Convergence]\label{thrm:prob_lb}
With $1+ \eta_k \tau_k = 1/\gamma$ and $\eta_k = \gamma^{-2(k+1)}$,
there exists  an optimal policy $\pi^* \in \Pi^*$, such that 
$
\lim_{k \to \infty} \pi_k = \pi^*.
$
Furthermore, the limiting policy corresponds to the optimal policy with the maximal entropy for every state,
\begin{align}\label{limit_policy}
\lim_{k \to \infty} \pi_k   (a|s) = \pi_U^*  (a|s) \coloneqq 
\begin{cases}
1/ \abs{\cA^*(s)}, ~ &a \in \cA^*(s), \\
0, ~ & a \notin \cA^*(s).
\end{cases}
\end{align}
In particular, for any $0 < \epsilon < 1$, we have  $\norm{\pi_{k+1} - \pi^*_U}_\infty \leq \epsilon$ whenever  
\begin{align}\label{kl_policy_convergence_finite_precision}
 k \geq \frac{1}{2} \log_\gamma \rbr{\frac{\Delta^*(\cM)}{2 \gamma \log (C_\gamma \abs{\cA} / \epsilon) }}
+ 2 \overline{K}_1
+ \log_\gamma \rbr{\frac{D}{2A}} 
+ 2 \log_\gamma \rbr{\frac{D}{2B}},
\end{align}
 where 
$
\overline{K}_1 = \min_{k} \cbr{k > K_1:  \Delta^*(\cM) \gamma^{-2k - 1} \geq 5k \log(\frac{1}{\gamma})},
$
 $K_1$ is defined as in Theorem \ref{thrm:convergence_to_optimal},
and  $A = \frac{2\varrho (4\log \abs{\cA} + C)}{(1-\gamma) (1-\gamma^2) \gamma} $,
$B =  \frac{4 \gamma C \abs{\cA} C_\gamma}{(1-\gamma^{1/2})(1-\gamma)\gamma}$,
and $D = 1 + \epsilon / 2$.
\end{theorem}

\begin{proof}
Note that as we have shown in the proof of Theorem \ref{thrm:convergence_to_optimal},  any action $a \notin \cA^*(s)$ satisfies $\pi_k(a|s) \to 0$. 
Thus it remains to discuss the convergence of $\pi_k(a|s)$ for every $a \in \cA^*(s)$.

Now let us consider any $i, j \in \cA^*(s)$, where it must holds that $Q^*(s, i) = Q^*(s,j)$. 
Recall that \eqref{difference_logit} still holds, that is, 
$
z_i^{k+1} - z_j^{k+1}  = - \tsum_{t=0}^{k} \gamma^{k+1 - t} \eta_t ( Q_i^t - Q_j^t).
$
Our goal is to show that $z_i^{k+1} - z_j^{k+1} \to 0$.

From Corollary \ref{corollary:value_convergence} and the definition of $Q$-function, 
we have that for any $(s,a) \in \cS \times \cA$, 
\begin{align}
Q^{\pi_k}(s,a) - Q^*(s,a) & = \gamma \tsum_{s' \in \cS} \cP(s' |s,a) \rbr{V^{\pi_k}(s) - V^*(s) }
\leq \frac{2 \gamma C \abs{\cA} C_\gamma}{(1-\gamma)^2} 
\exp \rbr{- \frac{{\Delta^*(\cM)} \gamma^{-2k - 1}}{2}  }, \label{convergence_q_super}
\end{align}
whenever $k \geq K_1$, where $K_1$ is defined as in Theorem \ref{thrm:convergence_to_optimal}.
Thus for any $\overline{K}_1 \geq K_1$, we have 
\begin{align}
 \tsum_{t=0}^{k} \gamma^{k+1 - t} \eta_t \abs{Q_i^t - Q_j^t}
& \leq   \underbrace{\tsum_{t=0}^{\overline{K}_1} \gamma^{k+1 - t} \eta_t \abs{ Q_i^t - Q_j^t}}_{(A)} +  \underbrace{\tsum_{t=\overline{K}_1+1}^{k} \gamma^{k+1 - t} \eta_t \abs{ Q_i^t - Q_j^t}}_{(B)}. \label{ineq:decomposition}
\end{align}
We proceed to first bound term (A) in \eqref{ineq:decomposition}.
 From \eqref{q_convergence} and the definition of $\eta_t = \gamma^{-2(t+1)}$, 
\begin{align*}
\gamma^{k+1 - t} \eta_t \abs{Q_i^t - Q_j^t} & \leq  \gamma^{k+1 - t} \eta_t \rbr{\abs{Q_i^t - Q^*(s,i)} + \abs{Q_j^t - Q^*(s, j)}  } \\
& \leq 2 \gamma^{k - 3t -1}
\cdot\varrho\gamma^t
  \frac{4  \log \abs{\cA} + C}{1-\gamma}
 =  \gamma^{k-2t} \frac{2 \varrho (4 \log \abs{\cA} + C)}{(1-\gamma)\gamma}.
\end{align*}
Hence we can bound term (A) as 
\begin{align}\label{bound:A}
(A) = \tsum_{t = 0}^{\overline{K}_1 } \gamma^{k+1 - t} \eta_t \abs{Q_i^t - Q_j^t} 
\leq \tsum_{t=0}^{\overline{K}_1}  \gamma^{k-2t} \frac{2 \varrho (4 \log \abs{\cA} + C)}{(1-\gamma)\gamma}
\leq \gamma^{k-2\overline{K}_1 } \frac{2 \varrho (4\log \abs{\cA} + C)}{(1-\gamma) (1-\gamma^2) \gamma}.
\end{align}
We then proceed to bound term (B) in \eqref{ineq:decomposition}.
From the definition of $\eta_t = \gamma^{-2(t+1)}$, the superlinear convergence of Q-function in \eqref{convergence_q_super}, and the condition that 
$Q^*(s,i) = Q^*(s,j)$, 
\begin{align}
\tsum_{t = \overline{K}_1 }^k \gamma^{k + 1 -t} \eta_t \abs{Q_i^t - Q_j^t} 
& \leq \tsum_{t = \overline{K}_1 }^k \gamma^{k + 1 -t} \eta_t \rbr{ \abs{Q_i^t - Q^*(s,i)} + \abs{Q_j^t - Q^*(s,j)} } \nonumber 
\\
&  \leq  \tsum_{t = \overline{K}_1 }^k \gamma^{k -3 t}  \frac{4 \gamma C \abs{\cA} C_\gamma}{(1-\gamma)^2 \gamma} 
\exp \rbr{- \frac{{\Delta^*(\cM)} \gamma^{-2k - 1}}{2} } \nonumber \\ 
& = \tsum_{t = \overline{K}_1 }^k \gamma^{k - t/2} \frac{4 \gamma C \abs{\cA} C_\gamma}{(1-\gamma)^2 \gamma} 
\exp \rbr{ - \frac{\Delta^*(\cM)}{2} \gamma^{-2t - 1} + \frac{5t}{2} \log(\frac{1}{\gamma}) } .
\nonumber
\end{align}
Now consider choosing
$
\overline{K}_1 = \min_{k} \cbr{k > K_1:  \Delta^*(\cM) \gamma^{-2k - 1} \geq 5k \log(\frac{1}{\gamma})},
$
then for any $k \geq \overline{K}_1$,
\begin{align}
(B)  \leq \tsum_{t = \overline{K}_1 }^k \gamma^{k + 1 -t} \eta_t \abs{Q_i^t - Q_j^t} 
    \leq  \tsum_{t = \overline{K}_1 }^k \gamma^{k - t/2} \frac{4 \gamma C \abs{\cA} C_\gamma}{(1-\gamma)\gamma} 
\leq \gamma^{k/2} \frac{4 \gamma C \abs{\cA} C_\gamma}{(1-\gamma^{1/2})(1-\gamma)\gamma} .
\label{bound:B}
\end{align}
Thus, by combing \eqref{bound:A} and \eqref{bound:B},  for any $k \geq \overline{K}_1$,
\begin{align}\label{finite_time_logit_diff_same}
 \abs{z_i^{k+1} - z_j^{k+1} } \leq \tsum_{t=0}^{k} \gamma^{k+1 - t} \eta_t \abs{Q_i^t - Q_j^t}
& \leq  \gamma^{k-2\overline{K}_1 } \frac{2\varrho (4\log \abs{\cA} + C)}{(1-\gamma) (1-\gamma^2) \gamma} 
+\gamma^{k/2} \frac{4 \gamma C \abs{\cA} C_\gamma}{(1-\gamma^{1/2})(1-\gamma)\gamma}.
\end{align}
Fixing $\overline{K}_1$, and taking $k \to \infty$ in the previous relation, we obtain  
\begin{align}\label{asymp_logit_diff}
\lim_{k \to \infty}  \tsum_{t=0}^{k} \gamma^{k+1 - t} \eta_t \abs{Q_i^t - Q_j^t} \to 0~~
\Rightarrow ~~ \abs{z_i^{k+1} - z_j^{k+1}} \to 0.
\end{align}
 
Since $\cbr{\pi_k}  \subset \Pi$ and $\Pi$ is compact,  $\cbr{\pi_k}  $ has at least one limit point.
Let  $\pi^*$ denote a limit point of the policy iterate $\cbr{\pi_k} $, i.e., there exists subsequence 
$\cbr{\pi_{k_l}}$ such that $\lim_{l \to \infty} \pi_{k_l} = \pi^*$.
Denoting $z_i^* = \log \pi^*(i|s)$ for any $i \in \cA$. 
Then from  \eqref{asymp_logit_diff} and the definition of $(z_i^k, z_j^k)$, we have 
\begin{align*}
\frac{\pi^*(i|s)}{\pi^*(j|s)} = \exp \rbr{\log z_i^* - \log z_j^* }
= \lim_{l \to \infty} \exp \rbr{ \log z_i^{k_l} - \log z_j^{k_l} }
= 1, ~ \forall i,j \in \cA^*(s).
\end{align*}
Combining the above observation with \eqref{state_wise_support_recovery}, we conclude that every limit point $\pi^*$ is given by $\pi^*_U$, and thus \eqref{limit_policy} follows immediately.

To establish \eqref{kl_policy_convergence_finite_precision}, first we observe  that in view of  \eqref{state_wise_support_recovery}, for any $k \geq \frac{1}{2} \log_\gamma \rbr{\frac{\Delta^*(\cM)}{2 \gamma \log (C_\gamma \abs{\cA} / \epsilon) }}$,  
\begin{align*}
\pi_{k+1}(j|s) \leq  \epsilon / \abs{\cA}, ~\forall s \in \cS, j \notin \cA^*(s).
\end{align*}
This in turn implies 
$
\tsum_{a \in \cA^*(s)} \pi_{k+1}(a|s) \geq 1 - \epsilon.
$
Note that \eqref{finite_time_logit_diff_same}, together with the definition of $z_i^k$, implies 
\begin{align*}
\frac{\pi_{k+1}(i | s) }{ \pi_{k+1}(j|s)} \leq \rho_{k+1}
= 
\exp \rbr{
 \gamma^{k-2\overline{K}_1 } \frac{2\varrho (4\log \abs{\cA} + C)}{(1-\gamma) (1-\gamma^2) \gamma} 
+\gamma^{k/2} \frac{4 \gamma C \abs{\cA} C_\gamma}{(1-\gamma^{1/2})(1-\gamma)\gamma}
},
~ \forall i, j \in \cA^*(s).
\end{align*}
Combining above two observations, we can invoke Lemma \ref{lemma_prob_ub_and_lb}, and obtain that 
whenever 
\begin{align}\label{condition_rho_eps_for_policy_convergence_kl}
\rho_{k+1} \leq  1 + \epsilon / 2,
\end{align}
 it holds
$
\norm{\pi_{k+1}(\cdot|s) - \pi^*_U(\cdot|s)}_\infty \leq \epsilon.
$
The desired claim then follows immediately, by noting that condition \eqref{condition_rho_eps_for_policy_convergence_kl} can be satisfied with 
\begin{align*}
k \geq \frac{1}{2} \log_\gamma \rbr{\frac{\Delta^*(\cM)}{2 \gamma \log (C_\gamma \abs{\cA} / \epsilon) }}
+ 2 \overline{K}_1
+ \log_\gamma \rbr{\frac{D}{2A}} 
+ 2 \log_\gamma \rbr{\frac{D}{2B}},
\end{align*}
where $A = \frac{2\varrho (4\log \abs{\cA} + C)}{(1-\gamma) (1-\gamma^2) \gamma} $,
$B =  \frac{4 \gamma C \abs{\cA} C_\gamma}{(1-\gamma^{1/2})(1-\gamma)\gamma}$,
and $D = 1 + \epsilon / 2$.
\qed \end{proof}

 To the best of our knowledge,  Theorem \ref{thrm:prob_lb} is the first result establishing the last-iterate  convergence of the policy, among existing first-order methods in the RL literature.  
We term the phenomenon of converging to the maximal-entropy optimal policy the  implicit (algorithmic) regularization of HPMD, as  there is no explicit regularization involved in the objective \eqref{eq:mdp_single_obj} to promote seeking maximal-entropy optimal policy.

 On a related note, it has been discussed in  \cite{hu2021actor} that the actor-critic method produces policy with bounded Kullback-Leibler divergence to the maximal-entropy optimal policy.
In contrast, we establish the exact convergence to this optimal policy, 
instead of bounding  the KL divergence.
See also \cite{derman2021twice} for discussions between entropy regularizations and the robustness of learned policies.

{\it Connections to Homotopy Methods.} To provide more context in interpreting Theorem \ref{thrm:prob_lb}, it is worth pointing out some interesting connections between HPMD and the homotopy method (i.e. regularization path) in the statistics literature \cite{zhao2007stagewise, hastie2004entire, park2007l1}.  

Consider an empirical risk minimization problem of the form 
$
\min_{\theta \in \RR^d } \cL (\theta) ,
$
where $\cL: \RR^d \to \RR$ denotes the empirical risk on the training data $\cbr{(x_i, y_i)}_{i =1}^n$.
 The regularization path $\cbr{\Gamma(s): s \geq 0}$ is a path in $\RR^d$, with $\Gamma(s)$ being a minimizer of  
 the regularized problem 
 $\min_{\theta \in \RR^d } \cL (\theta) + s \cR(\theta)$. 
 Here $\cR: \RR^d \to \RR$ denotes the regularization term to control the complexity, and $s  > 0$ denotes the regularization strength.
 
The algorithmic regularization of the homotopic method was first discussed in  \cite{rosset2004boosting}, which shows that when training linear classifiers with linearly separable data and an exponentially-tailed $\cL(\cdot)$, and $\cR(\cdot)$ being  $\ell_p$-norm, then taking $s \to 0$, the regularization path $\Gamma(s)$ converges in direction to the SVM solution:  
\begin{align}\label{svm_constraint}
 \theta^*  \leftarrow \min_{\theta} \norm{\theta}_p, ~\text{s.t.}~
\tsum_{i = 1}^n \max \rbr{ 1- y x_i^\top \theta, 0} \leq \tsum_{i = 1}^n \max \rbr{ 1- y x_i^\top \theta', 0} , ~ \forall \theta' \in \RR^d.
\end{align}

Similar to the regularization path $\cbr{\Gamma(s)}_{s \geq 0}$, HPMD can be viewed as solving a sequence of regularized MDP problems with diminishing regularizations. In particular, define the regularized weighted value as 
\begin{align*}
\textstyle
\min_{\pi} \cbr{f_{\tau}(\pi) \coloneqq \EE_{\nu^*} \sbr{V_{\tau}^{\pi}(s)}}, ~~ \mathrm{s.t.} ~~ \pi(\cdot|s) \in \Delta_{\cA}, \forall s \in \cS,
\end{align*}
where
$
V^{\pi}_{\tau} (s) 
 = \EE \sbr{\tsum_{t=0}^\infty \gamma^t \rbr{ c(s_t, a_t) - \tau \cH(\pi(\cdot|s_t)) } \big| s_0 = s, a_t \sim \pi(\cdot|s_t), s_{t+1} \sim \cP(\cdot|s_t,a_t)  }.
$
HPMD can be viewed as solving each regularized objective $f_{\tau_n}$  approximately using only  one step of policy gradient.
As Theorem \ref{thrm:prob_lb} suggests, the HPMD method yields the policy $\pi^*_U$, which  is also the solution of problem
\begin{align}\label{mdp_constraint_reg}
\pi^*_U \leftarrow  \min_{\pi \in \Pi} \tsum_{s \in \cS} - \cH(\pi(\cdot|s)), ~\text{s.t.}~ \tsum_{s \in \cS} V^{\pi}(s) \leq \tsum_{s \in \cS}  V^{\pi'}(s),  ~ \forall \pi' \in \Pi.
\end{align}
 
By comparing  \eqref{svm_constraint} and \eqref{mdp_constraint_reg}, 
it should be clear that the algorithmic regularization of HPMD parallels that of the homotopy method, in the sense that both seek the minimal-complexity solution (measured by $-\cH(\cdot)$ and $\ell_p$-norm, respectively) among all candidates that optimizes certain loss of interest.
Moreover, it is worth mentioning  that algorithmic regularization effects of first-order methods have also been extensively studied in the supervised learning literature \cite{ji2019implicit, soudry2018implicit, gunasekar2018characterizing, Li2020Implicit, li2021implicit}.
 In contrast,  Theorem \ref{thrm:prob_lb} seems to be the first algorithmic regularization result identified among first-order methods in the RL literature.

\subsection{Results for Sublinearly Converging HPMD}\label{subsec_results_sublinear_hpmd} 

We have so far focused on establishing the local acceleration and last-iterate policy convergence of the linearly converging HPMD variant. 
In this subsection, we proceed to show that these two observations hold fairly general as the computational properties of HPMD. 
Specifically, both local acceleration and the last-iterate policy convergence apply to the sublinearly converging HPMD variant, studied in Theorem \ref{thrm_sublinear_convergence}.

\begin{theorem}[Local Accelerated Convergence]\label{thrm_local_convergence_hpmd_sublinear}
With $\eta_k = k + k_0$, $\tau_k = {1}/{(k+k_0)^2}$, $k_0 = \ceil{{\gamma}/\rbr{1-\gamma}}$,~then
\begin{align}
\mathrm{dist}_{\ell_{1}} (\pi_{k+1} ,\Pi^*)  \leq   2 C_\gamma \abs{\cA}
\exp \rbr{ - {\Delta^*(\cM) k^2}/ 16 },  \label{eq_local_convergence_policy_hpmd_sublinear}
\end{align} 
for any iteration $k \geq K_1 \coloneqq (\underline{K}_1 + 2 k_0)^3$,
where $ \underline{K}_1 
= \frac{32 \varrho \log \abs{\cA}}{\Delta^*(\cM) (1-\gamma)}  \log \rbr{  \frac{32 \varrho \log \abs{\cA}}{\Delta^*(\cM) (1-\gamma)} }$,
 and $C_\gamma =  \exp \rbr{ \frac{C}{3 ( 1-\gamma ) } }$.
 Consequently, we also obtain 
 \begin{align}\label{eq_local_convergence_opt_gap_hpmd_sublinear}
 V^{\pi_{k+1}}(s) - V^{*}(s) \leq \frac{ 2 C \abs{\cA}  C_\gamma}{(1-\gamma)^2} \exp \rbr{ - \frac{\Delta^*(\cM) k^2}{16} }, ~\forall s\in \cS, a \in \cA.
 \end{align}
\end{theorem}

\begin{proof}
Let us adopt the same notations as in the proof of Theorem \ref{thrm:convergence_to_optimal}.
Fix state $s \in \cS$,  from \eqref{update_recursion_crude}, we have 
\begin{align*}
\log \pi_{k+1}(a|s) = \upsilon_k  \rbr{\log \pi_k(a|s) - \eta_k Q^{\pi_k}(s,a) }+ \zeta_k(s), ~
\upsilon_k \coloneqq 1 / \rbr{1+ \eta_k \tau_k}.
\end{align*} 
Recursively applying the above relation and using that fact $\pi_0$ is the uniform policy, we obtain 
\begin{align}\label{logit_recursion_sublinear}
z_i^{k+1} - z_j^{k+1} 
= - \tsum_{t=0}^k \eta_t  \beta_t (Q_i^t - Q_j^t) , ~ \beta_t \coloneqq \prod_{s = t}^k \upsilon_t.
\end{align}

For any $i \in \cA^*(s), ~ j \notin \cA^*(s)$,  similar to \eqref{diff_q_convergence}, by applying Theorem \ref{thrm_sublinear_convergence}, we obtain that 
if $ t \geq \underline{K}_1 
\coloneqq \frac{32 \varrho \log \abs{\cA}}{\Delta^*(\cM) (1-\gamma)}  \log \rbr{  \frac{32 \varrho \log \abs{\cA}}{\Delta^*(\cM) (1-\gamma)} }$, 
then 
$
Q_i^t - Q_j^t \leq - \Delta^*(\cM) /2 < 0 .
$
Combining this relation with \eqref{logit_recursion_sublinear},  for any $k \geq K_1 \coloneqq (\underline{K}_1 + 2 k_0)^3$, 
\begin{align}
z_i^{k+1} - z_j^{k+1} 
& \geq -\tsum_{t=0}^{\underline{K}_1} \frac{\eta_t \beta_t C}{1-\gamma} 
+ \tsum_{t = \underline{K}_1 + 1}^k \frac{\eta_t \beta_t \Delta^*(\cM)}{2}  \nonumber \\
& \overset{(a)}{=} - \frac{C}{1-\gamma} \tsum_{t=k_0 }^{\underline{K}_1 + k_0} \frac{t^2}{ k+1}
+ \tsum_{\underline{K}_1 + k_0 + 1}^{k + k_0} \frac{\Delta^*(\cM) t^2}{2 (k+1) } \nonumber  \\
& \overset{(b)}{\geq} 
- \frac{C}{3(1-\gamma)} + \frac{\Delta^*(\cM) k^2}{16} , \label{ineq_sublinear_hpmd_logit_diff_lb}
\end{align}
where $(a)$ follows from the choice of $\cbr{(\eta_t, \tau_t)}$, which simultaneously implying $\beta_t = \frac{t + k_0}{k + k_0 + 1}$;
$(b)$ follows from direct calculations and the choice of $k \geq (\underline{K}_1 + 2 k_0)^3$. 
By combining \eqref{ineq_sublinear_hpmd_logit_diff_lb} with the same arguments for  establishing \eqref{state_wise_support_recovery}, we obtain
\begin{align}
\pi_{k+1}(j | s)  \leq  C_\gamma
\exp \rbr{- {\Delta^*(\cM) k^2}/ 16 },
~ \forall j \notin \cA^*(s), 
 \label{state_wise_support_recovery_sublinear_hpmd}
\end{align}
and consequently 
\eqref{eq_local_convergence_policy_hpmd_sublinear}.
Finally, \eqref{eq_local_convergence_opt_gap_hpmd_sublinear} follows the same proof as in Corollary \ref{corollary:value_convergence}, but using \eqref{eq_local_convergence_policy_hpmd_sublinear} instead of \eqref{eq_local_policy_convegence_hpmd_linear}.
\qed \end{proof}

Similar to Theorem \ref{thrm:prob_lb}, sublinearly converging HPMD also exhibits the last-iterate policy convergence, with the same limiting policy.
\begin{theorem}[Last-iterate Policy Convergence]
With $\eta_k = k + k_0$, $\tau_k = {1}/{(k+k_0)^2}$, where $k_0 = \ceil{{\gamma}/\rbr{1-\gamma}}$,
then $
\lim_{k \to \infty} \pi_k = \pi^*_U
$.
In particular, for any $0 < \epsilon < 1$, we have  $\norm{\pi_{k+1} - \pi^*_U}_\infty \leq \epsilon$ whenever  
\begin{align}\label{kl_policy_convergence_finite_precision_sublinear_hpmd}
 k \geq \overline{K}_1 +
4 \sqrt{ \frac{\log\rbr{{C_\gamma \abs{\cA}}/\rbr{\epsilon}}}{\Delta^*(\cM)}}
+
  \frac{4 C (\overline{K}_1 + 2 k_0)^3}{(1-\gamma) \epsilon} 
+ \frac{8 \varrho C \abs{\cA} C_\gamma}{(1-\gamma)^2 \epsilon} ,
\end{align}
 where 
$
\overline{K}_1 = \min \cbr{k \geq K_1: \Delta^*(\cM) k^2 \geq 64 \log k},
$
and
 $K_1$ is defined as in Theorem \ref{thrm_local_convergence_hpmd_sublinear}.
\end{theorem}

\begin{proof}
Let us adopt the same notations as in the proof of Theorem \ref{thrm_local_convergence_hpmd_sublinear}.
For any $i, j \in \cA^*(s)$, 
 following similar lines as in the proof of Theorem \ref{thrm:prob_lb}, 
 for any $k > \overline{K}_1$, 
\begin{align*}
\abr{z_i^{k+1} - z_j^{k+1}} 
& \leq \tsum_{t = 0}^{\overline{K}_1} \frac{\eta_t \beta_t C}{1-\gamma} 
+ \tsum_{\overline{K}_1 + 1}^k \frac{2 \varrho C \abs{\cA} C_\gamma}{(1-\gamma)^2} \exp \rbr{ - \frac{\Delta^*(\cM) t^2}{16}} \eta_t \beta_t \\
& \overset{(a)}{\leq} 
\frac{C ( \overline{K}_1 + k_0 + 1)^3}{3( k+1) (1-\gamma)}
+ \frac{2 \varrho C \abs{\cA} C_\gamma}{(1-\gamma)^2 k} 
\tsum_{t = \overline{K}_1 + k_0 + 1}^{k + k_0}  t^{-2} \cdot
\exp \rbr{
- \frac{\Delta^*(\cM) t^2}{16} + 4 \log t
} \\
& \leq 
\frac{C ( \overline{K}_1 + 2 k_0)^3}{3k (1-\gamma)}
+ \frac{2 \varrho C \abs{\cA} C_\gamma}{(1-\gamma)^2 k},
\end{align*}
where $(a)$ follows from the choice of $\cbr{(\eta_t, \tau_t)}$.
The above relation in turn implies 
\begin{align}
\frac{\pi_{k+1}(i | s) }{ \pi_{k+1}(j|s)} \leq \rho_{k+1}
= 
\exp \rbr{
\frac{C ( \overline{K}_1 + 2 k_0)^3}{3k (1-\gamma)}
+ \frac{2 \varrho C \abs{\cA} C_\gamma}{(1-\gamma)^2 k}
},
~ \forall i, j \in \cA^*(s).
\label{logit_ratio_opt_sublinear_hpmd}
\end{align}
Now in view of \eqref{state_wise_support_recovery_sublinear_hpmd}, for any $k \geq K_1 + 4 \sqrt{\log\rbr{{C_\gamma \abs{\cA}}/\rbr{\epsilon}}/\Delta^*(\cM)}$, it holds that for any $s \in \cS$,
\begin{align}
\pi_{k+1}(j|s) \leq  \epsilon / \abs{\cA},  ~\forall  j \notin \cA^*(s) 
\Rightarrow 
\tsum_{a \in \cA^*(s)} \pi_{k+1}(a|s) \geq 1 - \epsilon.
\label{logit_sum_nonopt_sublinear_hpmd}
\end{align}

By combining \eqref{logit_ratio_opt_sublinear_hpmd}, \eqref{logit_sum_nonopt_sublinear_hpmd}, and Lemma \ref{lemma_prob_ub_and_lb}, 
to satisfy $\norm{\pi_{k+1} - \pi^*_U}_\infty \leq \epsilon$, 
it suffices to let $\rho_{k+1} \leq 1 + \epsilon / 2$.
This holds whenever \eqref{kl_policy_convergence_finite_precision_sublinear_hpmd} is satisfied. The proof is then completed.
\qed \end{proof}

\subsection{On the Removal of Assumption \ref{assump_support}}\label{subsec_discuss_assump}
In this subsection, we  discuss the feasibility of removing Assumption \ref{assump_support}. 
Consequently, we show that
Assumption \ref{assump_support} serves only the purpose of presentation simplicity, and 
all the results in this manuscript hold in an assumption-free fashion.

As should be clear from \eqref{mismatch_ratio_in_weighted_policy},  \eqref{q_convergence} and \eqref{convergence_q_super}, Assumption \ref{assump_support} is only needed to certify  pointwise linear convergence of  $V^{\pi_k}(\cdot)$ to $V^*(\cdot)$.
Given this observation, it is then clear that if one can establish the linear convergence of  
the general weighted objective $f_\rho(\pi)$ defined in \eqref{eq:mdp_single_obj_raw}, 
for distribution $\rho$ with $\mathrm{supp}(\rho) = \cS$,
then Assumption \ref{assump_support} can be removed. 

Clearly, $f_{\rho}(\pi)$ reduces to objective $f(\pi)$ defined in \eqref{eq:mdp_single_obj} by taking $\rho = \nu^*$.
We now show that establishing the linear convergence for $f_{\rho}(\pi)$ only takes a slightly modified analysis compared to that of $f(\pi)$.

Specifically,  instead of  taking    expectation with respect to $s \sim \nu^*$ on both sides of \eqref{three_point_decomposition},  we now take expectation with respect to $s \sim d_{\rho}^{\pi^*}$.
Then by reusing other elements in the proof of Lemma \ref{deterministic_generic}, one can obtain the following counterpart of \eqref{ineq_recursion_before_spec_parameters}, 
\begin{align}\label{opt_recursion_raw_for_general_weight}
f_\rho(\pi_{k+1}) - f(\pi^*) + \rbr{\frac{1}{\eta_k} + \tau_k} \phi'(\pi_{k+1}, \pi^*) 
\leq \gamma' \rbr{f_\rho(\pi_k) - f(\pi^*)}
+ \frac{1}{\eta_k} \phi'(\pi_k, \pi^*)
+ \frac{3\tau_k}{1-\gamma} \log \abs{\cA}, 
\end{align}
where 
\begin{align*}
\gamma' = 1 - \frac{1-\gamma}{\lVert  d_{\rho}^{\pi^*}/\rho \rVert_{\infty}},  ~ 
  \phi'(\pi, \pi^*) = \EE_{s \sim d_\rho^{\pi^*} } \frac{ D^{\pi^*}_{\pi}(s) }{\lVert  d_{\rho}^{\pi^*}/\rho \rVert_{\infty}}.
  \end{align*} 
  In view of \eqref{opt_recursion_raw_for_general_weight}, it is clear that Lemma \ref{deterministic_generic} holds for the weighted objective $f_\rho(\cdot)$, with $(\gamma', \phi')$ replacing the role of $(\gamma, \phi)$.
  Consequently, both Theorem \ref{thrm_sublinear_convergence} and \ref{thrm_convergence_fpmd} apply to the weighted objective $f_\rho(\cdot)$, with every term of $(1-\gamma)$ replaced by $(1-\gamma')$. 
The same argument can also be directly applied to the stochastic setting considered in Section \ref{sec:stochastic}.

\section{HPMD with Decomposable Bregman Divergences}\label{subsec_policy_convergence_decomposable}
In this section, we generalize the computational properties  of HPMD discussed in Section \ref{sec_deterministic} and \ref{sec_local_convergence_and_last_iterate}, which so far take the negative entropy function as the distance-generating function. 
Instead, we will consider a more general class of distance-generating functions that have decomposable structure.
Specifically, we assume the distance-generating function, denoted by $w: \Delta_{\cA}\to \overline{\RR} \coloneqq \RR \cup \cbr{\infty}$, takes the form of 
\begin{align}\label{decomposable_reg_main}
w(p)  = \tsum_{i=1}^{\abs{\cA}} v(p_i).
\end{align} 
We assume $v: \RR \to \overline{\RR} $ is a proper, closed and strictly convex function with $\mathrm{dom}(v) \supseteq \RR_+$ and is differentiable inside $\mathrm{Int}(\mathrm{dom}(v))$.
Our ensuing discussion also makes use of the restriction of $v$ onto $\RR_{+}$, defined as
\begin{align}\label{def_v_restriction}
\hat{v}(x) = v(x), ~ \forall x  \geq 0; ~~ \hat{v}(x) = \infty, ~ \forall x <0.
\end{align}
Accordingly, we define the restriction of $w$ as $\hat{w}(p) = \tsum_{i=1}^p \hat{v}(p_i)$, and a slightly generalized Bregman divergence associated with $\hat{w}$:
\begin{align}\label{def_generalized_bregman}
D^{\pi}_{\pi'}(s) =
\hat{w}(\pi(\cdot|s)) - \hat{w}(\pi'(\cdot|s)) - \inner{\nabla \hat{w}(\pi'(\cdot|s))}{\pi(\cdot|s) - \pi'(\cdot|s)},
\end{align}
Here $\nabla \hat{w}(p) \in \partial \hat{w}(p)$ denotes a subgradient of $w$ at $p$.

The generalized HPMD update rule takes the following form:
\begin{align}\label{hpmd_update_decomposable}
	\pi_{k+1} (\cdot| s) & =
	\argmin_{p(\cdot|s) \in \Delta_{\cA}} \eta_k\sbr{ \inner{Q^{\pi_k}(s, \cdot)}{ p(\cdot| s ) }  + \tau_k \hat{w} (p(\cdot|s)) } + D^{p}_{\pi_k}(s) \nonumber\\
	& = \argmin_{p(\cdot|s) \in \Delta_{\cA}} \eta_k\sbr{ \inner{Q^{\pi_k}(s, \cdot)}{ p(\cdot| s ) }  + \tau_k w (p(\cdot|s)) } + D^{p}_{\pi_k}(s),  ~ \forall s \in \cS,
\end{align}
where the equality is due to the constraint and the definition of $\hat{w}$.
The generalized Bregman divergence $D^p_{\pi_k}(s)$ is defined as in \eqref{def_generalized_bregman}, for some 
$\nabla \hat{w}(\pi_k(\cdot|s)) \in  \partial \hat{w}(\pi_k(\cdot|s)) $.
Compared to the KL-divergence considered before, working with the generalized Bregman divergence requires the additional attention to the fact that  $D^p_{\pi_k}(s)$ depends on the choice of $\nabla \hat{w}(\pi_k(\cdot|s))$, or equivalently, the choice of $\nabla \hat{v}(\pi_k(a|s))$.
Note that the generalized update recovers the update \eqref{fpmd:update} after taking $w(\cdot) = -\cH(\cdot)$.
The generalized update also does not require using the uniform policy as the initial policy.

Throughout our discussions in this section, we  choose the same $\cbr{\eta_k} $ and $\cbr{\tau_k} $ as in Theorem \ref{thrm_convergence_fpmd}.
We then proceed to establish that HPMD with the generalized update \eqref{hpmd_update_decomposable} converges linearly under certain technical conditions on $\hat{w}$.
As will be clear in our ensuing discussions, these conditions are fairly general and are satisfied by almost every practical choice of distance-generating function in the RL literature.
\begin{proposition}\label{claim_linear_decomposable_reg}
Suppose  the following holds:
\begin{itemize}
\item[(1)] $\sup_{\pi \in \Pi, s\in \cS} 2 \abs{ w(\pi(\cdot|s))} \leq \Phi < \infty $ for some $\Phi > 0$.
\item[(2)]  $\partial \hat{w} (\pi_k(\cdot|s))  \neq \emptyset$ for every $s \in \cS$ and $k \geq 0$.
\end{itemize}
Then by choosing the same $\cbr{\eta_k} $ and $\cbr{\tau_k} $ as in Theorem \ref{thrm_convergence_fpmd}, we have
 \begin{align*}
f(\pi_{k}) - f(\pi^*) 
\leq 
\gamma^k \rbr{
 f(\pi_{0}) - f(\pi^*)  + {4  \Phi}/\rbr{1-\gamma}
}.
\end{align*}
\end{proposition}
\begin{proof}
It can be straightforwardly verify that Lemma \ref{lemma_three_point} and Lemma \ref{deterministic_generic} hold for the generalized update \eqref{hpmd_update_decomposable}, with every term of the form $D^{\pi}_{\pi_0}(s)$ replaced by $w(\pi(\cdot|s))$. 
Hence by choosing $\Phi \geq \sup_{\pi \in \Pi, s \in \cS} 2 \abs{ w(\pi(\cdot|s))}$, and using the same choice of $\cbr{\eta_k} $ and $\cbr{\tau_k} $, we obtain, following the same lines in the proof of Theorem \ref{thrm_convergence_fpmd}, that 
\begin{align*}
f(\pi_{k}) - f(\pi^*) 
\leq 
\gamma^k \rbr{
 f(\pi_{0}) - f(\pi^*)  + {4  \Phi}/\rbr{1-\gamma}
}.
\end{align*}
The proof is then completed.
\qed \end{proof}

Below, we provide a simple and easily verifiable condition on the univariate function $v(\cdot)$ in the definition \eqref{decomposable_reg_main} of $w(\cdot)$, that in turn can guarantee a finite $\Phi$.

\begin{lemma}\label{lemma_bounded_divergence}
Given a proper closed convex function $v: \RR \to \overline{\RR}$ with $\mathrm{dom}(v) \supseteq [0,1]$,
then $-\infty < \inf_{x \in [0,1]} v(x) \leq \sup_{x \in [0,1]} v(x) < \infty$.
Consequently, we have $\sup_{\pi \in \Pi, s \in \cS} 2 \abs{ w(\pi(\cdot|s))} \leq \Phi <  \infty$ for some $\Phi >0$, where $w(\cdot)$ is defined as in \eqref{decomposable_reg_main}.
\end{lemma}

\begin{proof}
For any $x \in [0,1]$, from convexity we have 
$
v(x) \leq (1-x) v(0) + x v(1) \leq \max \cbr{v(0), v(1)} < \infty.
$
From this we conclude $  \sup_{x \in [0,1]} v(x) = \max \cbr{v(0), v(1)}$ and is attainable.
In addition, since $\mathrm{dom}(v) \supseteq [0,1]$, and $v$ is proper and closed, then the infimum is finite and attainable.
In addition, we clearly have
$\sup_{\pi \in \Pi} \abs{ w(\pi(\cdot|s))} \leq \abs{\cA} \sup_{x \in [0,1]} \abs{ v(x) }$.
\qed \end{proof}

In view of Lemma \ref{lemma_bounded_divergence}, the  existence of a finite $\Phi$ in Proposition \ref{claim_linear_decomposable_reg} can be readily satisfied by many practical regularizers, including the previously mentioned negative entropy, the $p$-th power of $\ell_p$-norm, and the negative Tsallis entropy.

Going forward, for presentation simplicity, we will focus on a single state $s$, and write $\pi_{k} $ in short for $\pi_k(\cdot|s)$, 
$Q^k$ in short for $Q^{\pi_k}(s, \cdot)$, 
 $Q^k_j$ in short for $Q^{\pi_k}(s, j)$,
  and $D^{p}_{\pi_k}$ in short for $D^{p}_{\pi_k}(s)$.
We begin by examining the update \eqref{hpmd_update_decomposable}, which provides an alternative characterization of the updated policy.
 
\begin{lemma}\label{lemma_decomposable_update_characterization}
Assuming $\partial \hat{v}(\pi_i^k) \neq \emptyset$ for every $i \in \cA$, then
 HPMD with the generalized update rule \eqref{hpmd_update_decomposable} satisfies 
\begin{align}
\pi^{k+1}_i 
= 
\nabla \hat{v}^* \rbr{ \frac{\nabla  \hat{v}(\pi^k_i) -\eta_k Q^k_i -  \lambda_k}{1+\eta_k \tau_k} },
 \label{connection_to_grad}
\end{align}
for some $\lambda_k \in \RR$ that does not depend on the action $i \in \cA$.
\end{lemma}  
\begin{proof}
Note that the update \eqref{hpmd_update_decomposable} is equivalent to 
\begin{align*}
\pi_{k+1} = 
\argmin_{p \in \RR^{\abs{\cA}}_{+} } \eta_k\sbr{ \inner{Q^k}{ p }  + \tau_k \hat{w}(p) } + D^{p}_{\pi_k},  ~~~
\mathrm{s.t.} ~ \tsum_{i=1}^{\abs{\cA}} p_i = 1,
\end{align*}
where $\RR^{\abs{\cA}}_{+}$ denotes the nonnegative orthant in $\RR^{\abs{\cA}}$.
The previous constrained convex optimization problem satisfies Slater condition. Hence there exists a Lagrange multiplier $\lambda_k \in \RR$, such that 
\begin{align}\label{general_update_equiv}
\pi_{k+1} = 
\argmin_{p \in \RR^{\abs{\cA}}_{+} } \eta_k\sbr{ \inner{Q^k}{ p }  + \tau_k \hat{w}(p) } + D^{p}_{\pi_k}
+ \lambda_k \rbr{ \tsum_{i=1}^{\abs{\cA}} p_i - 1}.
\end{align}
Since $\hat{w}$ is coordinate-wise decomposable and so is the constraint, the computation of \eqref{general_update_equiv} is separable. 
By denoting $\pi^k_i$ in short for $\pi_k(i|s)$, we equivalently have 
\begin{align}
\pi^{k+1}_i  = \argmin_{p_i \in \RR_+}  \eta_k\sbr{ \inner{Q^k_i}{ p_i }  + \tau_k \hat{v}(p_i) } + \hat{v}(p_i) - \hat{v}(\pi^k_i) - \inner{\nabla \hat{v}(\pi^k_i) }{p_i - \pi^k_i} + \lambda_k p_i.
\label{general_update_reg}
\end{align}
where $\nabla \hat{v}(\pi_i^k) \in \partial \hat{v}(\pi_i^k)$ denotes a subgradient of $ \hat{v}(\cdot)$ at $x$.
Hence we obtain
\begin{align}
\pi^{k+1}_i  & = \argmin_{p_i \in \RR_+}  \eta_k\sbr{ \inner{Q^k_i}{ p_i }  + \tau_k  \hat{v}(p_i) } +  \hat{v}(p_i) -  \hat{v}(\pi^k_i) - \inner{ \nabla  \hat{v}(\pi^k_i) }{p_i - \pi^k_i} + \lambda_k p_i \nonumber \\ 
& = \argmin_{p_i \in \RR_+} \inner{\eta_k Q^k_i -  \nabla  \hat{v}(\pi^k_i) + \lambda_k}{p_i} 
+(1+ \eta_k \tau_k)  \hat{v}(p_i) \nonumber \\
& =  \argmax_{p_i \in \RR_+} \inner{ \nabla  \hat{v}(\pi^k_i) -\eta_k Q^k_i -  \lambda_k}{p_i} 
- (1 +  \eta_k \tau_k)  \hat{v}(p_i) \nonumber \\
& \overset{(a)}{=} \argmax_{p_i \in \RR} \inner{ \frac{\nabla  \hat{v}(\pi^k_i) -\eta_k Q^k_i -  \lambda_k}{1+\eta_k \tau_k}}{p_i} 
- \hat{v}(p_i) \nonumber \\
& = \nabla \hat{v}^* \rbr{ \frac{\nabla  \hat{v}(\pi^k_i) -\eta_k Q^k_i -  \lambda_k}{1+\eta_k \tau_k} },\nonumber
\end{align}
where 
$(a)$ uses the definition of $\hat{v}$.
Note that here $\hat{v}^*$ is differentiable at $(\nabla  \hat{v}(\pi^k_i) -\eta_k Q^k_i -  \lambda_k)/\rbr{1+\eta_k \tau_k} $ as $\hat{v}$ is strictly convex, implying the maximizer of $(a)$ being unique.
\qed \end{proof}

To proceed,  the following lemma provides two simple conditions, under which $\partial \hat{v}(\pi_i^k) \neq \emptyset$ for every $k \geq 0$ and $i \in \cA$, and consequently certifying condition $(2)$ in Proposition \ref{claim_linear_decomposable_reg}.
\begin{lemma}\label{lemma_range_condition}
For HPMD with the generalized update \eqref{hpmd_update_decomposable}, we have 
\begin{itemize}
\item[I.]  If $\partial v(0) \neq \emptyset$, then $\partial v(0) \subseteq \partial \hat{v}(0)$.
\item[II.] If $\partial v(0) = \emptyset$, and $\pi_i^0 > 0$ for every $i \in \cA$, then $\pi_i^k > 0$  for all $k \geq 0$ and $i \in \cA$.
\end{itemize}
In either of the cases above, we have $ \partial \hat{v}(\pi_i^k) \neq \emptyset$ for every $i \in \cA$ and $k \geq 0$.
\end{lemma}
\begin{proof}
The first claim follows immediately from the definition of subgradient, $v(0) = \hat{v}(0)$, and $v(\cdot) \leq \hat{v}(\cdot)$.
To show the second claim, it suffices to show that $\pi_i^{k+1}$ in \eqref{general_update_reg} is always strictly positive if $\partial v(0) = \emptyset$ and $\pi_i^k > 0$.
Note that since $\pi_i^k > 0$, then $w$ is differentiable at $\pi_i^k$ and hence $\nabla v(\pi_i^k)$ exists and is finite. 
The subproblem \eqref{general_update_reg} then takes the form of 
\begin{align*}
\min_{p_i \in \RR_+} \psi(p_i) \coloneqq \inner{a}{p_i} + (1 + \eta_k \tau_k) v(p_i).
\end{align*}
Since $\partial v(0) = \emptyset$, we must have $\partial \psi(0) = \emptyset$, since other wise 
$\partial v(0) \supseteq \frac{1}{1+\eta_k \tau_k} \rbr{ \partial \psi(0) - a}$ is nonempty.
Thus the minimizer cannot  be $0$ and hence $\pi_i^{k+1} > 0$.

We then make the following observations.
If $\partial v(0) \neq \emptyset$, then 
 $ \partial \hat{v}(\pi_i^k)  \neq \emptyset$.
 If $\partial v(0) = \emptyset$, then it holds $\pi_i^k >0$ for all $k\geq 0$ and $i \in \cA$ whenever $\pi_i^0 > 0$ for all $i \in \cA$.
Hence $\hat{v}$ is differentiable at $\pi_i^k$, and  $ \partial \hat{v}(\pi_i^k)  \neq \emptyset$.
The proof is then completed.
\qed \end{proof}

Combining Proposition \ref{claim_linear_decomposable_reg}, Lemma \ref{lemma_bounded_divergence} and \ref{lemma_range_condition}, we can obtain the following global linear convergence of HPMD for any choice of subgradients $\{\nabla \hat{v}(\pi_i^k)\}$ in the generalized HPMD update \eqref{hpmd_update_decomposable}.
\begin{corollary}\label{cor_generalized_hpmd_linear_convergence}
Suppose either  (a) $\partial v(0) \neq \emptyset$; or (b) $\partial v(0) = \emptyset$, and $\pi_i^0 > 0$ for every $i \in \cA$.
Then 
for any choice of $\{ \nabla \hat{v}(\pi_i^k) \}$ with $\nabla \hat{v}(\pi_i^k) \in \partial \hat{v}(\pi_i^k) \neq \emptyset$,
HPMD with the generalized update \eqref{hpmd_update_decomposable} satisfies 
\begin{align*}
f(\pi_{k}) - f(\pi^*) 
\leq 
\gamma^k \rbr{
 f(\pi_{0}) - f(\pi^*)  + {4  \Phi}/\rbr{1-\gamma}
},
\end{align*}
for some $0 < \Phi < \infty$.
\end{corollary}

We next
consider a concrete choice of subgradients $\{\nabla \hat{v}(\pi_i^k)\}$ in the generalized update \eqref{hpmd_update_decomposable},
 for which we obtain an important recursion.
\begin{lemma}\label{lemma_dual_variable_recursion}
Consider HPMD with the generalized update \eqref{hpmd_update_decomposable}.
Suppose either (a) $\partial v(0) \neq \emptyset$; or (b) $\partial v(0) = \emptyset$, and $\pi_i^0 > 0$ for every $i \in \cA$.
In addition, let the subgradients $\{\nabla \hat{v}(\pi_i^k) \}$ in the update \eqref{hpmd_update_decomposable} be chosen as 
\begin{align}\label{eq_choice_of_subgradient}
\nabla \hat{v}(\pi_i^0) \in  \partial \hat{v}(\pi_i^0);~
\nabla \hat{v}(\pi_i^{k+1}) = \frac{\nabla  \hat{v}(\pi^k_i) -\eta_k Q^k_i -  \lambda_k}{1+\eta_k \tau_k} ,
~ \forall k \geq 0,
\end{align}
where $\lambda_k$ is defined as in Lemma \ref{lemma_decomposable_update_characterization}.
Then 
we obtain for any $k > 0$,
\begin{align}
\nabla \hat{v}(\pi^{k}_i)  - \nabla \hat{v}(\pi^{k}_j) 
 = \gamma^k \rbr{ \nabla \hat{v}(\pi^{0}_i)  - \nabla \hat{v}(\pi^{0}_j) } 
+ \tsum_{t=0}^{k-1} \gamma^{k - t} \eta_t (Q^t_i - Q^t_j), ~ \forall i,j \in \cA. \label{recursion_iter_decomposable_reg}
\end{align}
\end{lemma}

\begin{proof}
Since $\hat{v}$ is closed, 
$\partial \hat{v} = (\partial \hat{v}^*)^{-1}$. Combining this observation with   \eqref{connection_to_grad}, we obtain  
\begin{align*}
\partial \hat{v}(\pi_i^{k+1}) \ni 
\frac{\nabla \hat{v}(\pi^k_i) -\eta_k Q^k_i -  \lambda_k}{1+\eta_k \tau_k}  \overset{(a)}{=} \gamma \rbr{\nabla \hat{v}(\pi^k_i) - \eta_k Q^k_i } + \zeta_k,
\end{align*}
where $(a)$ uses the fact that $1 + \eta_k \tau_k = 1/\gamma$ and denote $\zeta_k = \gamma \lambda_k$.
Note that the definition of $\zeta_k$ does not depend on the choice of $i \in \cA$.
Given the choice of subgradients in \eqref{eq_choice_of_subgradient}, we can  recursively apply the prior relation, and obtain that for any $i,j\in \cA$,
\begin{align*}
\nabla \hat{v}(\pi^{k}_i)  - \nabla \hat{v}(\pi^{k}_j) 
& = \gamma \rbr{\nabla \hat{v}(\pi^{k-1}_i)  - \nabla \hat{v}(\pi^{k-1}_j)}
- \gamma \eta_{k-1} \rbr{Q^{k-1}_i -Q^{k-1}_j }  \\
& = \gamma^k \rbr{ \nabla \hat{v}(\pi^{0}_i)  - \nabla \hat{v}(\pi^{0}_j) } 
+ \tsum_{t=0}^{k-1} \gamma^{k - t} \eta_t (Q^t_i - Q^t_j) .
\end{align*}
The proof of \eqref{recursion_iter_decomposable_reg} is then completed.
\qed \end{proof}



Next, we introduce a key intermediate result, which shows that
$\cbr{\pi_k}$ converges to $\Pi^*$,
 with a local convergence rate explicitly depending on the tail of $\nabla \hat{v}^*$.

\begin{lemma}\label{hpmd_generalized_converge_to_optimal}
Under the same conditions in Lemma \ref{lemma_dual_variable_recursion}, 
let subgradients $\{\nabla \hat{v}(\pi_i^k)\}$ be set as in \eqref{eq_choice_of_subgradient}, 
then HPMD with the generalized update \eqref{hpmd_update_decomposable} satisfies 
\begin{align}\label{pi_ub_via_logit}
\pi_j^{k+1} \leq \nabla \hat{v}^* \rbr{- \gamma^{-{2k-1}} \frac{\Delta^*(\cM)}{2}  + \overline{v} + 2   \max_{i \in \cA} \abs{ \nabla  \hat{v}(\pi_i^0)}  + \frac{2 C }{(1-\gamma^3)(1-\gamma) \gamma} }, ~\forall j \notin \cA^*(s), 
\end{align}
for any $ k \geq K_1 = \max \cbr{ 3 { \log_\gamma \rbr{ \frac{ \Delta^*(\cM) (1-\gamma) }{ 2\varrho\rbr{ 4 \Phi + C} }} }, \frac{1}{2} \log_\gamma \rbr{
\frac{\Delta^*(\cM) (1-\gamma^3) ( 1-\gamma) \gamma}{4 ( \max_{i \in \cA} \abs{ \nabla  \hat{v}(\pi_i^0)} + C) }
}
}$
and $\overline{v} =  \nabla \hat{v}(1)$.
In addition,
\begin{align}\label{eq_hpmd_generalized_to_optimal_policy_set}
\pi_k \to \Pi^*.
\end{align}
\end{lemma}
\begin{proof}
It should be clear that relation \eqref{recursion_iter_decomposable_reg} takes the same structure as \eqref{difference_logit}, with $z_i^k = \log \pi_i^k$ replaced by  $\nabla \hat{v}(\pi^{k}_i)$, and 
an additional linearly converging term $ \gamma^k \rbr{ \nabla \hat{v}(\pi^{0}_i)  - \nabla \hat{v}(\pi^{0}_j) }$.
Our next few steps closely mirror those in the proof of Theorem \ref{thrm:convergence_to_optimal},
with some  overloading of notations.

Consider any pair of actions $(i, j)$, with $i \in \cA^*(s)$ and $j \notin \cA^*(s)$,
for which it holds that $Q^*(s,i) < Q^*(s,j)$. 
Similar to \eqref{q_convergence}, 
with Assumption \ref{assump_support}  and Proposition \ref{claim_linear_decomposable_reg}, we obtain
\begin{align}\label{q_convergence_generalized}
Q^{\pi_k}(s,a) - Q^*(s,a) \leq   \gamma^k 
\varrho(4  \Phi + C )/ (1-\gamma),
\end{align}
 from which we obtain
\begin{align}\label{diff_q_convergence_generalized}
Q_i^t - Q_j^t \leq (Q^*(s, i) - Q^*(s,j))/2 < 0, ~ \forall t \geq \underline{K}_1(s) \coloneqq  { \log_\gamma \rbr{ \frac{ \sbr{Q^*(s,j) - Q^*(s,i)} (1-\gamma) }{ 2\varrho\rbr{ 4 \Phi + C} }} }.
\end{align}
Recall that we choose $\eta_k = \gamma^{-2(k+1)}$.
Thus for any $k > \underline{K}_1(s)$, given \eqref{recursion_iter_decomposable_reg}, we have 
\begin{align*}
& ~~~~ \nabla \hat{v}(\pi_i^{k+1})  - \nabla \hat{v}(\pi_j^{k+1}) \\
& =  \gamma^{k+1} \rbr{ \nabla \hat{v}(\pi^{0}_i)  - \nabla  \hat{v}(\pi^{0}_j) }  - \tsum_{t=0}^{ \underline{K}_1(s) } \gamma^{k+1 - t} \eta_t ( Q_i^t - Q_j^t)
-  \tsum_{t=\underline{K}_1(s) + 1}^{ k } \gamma^{k+1 - t} \eta_t ( Q_i^t - Q_j^t) \\
& \geq
-2  \max_{i \in \cA} \abs{ \nabla  \hat{v}(\pi_i^0)}
 - \tsum_{t=0}^{ \underline{K}_1(s) } \gamma^{k-3t-1} ( Q_i^t - Q_j^t)
-  \tsum_{t=\underline{K}_1(s) + 1 }^{ k } \gamma^{k-3t-1}( Q_i^t - Q_j^t) \\
& \overset{(a)}{\geq} -2   \max_{i \in \cA} \abs{ \nabla  \hat{v}(\pi_i^0)} - \tsum_{t=0}^{ \underline{K}_1(s) } \gamma^{k-3t-1} \cdot \frac{2C}{1-\gamma}
+ \frac{ \gamma^{-2k - 1}}{2} \sbr{Q^*(s, j) - Q^*(s,i)}  \\
& \geq - 2 \max_{i \in \cA} \abs{ \nabla  \hat{v}(\pi_i^0)} - \frac{2 C \gamma^{k-3\underline{K}_1(s) }}{(1-\gamma^3)(1-\gamma) \gamma} 
+ \frac{\gamma^{-2k - 1}}{2} \sbr{Q^*(s, j) - Q^*(s,i)} ,
\end{align*}
where 
$(a)$ uses \eqref{diff_q_convergence_generalized}.
Thus  we obtain for any $  k \geq \tilde{K}_1 \coloneqq 3 { \log_\gamma \rbr{ \frac{ \Delta^*(\cM) (1-\gamma) }{ 2\varrho\rbr{ 4 \Phi + C} }} }$, 
\begin{align}
 \nabla \hat{v}(\pi_i^{k+1})  - \nabla \hat{v}(\pi_j^{k+1})  & \geq  -2   \max_{i \in \cA} \abs{ \nabla  \hat{v}(\pi_i^0)}  - \frac{2 C }{(1-\gamma^3)(1-\gamma) \gamma} + \frac{\gamma^{-2k - 1}}{2} \sbr{Q^*(s, j) - Q^*(s,i)} . \label{logit_diff_exp_generalized}
\end{align}

Going forward, we make use of the following fact for a univariate convex function $v: \RR \to \overline{\RR} $.
\begin{fact}\label{fact_monotone}
For any $x \neq y$ with non-empty subdifferential, we have $\inner{\nabla v(x) - \nabla v(y)}{x- y} \geq 0$, where $\nabla v(x) \in \partial v(x)$, $\nabla v(y) \in \partial v(y)$.
If $v$ is strictly convex, then the inequality is strict.
\end{fact}

Note that $\pi_i^{k+1} \leq 1$, which combined with  Fact \ref{fact_monotone}, $\mathrm{dom}(\hat{v}) = \RR_+$ and $\hat{v}$ being differentiable inside $\RR_{++}$, shows that for $\overline{v} =  \nabla \hat{v}(1)$, $\nabla \hat{v}( \pi_i^k) \leq  \overline{v}$ for any $i \in \cA^*(s)$.
From  \eqref{logit_diff_exp_generalized}, we then conclude that 
\begin{align}\label{logit_ub_exp_generalized}
\nabla \hat{v}(\pi_j^{k+1}) \leq - \gamma^{-{2k-1}} \frac{\Delta^*(\cM)}{2}  + \overline{v} + 2   \max_{i \in \cA} \abs{ \nabla  \hat{v}(\pi_i^0)}  + \frac{2 C }{(1-\gamma^3)(1-\gamma) \gamma} , ~ \forall j \notin \cA^*(s),
\end{align}
occurs for any $k \geq \tilde{K}_1$.

We proceed to establish that $\hat{v}^*(x) $ is differentiable  at any $x \leq \overline{v}$, with
$\lim_{x \to -\infty} \nabla \hat{v}^*(x) = 0$.
Recall 
\begin{align}
\nabla \hat{v}^*(x) = \argmax_{y \in \RR} \inner{x}{y} - \hat{v}(y) 
= \argmin_{y \in \RR} \inner{-x}{y} + \hat{v}(y)
\coloneqq \argmin_{y \in \RR} \phi(y), \label{def_subgrad_v_conjugate}
\end{align}
whenever the optimization problem above is solvable.
For any $\epsilon > 0$,   $\hat{v} \equiv v$ is differentiable at $\epsilon$.
Consider 
$
u(y) = \sbr{
\inner{-x}{y} + \hat{v}(\epsilon) + \inner{\nabla \hat{v}(\epsilon)}{y - \epsilon}
}
\cX_{\RR_+}(y),
$
where $\cX_{\RR_+}$ denotes the characteristic function of $\RR_+$.
Since $\hat{v}$ is strictly convex, it is clear that
\begin{align}
u(y) < \phi(y), ~ \forall y > \epsilon; ~ u(\epsilon) = \phi(\epsilon), \label{ineq_aux_function}
\end{align}
for any $x \in \RR$.
On the other hand, for $x \leq \nabla \hat{v}(\epsilon)$, $u(\cdot)$ is non-decreasing over $\RR_+$.
Combining this with \eqref{ineq_aux_function}, we obtain 
$
\phi(y) > u(y) \geq u(\epsilon) = \phi(\epsilon)$, for  $y > \epsilon$.
Thus it must holds $\nabla \hat{v}^*(x) = \argmin_{y \in \RR} \phi(y) = \argmin_{y \in  \RR_+} \phi(y) \leq \epsilon$.
In conclusion, for any $\epsilon > 0$, 
\begin{align}\label{existence_of_grad_vhat_conjugate}
\text{
if $x \leq \nabla \hat{v}(\epsilon)$, then $\nabla \hat{v}^*(x)$ exists, with $\nabla \hat{v}^*(x) \leq \epsilon$,
}
\end{align}
from which we immediately obtain  $\lim_{x \to -\infty} \nabla \hat{v}^*(x) = 0$.
Hence for any $j \notin \cA^*(s)$, and any $k \geq K_1$, 
\begin{align*}
\pi_j^{k+1} & = \nabla \hat{v}^* (\nabla \hat{v}(\pi_j^{k+1})) \leq \nabla \hat{v}^* \rbr{ - \gamma^{-{2k-1}} \frac{\Delta^*(\cM)}{2}  + \overline{v} + 2   \max_{i \in \cA} \abs{ \nabla  \hat{v}(\pi_i^0)}  + \frac{2 C }{(1-\gamma^3)(1-\gamma) \gamma} },
\end{align*}
where the first equality holds since $\hat{v}$ is closed and strictly convex,
and the second inequality uses \eqref{logit_ub_exp_generalized},   Fact \ref{fact_monotone}, and the fact that for
$
k \geq \frac{1}{2} \log_\gamma \rbr{
\frac{\Delta^*(\cM) (1-\gamma^3) ( 1-\gamma) \gamma}{4 ( \max_{i \in \cA} \abs{ \nabla  \hat{v}(\pi_i^0)} + C) }
}
$, 
\begin{align*}
x_k \coloneqq - \gamma^{-{2k-1}} \frac{\Delta^*(\cM)}{2}  + \overline{v} + 2   \max_{i \in \cA} \abs{ \nabla  \hat{v}(\pi_i^0)}  + \frac{2 C }{(1-\gamma^3)(1-\gamma) \gamma}
\leq \overline{v},
\end{align*}
and consequently $\hat{v}^*$ is differentiable at $x_k$ given the \eqref{existence_of_grad_vhat_conjugate}. 
Hence \eqref{pi_ub_via_logit} is proved. 
 \eqref{eq_hpmd_generalized_to_optimal_policy_set} then follows from \eqref{pi_ub_via_logit}, and $\lim_{x \to -\infty} \nabla \hat{v}^*(x) = 0$.
\qed \end{proof}

As an immediate corollary of Lemma \ref{hpmd_generalized_converge_to_optimal}, HPMD with the generalized update rule \eqref{hpmd_update_decomposable} converges at a faster-than-linear rate locally, provided that $\nabla \hat{v}^*$ has a light tail.
We will also provide readily verifiable conditions that can certify such light-tailed behavior of $\nabla \hat{v}^*$.
In particular, there exists a fairly broad class of distance-generating functions that can guarantee the finite-time exact convergence of the optimality gap. 

\begin{lemma}[Local Convergence of Generalized HPMD]\label{lemma_hpmd_generalized_local_convergence}
Assume the same settings in Lemma \ref{hpmd_generalized_converge_to_optimal}. 
Suppose $\lim_{x \to -\infty}  - \nabla \hat{v}^*(x) x = 0$, then for any $\epsilon > 0$, there exists $\underline{K}_2  (\epsilon) \geq K_1$,
where $K_1$ is defined as in Lemma \ref{hpmd_generalized_converge_to_optimal}, 
 such that for any $k \geq \underline{K}_2(\epsilon)$,
\begin{align}
\mathrm{dist}_{\ell_{1}} (\pi_{k+1} ,\Pi^*)  & \leq 2 \epsilon  \abs{\cA} \gamma^{2k} ,  \label{eq_policy_sup_convergenvce_generalized}\\
V^{\pi_{k+1}}(s) - V^*(s)   \leq
\frac{2 \epsilon  \abs{\cA} C}{(1-\gamma)^2} \gamma^{2k}  ~ & \text{and} ~~
Q^{\pi_{k+1}}(s, a) - Q^*(s, a)   \leq
\frac{2 \epsilon   \abs{\cA} C \gamma}{(1-\gamma)^2} \gamma^{2k}, ~ \forall (s,a) \in \cS \times \cA. \label{eq_value_sup_convergence_generalized}
\end{align}
In particular, if $0 \in \partial \hat{v}(0)$, then one can take $\underline{K}_2
\equiv
\frac{1}{2} \log_\gamma \rbr{
\frac{\Delta^*(\cM) (1-\gamma^3) ( 1-\gamma) \gamma}{4 ( \max_{i \in \cA} \abs{ \nabla  \hat{v}(\pi_i^0)} + C + \abs{\overline{v}}) }
} + K_1$, 
and for any $k \geq \underline{K}_2$, it holds 
\begin{align*}
\mathrm{dist}_{\ell_{1}} (\pi_{k+1} ,\Pi^*) = 0; ~
V^{\pi_{k+1}}(s) = V^*(s) ; ~
Q^{\pi_{k+1}}(s, a) = Q^*(s, a) . ~ \forall (s,a) \in \cS \times \cA.
\end{align*}
\end{lemma}

\begin{proof}
Since \eqref{pi_ub_via_logit} in Lemma \ref{hpmd_generalized_converge_to_optimal} holds for any state $s \in \cS$, then 
\begin{align*}
\mathrm{dist}_{\ell_{1}} (\pi_{k+1} ,\Pi^*)  \leq 2 \abs{\cA} 
 \nabla \hat{v}^* \rbr{- \gamma^{-{2k-1}} \frac{\Delta^*(\cM)}{2}  + \overline{v} + 2   \max_{i \in \cA} \abs{ \nabla  \hat{v}(\pi_i^0)}  + \frac{2 C }{(1-\gamma^3)(1-\gamma) \gamma} },
\end{align*}
for any $ k \geq K_1$.
In addition, recall that \eqref{ineq_from_dist_to_value} still holds, then 
\begin{align*}
V^{\pi_{k+1}}(s) - V^*(s)  
& \leq \frac{C }{(1-\gamma)^2} \mathrm{dist}_{\ell_{1}} (\pi_{k+1} ,\Pi^*) \\
& \leq 
\frac{2 \abs{\cA} C}{(1-\gamma)^2}
 \nabla \hat{v}^* \rbr{- \gamma^{-{2k-1}} \frac{\Delta^*(\cM)}{2}  + \overline{v} + 2   \max_{i \in \cA} \abs{ \nabla  \hat{v}(\pi_i^0)}  + \frac{2 C }{(1-\gamma^3)(1-\gamma) \gamma} }.
\end{align*}
Finally, since $\nabla \hat{v}^*(x) = \smallO(-x^{-1})$, then one can find, for any $\epsilon > 0$, a $\underline{K}_2 \geq K_1$ such that 
\begin{align}
\nabla \hat{v}^* \rbr{- \gamma^{-{2k-1}} \frac{\Delta^*(\cM)}{2}  + \overline{v} + 2   \max_{i \in \cA} \abs{ \nabla  \hat{v}(\pi_i^0)}  + \frac{2 C }{(1-\gamma^3)(1-\gamma) \gamma} }
\leq 
\epsilon \cdot  \gamma^{2k}, ~ \forall k \geq \underline{K}_2. \label{def_uK2_from_epsilon}
\end{align}
The proof for \eqref{eq_policy_sup_convergenvce_generalized} and \eqref{eq_value_sup_convergence_generalized} is completed by combining above observations with \eqref{convergence_q_super}.

To show the second part of the claim, recall that \eqref{def_subgrad_v_conjugate} provides a characterization of $\nabla \hat{v}^*(x)$. 
In particular, if $x \leq 0$, and $0 \in \partial \hat{v}(0)$,
then $- x \in \partial \phi(0)$ and $\inner{-x}{y} \geq 0$ for any $y \in \RR_+$. 
Consequently, we have $\phi(y) > \phi(0)$ for any $y \geq 0$ given the strict convexity of $\phi$,  and thus $\nabla \hat{v}^*(x) = \argmin_{y \in \RR} \phi(y) = 0$ for $x \leq 0$.
Hence by taking $\underline{K}_2 =  
\frac{1}{2} \log_\gamma \rbr{
\frac{\Delta^*(\cM) (1-\gamma^3) ( 1-\gamma) \gamma}{4 ( \max_{i \in \cA} \abs{ \nabla  \hat{v}(\pi_i^0)} + C + \abs{\overline{v}}) }
} + K_1$,
we have
\begin{align*}
\nabla \hat{v}^* \rbr{- \gamma^{-{2k-1}} \frac{\Delta^*(\cM)}{2}  + \overline{v} + 2   \max_{i \in \cA} \abs{ \nabla  \hat{v}(\pi_i^0)}  + \frac{2 C }{(1-\gamma^3)(1-\gamma) \gamma} } = 0, ~ \forall k \geq \underline{K}_2.
\end{align*}
The proof is then completed.
\qed \end{proof}

The next lemma then shows that for any pair of actions $i, j \in \cA^*(s)$, then their corresponding dual variables $\{\nabla \hat{v}(\pi_i^{k})\}$ and $\{\nabla \hat{v}(\pi_j^{k})\}$ will be asymptotically equal to each other. 

\begin{lemma}\label{lemma_hpmd_dual_asymp_equal_opt_action}
With the same settings in Lemma \ref{lemma_hpmd_generalized_local_convergence}, 
then for any $\epsilon > 0$, and any 
\begin{align*}
k \geq K_2 (\epsilon) \coloneqq 2 \underline{K}_2 \rbr{\frac{(1-\gamma)^3 \epsilon}{4 \abs{\cA} C}}
+ \log_\gamma \rbr{
\frac{ (1-\gamma^2) (1-\gamma) \gamma \epsilon}{8 \varrho (4 \Phi + C) }
}
+ 
\log_\gamma \rbr{
\frac{ \epsilon}{ 8 \max_{i \in \cA} \abs{\nabla \hat{v}(\pi_i^0)} }
}, 
\end{align*}
it holds that
$\abr{ \nabla \hat{v}(\pi_i^{k+1})  - \nabla \hat{v}(\pi_j^{k+1})}
\leq \epsilon$, 
where $\underline{K}_2 (\cdot) $ is defined as in Lemma \ref{lemma_hpmd_generalized_local_convergence}.
\end{lemma}
\begin{proof}
Given recursion \eqref{recursion_iter_decomposable_reg} in Lemma \ref{lemma_dual_variable_recursion}, 
for any $k \geq \underline{K}_2 (\epsilon)$, defined as in Lemma \ref{lemma_hpmd_generalized_local_convergence}, we have for any $i, j \in \cA^*(s)$, 
\begin{align*}
& ~~~~\abr{ \nabla \hat{v}(\pi_i^{k+1})  - \nabla \hat{v}(\pi_j^{k+1})} \\
& \leq  \gamma^{k+1} \abr{ \nabla \hat{v}(\pi^{0}_i)  - \nabla  \hat{v}(\pi^{0}_j) }  + \underbrace{\tsum_{t=0}^{ \underline{K}_2 } \gamma^{k+1 - t} \eta_t \abr{ Q_i^t - Q_j^t }}_{(A)}
+  \underbrace{\tsum_{t=\underline{K}_2 + 1}^{ k } \gamma^{k+1 - t} \eta_t \abr{ Q_i^t - Q_j^t}}_{(B)}.
\end{align*}
We proceed to bound term $(A)$ and $(B)$ separately. 
For term $(A)$, following the same reasoning as in \eqref{bound:A} in conjunction with Corollary \ref{cor_generalized_hpmd_linear_convergence}, 
\begin{align*}
(A) = \tsum_{t = 0}^{ \underline{K}_2} \gamma^{k+1 - t} \eta_t \abs{Q_i^t - Q_j^t} 
\leq \tsum_{t=0}^{\underline{K}_2}  \gamma^{k-2t} \frac{2 \varrho (4  \Phi + C)}{(1-\gamma)\gamma}
\leq \gamma^{k-2\underline{K}_2 } \frac{2 \varrho (4\log \Phi + C)}{(1-\gamma) (1-\gamma^2) \gamma}.
\end{align*}
For term $(B)$, since $Q^*(s,i) = Q^*(s,j)$, then applying Lemma \ref{lemma_hpmd_generalized_local_convergence} gives
\begin{align*}
\tsum_{t = \underline{K}_2 }^k \gamma^{k + 1 -t} \eta_t \abs{Q_i^t - Q_j^t} 
& \leq \tsum_{t =\underline{K}_2 }^k \gamma^{k + 1 -t} \eta_t \rbr{ \abs{Q_i^t - Q^*(s,i)} + \abs{Q_j^t - Q^*(s,j)} } \nonumber 
\\
&  \leq  \tsum_{t =\underline{K}_2 }^k \gamma^{k -3 t}  \frac{2 \epsilon   \abs{\cA} C }{(1-\gamma)^2 } \gamma^{2t}  
\leq \frac{2 \abs{\cA} C}{(1-\gamma)^3} \epsilon.
\end{align*}
Hence by combining the above observations, it holds that
\begin{align}\label{hpmd_generalized_dual_close_for_opt_action}
\abr{ \nabla \hat{v}(\pi_i^{k+1})  - \nabla \hat{v}(\pi_j^{k+1})}
\leq 
2 \gamma^{k+1} \max_{i \in \cA} \abs{\nabla \hat{v}(\pi_i^0)} 
+ \gamma^{k-2\underline{K}_2 } \frac{2 \varrho (4 \Phi + C)}{(1-\gamma) (1-\gamma^2) \gamma}
+ \frac{2 \abs{\cA} C}{(1-\gamma)^3} \epsilon.
\end{align}

Based on \eqref{hpmd_generalized_dual_close_for_opt_action}, it is then clear that 
for any $\epsilon > 0$, 
 for 
\begin{align*}
K_2 = 2 \underline{K}_2 (\epsilon)
+ \log_\gamma \rbr{
\frac{\abs{\cA} C (1-\gamma^2) \gamma \epsilon}{2 (1-\gamma)^2 \varrho (4 \Phi + C) }
}
+ 
\log_\gamma \rbr{
\frac{\abs{\cA} C \epsilon}{2 (1-\gamma)^3 \max_{i \in \cA} \abs{\nabla \hat{v}(\pi_i^0)} }
},
\end{align*}
then 
$
\abr{ \nabla \hat{v}(\pi_i^{k+1})  - \nabla \hat{v}(\pi_j^{k+1})}
\leq 
\frac{4 \abs{\cA} C}{(1-\gamma)^3} \epsilon
$,
 whenever $k \geq K_2$.
 Taking $\epsilon \to \frac{(1-\gamma)^3 \epsilon}{4 \abs{\cA} C}$ completes the proof.
\qed \end{proof}

In view of Lemma \ref{lemma_hpmd_dual_asymp_equal_opt_action}, one can then certify that
for any $i,j\in \cA^*(s)$, HPMD with the generalized update rule \eqref{hpmd_update_decomposable} satisfies
$\pi_i^k - \pi_j^k \to 0$.

\begin{lemma}\label{lemma_logit_equal_decomposable}
With the same settings in Lemma \ref{lemma_hpmd_generalized_local_convergence},
we have 
$\pi_i^k - \pi_j^k \to 0$ for any $i,j\in \cA^*(s)$.
\end{lemma}
\begin{proof}
Suppose the claim does not hold, then
for some action pair $i, j \in \cA^*(s)$, 
 there exists $\epsilon > 0$ such that for any $N \geq 0$, one can find $k > N$ with 
$\abs{\pi_i^k - \pi_j^k} > \epsilon$.
Additionally, since $\pi_i^k, \pi_j^k$ in $[0,1]$, we can then construct a subsequence $\cbr{k_l}$ such that 
$
\abs{\pi^{k_l}_i - \pi^{k_l}_j} > \epsilon, ~ \pi^{k_l}_i \to \pi_i^{\dagger}, ~ \pi^{k_l}_j \to \pi_j^{\dagger},
$
which also implies $\abs{\pi_i^{\dagger} - \pi_j^{\dagger}} \geq \epsilon$.
Suppose, without loss of generality, that $\pi_i^\dagger > \pi_j^\dagger$. Then given $0 < \delta < (\pi_i^\dagger - \pi_j^\dagger) / 2$,  there exists $L$ such that
$\pi^{k_l}_i \geq \pi_i^\dagger - \delta$, and $\pi^{k_l}_j < \pi_j^\dagger + \delta$ for $l \geq L$.
Hence given Fact \ref{fact_monotone} and the strict convexity of $\hat{v}$, 
\begin{align*}
\nabla \hat{v}(\pi_j^{k_l}) \leq \nabla  \hat{v}( \pi_j^\dagger + \delta) < \nabla  \hat{v}( \pi_i^\dagger - \delta) \leq \nabla  \hat{v}(\pi_i^{k_l}) , ~ \forall l \geq L,
\end{align*}
for any choice of $\nabla \hat{v}(\pi_j^{k_l}) \in \partial  \hat{v}(\pi_j^{k_l}), ~ \nabla \hat{v}(\pi_i^{k_l}) \in \partial  \hat{v}(\pi_i^{k_l})$.
This clearly contradicts with Lemma \ref{lemma_hpmd_dual_asymp_equal_opt_action}, and hence the proof is completed.
\qed \end{proof}

We are now ready to state the main result in this section, which certifies the existence of a limiting policy, and establishes its correspondence to the minimal-complexity optimal policy, where the complexity is measured exactly by the distance-generating function \eqref{decomposable_reg_main}.
\begin{theorem}\label{thrm_limit_policy_decomposable}
For HPMD with the generalized update \eqref{hpmd_update_decomposable}, suppose  
\begin{enumerate}
\item $v(\cdot)$ is proper, closed, strictly convex, with $\mathrm{dom}(v) \supseteq \RR_+$,  and differentiable inside $\mathrm{Int}(\mathrm{dom}(v))$;
\item $\lim_{x \to \infty} \nabla \hat{v}^*(-x)  x = 0$, or equivalently, 
$\nabla \hat{v}^* (-x) = \smallO(x^{-1})$ as $x \to \infty$.
\item  The subgradients $\{\nabla \hat{v}(\pi_i^k) \}$ in the update \eqref{hpmd_update_decomposable} are chosen as in \eqref{eq_choice_of_subgradient}.
\end{enumerate}
Then for any initial policy $\pi_0$ satisfying $\min_{s\in \cS, a \in \cA} \pi_0(a|s) >0$,  we have
\begin{align}\label{limit_policy_decomposable_main}
\lim_{k \to \infty} \pi_k   (a|s) = \pi_U^*  (a|s) \coloneqq 
\begin{cases}
1/ \abs{\cA^*(s)}, ~ &a \in \cA^*(s), \\
0, ~ & a \notin \cA^*(s).
\end{cases}
\end{align}
Furthermore, if $\partial v(0) \neq \infty$, then \eqref{limit_policy_decomposable_main} holds with any initial policy $\pi_0 \in \Pi$.

Moreover,
$\pi^*_U$ corresponds to the unique optimal policy with the minimal complexity measured by $w$.
That is, 
\begin{align}\label{eq_opt_minimal_complexity_policy}
w( \pi^*_U (\cdot|s) ) =  \min_{\pi \in \Pi^*} w(\pi(\cdot|s)), ~ \forall s \in \cS,
\end{align}
where $\Pi^*$ denotes the set of optimal policies, and the minimizer above is unique.
\end{theorem}

\begin{proof}
Note that conditions in Lemma  \ref{hpmd_generalized_converge_to_optimal}  and \ref{lemma_logit_equal_decomposable}  are satisfied. 
Hence for any $i \notin \cA^*(s)$, $\pi_i^k \to 0$; for any $i, j \in \cA^*(s)$, $ \pi_i^k - \pi_j^k \to 0$.
Thus \eqref{limit_policy_decomposable_main} follows immediately from this observation.

We proceed to establish \eqref{eq_opt_minimal_complexity_policy}. 
Given Lemma \ref{lemma:optimal_policy_set}, for each state $s \in \cS$,  \eqref{eq_opt_minimal_complexity_policy} is equivalent to 
\begin{align*}
p^* =  \argmin_{ p \in \RR^{\abs{\cA}} } w(p), ~ \mathrm{s.t.}~ \tsum_{i \in \cA^*(s)} p_i = 1, ~  p_j \geq 0, ~\forall j.
\end{align*}
Here $p^*$ is unique since $w$ is strictly convex.
Given definition \eqref{decomposable_reg_main}, $w(x) = w(x')$ if $x'$ results from permutation of entries in $x$.
We claim that $p^*$ satisfies $p^*_i =  p^*_j$ for all $i,j \in \cA^*(s)$, from which  \eqref{eq_opt_minimal_complexity_policy} follows. 
If not, then $p^*_i \neq p^*_j$ for some $i, j \in \cA^*(s)$. 
Consider $\tilde{p}^*$, given by exchanging $p^*_i$ and $p^*_j$ in $p^*$. 
Then from the strict convexity of $w$, we have 
$w\rbr{ \frac{1}{2} ( p^* + \tilde{p}^*)} < \frac{1}{2} w(p^*) + \frac{1}{2} w(\tilde{p}^*) = w(p^*)$, a contraction. 
Thus the proof of \eqref{eq_opt_minimal_complexity_policy} is completed.
\qed \end{proof}

By  specializing Theorem \ref{thrm_limit_policy_decomposable}, we can then recover the previously obtained result (Theorem \ref{thrm:prob_lb}), when $w$ corresponds to the negative entropy function.
 
\begin{corollary}[Negative entropy, restatement of Theorem \ref{thrm:prob_lb}]\label{cor_recover_entropy_parameter}
Let $v(x) = x\log x$ for $x > 0$ and $v(0) = 0$.
Then for any initial policy $\pi_0$ with $\min_{s \in \cS, a \in \cA}\pi_0(a|s) > 0$,
HPMD with the generalized update \eqref{hpmd_update_decomposable} exhibits the last-iterate policy convergence,  with $\lim_{k \to \infty} \pi_k = \pi^*_U$.
\end{corollary}

\begin{proof}
Note that $v(x) = x \log x$ satisfies condition 1 and 3 in Theorem \ref{thrm_limit_policy_decomposable} trivially.
In particular, the subgradients in \eqref{eq_choice_of_subgradient} reduces to the gradient in this case.
Direct calculation also yields $\nabla \hat{v}^*(x) = \exp ( x-1)$ for any $x \in \RR$. Hence condition 2 is also satisfied, the proof is then completed.
\qed \end{proof}

To illustrate the broad applicability of our discussions in this section, we proceed to establish non-asymptotic characterization for both the optimality gap and the policy, for other common distance-generating functions.

\begin{corollary}[$p$-th power of $\ell_p$-norm]\label{corr_p_norm}
For any $p \in (1, \infty)$, let $v(x) = \abs{x}^p$, i.e., let the  $w(\cdot) = \norm{\cdot}_p^p$ be the distance-generating function for HPMD with the generalized update \eqref{hpmd_update_decomposable}. 
Then the following holds. 
\begin{itemize}
\item[1.] Linear convergence of the optimality gap: 
\begin{align}\label{pnorm_linear_convergence}
f(\pi_{k}) - f(\pi^*) 
\leq 
\gamma^k \rbr{
 f(\pi_{0}) - f(\pi^*)  + 8 / (1-\gamma)
}, ~\forall k \geq 0.
\end{align}

\item[2.]
Finite-time exact convergence:
\begin{align}\label{pnorm_finite_time_convergence}
\pi_{k+1} \in \Pi^*, ~ f(\pi_{k+1}) = f(\pi^*), 
~ \forall k \geq \underline{K}_2, 
\end{align}
where 
$
\underline{K}_2
= 
 \log_\gamma \rbr{
\frac{\Delta^*(\cM) (1-\gamma^3) ( 1-\gamma) \gamma}{4 ( \max_{i \in \cA} \abs{ \nabla  \hat{v}(\pi_i^0)} + C + p) }
}
+ 
  3 { \log_\gamma \rbr{ \frac{ \Delta^*(\cM) (1-\gamma) }{ 2\varrho\rbr{ 8 + C} }} }.
  $
  \vspace{0.1in}
\item[3.] Last-iterate policy convergence: 
for any $\epsilon \in (0,1)$, it holds that
\begin{align}\label{pnorm_policy_convergence}
\norm{\pi_{k+1} - \pi^*_U}_\infty \leq \epsilon,
~ \forall k \geq K_2 \coloneqq
2 \underline{K}_2
+ \log_\gamma \rbr{
\frac{ (1-\gamma^2) (1-\gamma) \gamma \epsilon'}{8 \varrho (8 + C) }
}
+ 
\log_\gamma \rbr{
\frac{ \epsilon'}{ 8 \max_{i \in \cA} \abs{\nabla \hat{v}(\pi_i^0)} }
},
\end{align}
where $\epsilon' 
= 
\min \cbr{
{\epsilon}/{\abs{\cA}}, 
p(p-1) \epsilon, 
 \rbr{\abs{\cA}}^{1-p} p / 2
}$.
\end{itemize}

\end{corollary}

\begin{proof}
By definition, it is immediate that $\sup_{\pi \in \Pi, s\in \cS} 2 \abs{ w(\pi(\cdot|s))} \leq 2 $, and hence both conditions in Proposition \ref{claim_linear_decomposable_reg} are satisfied with $\Phi = 2$, hence applying Proposition \ref{claim_linear_decomposable_reg} gives \eqref{pnorm_linear_convergence}.
In addition, since $0 \in \partial \hat{v}(0)$, one can then apply Lemma \ref{lemma_hpmd_generalized_local_convergence} and obtain
\eqref{pnorm_finite_time_convergence}.

We proceed to establish \eqref{pnorm_policy_convergence}.
Note that $\tsum_{i \notin \cA^*(s)} \pi_i^{k+1} = 0$ for $k \geq \underline{K}_2$.
In view of Lemma \ref{lemma_hpmd_dual_asymp_equal_opt_action}, suppose we have 
\begin{align}\label{precondition_close_dual}
\abr{ \nabla \hat{v}(\pi_i^{k+1})  - \nabla \hat{v}(\pi_j^{k+1})}
\leq \epsilon', ~ \forall i , j \in \cA^*(s), 
\end{align}
for some $\epsilon' > 0$, and $k \geq K_2(\epsilon')$.
Let $i^* \in \Argmax_{i \in \cA^*(s)}  \pi_i^{k+1}$, then it is clear 
$\pi_{i^*}^{k+1} \geq 1/ \abs{\cA^*(s)}$.
Combining this observation with \eqref{precondition_close_dual} and Fact \ref{fact_monotone}, 
\begin{align*}
\nabla \hat{v} (\pi_i^{k+1})
\geq \nabla v(\pi_{i^*}^{k+1}) - \epsilon' 
\geq \nabla v(1/ \abs{\cA^*}) - \epsilon' > 0,  ~ \forall \nabla \hat{v} (\pi_i^{k+1}) \in \partial \hat{v} (\pi_i^{k+1}), ~ \forall i \in \cA^*(s), 
\end{align*}
if $\epsilon ' < \nabla v(1/ \abs{\cA^*}) = p(1/\abs{\cA^*(s)})^{p-1}$.
Given definition of $\hat{v}$, this in turn implies $\pi_i^{k+1} > 0$ for any $i \in \cA^*(s)$, for which $v$ is differentiable.
In this case, from \eqref{precondition_close_dual} we obtain 
\begin{align}
 \nabla v(\pi_{i^*}^{k+1})  - \nabla v(\pi_i^{k+1}) \leq \epsilon', 
 ~
 \Rightarrow
 ~
 \pi_i^{k+1} \geq \sbr{ \rbr{\abs{\cA^*(s)}}^{1-p} - {\epsilon'}/{p}}^{\frac{1}{p-1}}, ~ \forall i \in \cA^*(s).
 \label{eq_pnorm_prob_lb}
\end{align}

Since by definition, $ \nabla v(\pi_{i^*}^{k+1})  - \nabla v(\pi_i^{k+1}) = p (p-1) \zeta^{p-2} \rbr{\pi_{i^*}^{k+1} - \pi_i^{k+1}} $
for some $\zeta \in [ \pi_i^{k+1}, \pi_{i^*}^{k+1}]$, then for $p \geq 2$, we obtain 
\begin{align*}
\epsilon' 
\geq 
 \nabla v(\pi_{i^*}^{k+1})  - \nabla v(\pi_i^{k+1}) 
 & = p (p-1) \zeta^{p-2} \rbr{\pi_{i^*}^{k+1} - \pi_i^{k+1}} \\
&  \geq 
 p (p -1) \sbr{ \rbr{\abs{\cA^*(s)}}^{1-p} - {\epsilon'}/{p}}^{\frac{1}{p-1}}
  \rbr{\pi_{i^*}^{k+1} - \pi_i^{k+1}} \\
 & \geq 
  \rbr{\pi_{i^*}^{k+1} - \pi_i^{k+1}} / \abs{\cA^*(s)},
\end{align*}
provided $\epsilon '  \leq p(1/\abs{\cA^*(s)})^{p-1} /2 $.
Hence $\pi_{i^*}^{k+1} - \pi_i^{k+1} \leq \epsilon ' \abs{\cA^*(s)}$. 
For $p \in (1,2)$, we have 
$\epsilon ' \geq  \nabla v(\pi_{i^*}^{k+1})  - \nabla v(\pi_i^{k+1}) = p (p-1) \zeta^{p-2} \rbr{\pi_{i^*}^{k+1} - \pi_i^{k+1}} \geq p (p-1) \rbr{\pi_{i^*}^{k+1} - \pi_i^{k+1}} $.
In summary, if $\epsilon ' \leq p(1/\abs{\cA^*(s)})^{p-1} /2 $, 
then 
\begin{align}\label{ineq_pnorm_close_policy}
\pi_{i^*}^{k+1} - \pi_i^{k+1} \leq \max \cbr{
 \epsilon ' \abs{\cA^*(s)},
 {\epsilon'}/\sbr{p (p-1)}
} \coloneqq \delta, ~ \forall i \in \cA^*(s).
\end{align}
Thus following the same lines as in the proof of Lemma \ref{lemma_prob_ub_and_lb} for showing \eqref{lemma_eq_prob_ub_and_lb}, we obtain 
$\abr{ \pi_i^{k+1} - 1/\abs{\cA^*(s)} } \leq \delta$ 
for any $k \geq \underline{K}_2$, provided $\epsilon '  \leq p(1/\abs{\cA^*(s)})^{p-1} /2 $.
To obtain $\norm{\pi_k(\cdot|s)  - \pi^*_U(\cdot|s)}_\infty \leq \epsilon$, it then suffices to take
$\epsilon' 
= 
\min \cbr{
{\epsilon}/{\abs{\cA^*(s)}}, 
p(p-1) \epsilon, 
 \rbr{\abs{\cA^*(s)}}^{1-p} p / 2
}$. 
Finally, given Lemma \ref{lemma_hpmd_dual_asymp_equal_opt_action},  \eqref{precondition_close_dual} with $\epsilon'$ specified above can be satisfied 
whenever $k \geq K_2$, where 
\begin{align*}
K_2 = 2 \underline{K}_2
+ \log_\gamma \rbr{
\frac{ (1-\gamma^2) (1-\gamma) \gamma \epsilon'}{8 \varrho (8 + C) }
}
+ 
\log_\gamma \rbr{
\frac{ \epsilon'}{ 8 \max_{i \in \cA} \abs{\nabla \hat{v}(\pi_i^0)} }
}.
\end{align*}
Hence the proof for \eqref{pnorm_policy_convergence} is completed.
\qed \end{proof}

To proceed, let us first recall the Tsallis entropy for a discrete distribution. 
\begin{definition}[Tsallis Entropy]
For a discrete distribution $\cbr{p_i}$, the Tsallis entropy with entropic index $q \in \RR$ is defined as 
$
S_q(p) = \frac{z}{q-1} \rbr{1 - \tsum_i p_i^q}
$,
where $z$ is a positive constant.
\end{definition}

Now consider function $w_q(p) = \tsum_{i} p_i^q$ if $q > 1$, and $w_q(p) = -\tsum_i p_i^q$ if $q  \in (0,1)$.
It should be clear that from the definition of the generalized update rule \eqref{hpmd_update_decomposable}, setting $w(\cdot) = - S_q(\cdot)$ with parameters $\cbr{(\eta_k, \tau_k)}$ 
 is equivalent to using $w(\cdot) = - w_q(\cdot)$ with $\cbr{(\eta_k', \tau_k')}$, for $\eta_k' =  \eta_k z / \abs{q-1}$ and $\tau_k' = \tau_k \abs{q-1} / z$.
 To simplify our ongoing discussions, we will identify the Tsallis entropy with the entropic index $q$ with function $w_q(\cdot)$.

\begin{corollary}[Negative Tsallis entropy]\label{cor_tsallis_entropy_parameter}
 Let the negative Tsallis entropy $w_q(\cdot)$ with entropic index $q >0 $ and $q \neq 1$ be the distance-generating function for HPMD with the generalized update \eqref{hpmd_update_decomposable}. 
\begin{itemize}
\item[I.]  If entropic index $q > 1$.
 Then \eqref{pnorm_linear_convergence}, \eqref{pnorm_finite_time_convergence}, and \eqref{pnorm_policy_convergence} hold.

\item[II.] If entropic index $q \in  (0,1)$,
and $\min_{s \in \cA, a \in \cA} \pi_0(a|s)  > 0$, 
 then the following holds.
 \begin{itemize}
\item[1.] Linear convergence of the optimality gap: 
\begin{align}\label{tsallis_linear_convergence}
f(\pi_{k}) - f(\pi^*) 
\leq 
\gamma^k \rbr{
 f(\pi_{0}) - f(\pi^*)  + 8 \abs{\cA} / (1-\gamma)
}, ~\forall k \geq 0.
\end{align}

\item[2.]
Local  accelerated convergence:
for any $\epsilon > 0$, we have 
\begin{align}
\mathrm{dist}_{\ell_{1}} (\pi_{k+1} ,\Pi^*)  & \leq 2 \epsilon  \abs{\cA} \gamma^{2k} ,~ 
f(\pi_{k+1}) - f(\pi^*) \leq
\frac{2 \epsilon  \abs{\cA} C}{(1-\gamma)^2} \gamma^{2k},  
~
\forall k \geq \underline{K}_2(\epsilon), 
 \label{tsallis_sup_convergence_generalized}
\end{align}
where 
\begin{align}
\underline{K}_2(\epsilon) 
 = &   \log_\gamma \rbr{
\frac{\Delta^*(\cM) (1-\gamma^3) ( 1-\gamma) \gamma}{8 ( \max_{i \in \cA} \abs{ \nabla  \hat{v}(\pi_i^0)} + C + p) }
}
+ 
\frac{1-p}{2p} \log_\gamma (\epsilon) 
+ 
\frac{1}{2p} \log_\gamma \rbr{ \frac{\Delta^*(\cM)}{4p}} \nonumber 
\\
&  
+ 
3 { \log_\gamma \rbr{ \frac{ \Delta^*(\cM) (1-\gamma) }{ 2\varrho\rbr{ 8 \abs{\cA} + C} }} }.
\label{tsallis_uk2}
 \end{align}
\item[3.] Last-iterate policy convergence: 
for any $\epsilon \in (0,1)$, it holds that
\begin{align}\label{tsallis_policy_convergence}
\norm{\pi_{k+1} - \pi^*_U}_\infty \leq \epsilon,
~ 
k \geq K_2 \rbr{ \frac{p (1-p) \epsilon}{2}}  + \underline{K}_2 \rbr{ \frac{\epsilon}{2 \abs{\cA}}},
\end{align}
where $K_2(\cdot)$ is defined as 
\begin{align*}
K_2(\epsilon) \coloneqq
2 \underline{K}_2 \rbr{\frac{(1-\gamma)^3 \epsilon}{4 \abs{\cA} C}}
+ \log_\gamma \rbr{
\frac{ (1-\gamma^2) (1-\gamma) \gamma \epsilon}{8  \varrho (8 \abs{\cA} + C) }
}
+ 
\log_\gamma \rbr{
\frac{ \epsilon}{ 8  \max_{i \in \cA} \abs{\nabla \hat{v}(\pi_i^0)} }
}.
\end{align*}
\end{itemize}
\end{itemize}

\end{corollary}

\begin{proof}
The case for $q > 1$ is readily implied by Corollary \ref{corr_p_norm}. 

We proceed to the proof for $q \in (0,1)$.
By definition, $ - \abs{\cA}^{1-p} \leq w(\pi(\cdot|s)) \leq 0 $,
hence one can take $\Phi = 2 \abs{\cA}$.
 Then \eqref{tsallis_linear_convergence} comes immediately after applying Corollary \ref{cor_generalized_hpmd_linear_convergence}.

Next, we determine $\underline{K}_2(\cdot)$, defined in Lemma \ref{lemma_hpmd_generalized_local_convergence}. 
Direct calculation shows $\nabla \hat{v}^*(x) = (1/p)^{\frac{1}{p-1}} \abs{x}^{\frac{1}{p-1}}$ for $x < 0$. 
Combining this with the definition of $K_1$ in Lemma \ref{hpmd_generalized_converge_to_optimal}, it can be readily verified that 
$
\underline{K}_2(\epsilon) 
$ defined in \eqref{tsallis_uk2}
satisfies \eqref{def_uK2_from_epsilon}.
Consequently, we obtain \eqref{tsallis_sup_convergence_generalized} after applying Lemma \ref{lemma_hpmd_generalized_local_convergence}. 

We then proceed to establish the policy convergence.
 Since $\partial v(0) = \emptyset$,  from Theorem \ref{thrm_limit_policy_decomposable} it holds $\pi_i^k > 0$ for any $k \geq 0$ and $i \in \cA$, and $\hat{v}$ is differentiable at $\pi_i^k$.
Fix $\epsilon \in (0,1)$ and $\epsilon ' > 0$,  suppose for some $K > 0$, 
\begin{align}\label{tsallis_close_dual_and_close_to_opt_policy}
\tsum_{i \in \cA^*(s)} \pi_i^{k+1} \geq 1-\epsilon, 
~
\abr{ \nabla v(\pi_i^{k+1})  - \nabla v(\pi_j^{k+1})}
\leq \epsilon', ~ \forall i , j \in \cA^*(s), 
\end{align} 
for  any $k \geq K$.
Note that if $\abs{\cA^*(s)} = 1$, then \eqref{tsallis_policy_convergence} follows immediately. 
Thus it suffices consider the case of $\abs{\cA^*(s)} > 1$.
Let $i^* \in \Argmax_{i \in \cA^*(s)}  \pi_i^{k+1}$, then the second inequality in \eqref{tsallis_close_dual_and_close_to_opt_policy}  then implies 
\begin{align*}
\epsilon ' \geq  \nabla v(\pi_{i^*} ^{k+1})  - \nabla v(\pi_i^{k+1})
\geq 
p \rbr{ - (\pi_{i^*} ^{k+1})^{p-1} +  (\pi_{i} ^{k+1})^{p-1} }
\geq p (1-p) \rbr{ \pi_{i^*} ^{k+1} - \pi_{i} ^{k+1}}
\end{align*}
for any $i \in \cA^*(s)$.
Hence $\abs{ \pi_{i} ^{k+1} - \pi_{j} ^{k+1}} \leq \epsilon'$ for any $i,j \in \cA^*(s)$. 
Combing the above inequality and \eqref{tsallis_close_dual_and_close_to_opt_policy}, 
we can obtain, following similar lines as in the proof of Lemma \ref{lemma_prob_ub_and_lb} for showing \eqref{lemma_eq_prob_ub_and_lb}, 
that 
$
\frac{1-\epsilon}{\abs{\cA^*(s)}} - \frac{\epsilon'}{p (1-p)} 
\leq \pi_i^{k+1} \leq \frac{1}{\abs{\cA^*(s)}} + \frac{\epsilon' }{p(1-p)}.
$
Thus \eqref{tsallis_policy_convergence} follows by choosing $\epsilon' \leq p (1-p) \epsilon / 2$. 

It remains to choose $K$ so that both conditions \eqref{tsallis_close_dual_and_close_to_opt_policy} are satisfied with the above choice of $\epsilon'$.
In view of the first inequality in \eqref{tsallis_sup_convergence_generalized}, the first condition in \eqref{tsallis_close_dual_and_close_to_opt_policy} holds when 
$k \geq \underline{K}_2(\frac{\epsilon}{2 \abs{\cA}})$. 
Meanwhile, in view of Lemma \ref{lemma_hpmd_dual_asymp_equal_opt_action}, 
the second condition in \eqref{tsallis_sup_convergence_generalized} holds when 
$ k \geq K_2(\epsilon')$.
In conclusion, \eqref{tsallis_close_dual_and_close_to_opt_policy} holds when $k \geq K_2(\epsilon') + \underline{K}_2(\frac{\epsilon}{2 \abs{\cA}})$.
The proof for $q \in (0,1)$ is then completed.
\qed \end{proof}

To conclude our discussions in this section, 
it is worth mentioning that  the finite time exact convergence of HPMD with the distance-generating function either being the $p$-th power of $\ell_p$-norm (Corollary \ref{corr_p_norm}), or the negative Tsallis entropy with entropy index $q > 1$ (Corollary \ref{cor_tsallis_entropy_parameter}),  appears to be new in the literature of PG  methods.
To the best of our knowledge, we are not aware of any other result on the finite time convergence of PG methods, regardless the choice of the distance-generating function.
Interestingly, both corollaries also show that HPMD would continue its convergence to the limiting policy $\pi^*_U$, even if the current policy is already optimal after a finite number of iterations.

\section{Stochastic Homotopic Policy Mirror Descent}\label{sec:stochastic}

In this section, we introduce the stochastic homotopic policy mirror descent (SHPMD) method, and study its convergence properties for both the optimality gap and the policy, and accordingly establish its sample complexity. 
In addition, by establishing and exploiting the local superlinear convergence of SHPMD, we can further  improve the sample complexity  when searching for a close-to-optimal policy.

The SHPMD method (Algorithm \ref{alg:sfpmd}) minimizes the objective \eqref{eq:mdp_single_obj} by performing the following update at every iteration $k \geq 0$, 
\begin{align}
\label{shpmd:update}
\pi_{k+1}(\cdot | s) = \argmin_{p(\cdot |s) \in \Delta_{\cA}} \eta_k \sbr{\inner{Q^{\pi_k, \xi_k}(s, \cdot)}{p(\cdot|s)} + \tau_k D^{p}_{\pi_0} (s) } + D^{p}_{\pi_k}(s), ~~ \forall s \in \cS.
\end{align}
For the simplicity of computing divergence-dependent constant, we will restrict our attention to $D^{\pi}_{\pi'}$ being the KL divergence, similar to our discussions in Section \ref{sec_deterministic} and \ref{sec_local_convergence_and_last_iterate}.

\begin{algorithm}[b!]
    \caption{The stochastic homotopic  policy mirror descent (SHPMD) method}
    \label{alg:sfpmd}
    \begin{algorithmic}
    \STATE{\textbf{Input:} Initial policy $\pi_0$, nonnegative parameters $\cbr{\tau_k} $, and stepsizes $\{\eta_k\} $.}
    \FOR{$k=0, 1, \ldots$}
	\STATE{Update policy:
			\vspace{-0.1in}
	\begin{align*}
	\pi_{k+1} (\cdot| s) =
	\argmin_{p(\cdot|s) \in \Delta_{\cA}} \eta_k\sbr{ \inner{Q^{\pi_k, \xi_k}(s, \cdot)}{ p(\cdot| s ) }  + \tau_k D^{p}_{\pi_0} (s) } + D^{p}_{\pi_k}(s),  ~ \forall s \in \cS
	\end{align*}}
	\vspace{-0.15in}
    \ENDFOR
    \end{algorithmic}
\end{algorithm}

Different from the deterministic HPMD method, SHPMD uses a stochastic estimate $Q^{\pi_k, \xi_k}$ of the state-action value function $Q^{\pi_k}$ to perform the policy update.
To facilitate our analysis, we impose following conditions on the stochastic estimate $Q^{\pi_k, \xi_k} \in \RR^{\abs{\cS} \times \abs{\cA}}$:
\begin{align}
\EE_{\xi_k} Q^{\pi_k, \xi_k} = \overline{Q}^{\pi_k}, ~~
\norm{\overline{Q}^{\pi_k} - Q^{\pi_k} }_{\infty}  \leq \varepsilon_k,  ~~
\EE \norm{{Q}^{\pi_k, \xi_k} - Q^{\pi_k} }_{\infty}^2  \leq \sigma_k^2,  \label{noise_conditions}
\end{align}

To construct an estimate $Q^{\pi_k, \xi_k}$ satisfying the above conditions, we consider the simple scenario where we have the access to a generative model of the MDP.  
That is, for any state-action pair $(s, a) \in \cS \times \cA$, we can generate $M_k$ independent trajectories, each with length $T_k$, by following policy $\pi_k$ and starting from $(s,a)$. The $i$-th trajectory, denoted by $\chi_k^i (s,a)$, takes the form of
$
\chi_k^i(s,a) = \{ (s_0^i = s, a_0^i = a), (s_1^i, a_1^i), \ldots, (s_{T_k - 1}^i, a_{T_k - 1}^i ) \}.
$
We then denote the set of random variables for constructing the stochastic estimate as $\xi_k = \{ \chi_k^i(s, a):  s\in \cS, a \in \cA , i \in [M_k]\}$. Accordingly,  the estimator $Q^{\pi_k, \xi_k}$ is defined by 
\begin{align}\label{estimator_ind_traj}
Q^{\pi_k, \xi_k} (s, a ) = 
  \frac{1}{M_k}  \tsum_{i=1}^{M_k}  \tsum_{t=0}^{T_k - 1} \gamma^t c(s_t^i, a_t^i)  , &  ~~ s \in \cS,   a \in \cA.
  \end{align}

\subsection{Linear Convergence and Sample Complexity}

In this subsection, we establish the global linear convergence of SHPMD with certain conditions on   $\cbr{(\sigma_k ,\varepsilon_k)} $, together with proper specification of $\cbr{(\tau_k,\eta_k)} $.
We begin by the following lemma characterizing the update of  SHPMD.


\begin{lemma}\label{lemma_three_point_stochastic}
For any $p\in \Pi$ and any $s\in\cS$,  we have 
\begin{align}\label{eq_three_point_stochastic}
& \eta_k \inner{Q^{\pi_k, \xi_k}(s, \cdot)}{\pi_{k+1}(\cdot|s) - p(\cdot|s) } + \eta_k \tau_k \rbr{D^{\pi_{k+1}}_{\pi_0}(s) - D^{p}_{\pi_0}(s)  }
+ D^{\pi_{k+1}}_{\pi_k}(s) \nonumber \\
\leq & D^{p}_{\pi_k}(s) - (\tau_k \eta_k + 1) D^{p}_{\pi_{k+1}}(s).
\end{align}
\end{lemma}

The proof of Lemma \ref{lemma_three_point_stochastic} follows exactly the same lines as in Lemma \ref{lemma_three_point}.
We then proceed to establish some generic convergence properties of the SHPMD method.

\begin{lemma}\label{stochastic_generic}
Suppose $1 + \eta_k \tau_k = 1/\gamma$, and $\alpha_k > 0$ satisfies 
\begin{align}\label{alpha_choice_stoch}
\alpha_k \geq \alpha_{k+1} \gamma, ~
\frac{\alpha_k}{\gamma \eta_k} \geq \frac{\alpha_{k+1}}{\eta_{k+1}},
\end{align}
then for any iteration $k$, SHPMD satisfies 
\begin{align}
&\EE \sbr{ \alpha_k \rbr{f(\pi_{k+1}) - f(\pi^*)} + \frac{\alpha_k}{\gamma \eta_k} \phi(\pi_{k+1}, \pi^*) } \nonumber \\
 \leq &  
\alpha_0 \gamma \rbr{f(\pi_{0}) - f(\pi^*) } + \frac{\alpha_0}{\eta_0} \phi(\pi_0, \pi^*) 
+  2   \tsum_{t=0}^k \alpha_t \varepsilon_t + 
\frac{2}{1-\gamma}  \tsum_{t=0}^k \alpha_t \eta_t \sigma_t^2
+ \frac{3}{\gamma}   \tsum_{t=0}^k \frac{\alpha_t}{\eta_t} \log \abs{\cA}. \label{stoch_generic_recursion}
\end{align}
\end{lemma}

\begin{proof}
For each $s \in \cS$, by plugging  $p = \pi_k$ in \eqref{eq_three_point_stochastic}, then 
\begin{align}\label{three_point_negative_stochastic}
& \eta_k \inner{Q^{\pi_k, \xi_k}(s, \cdot)}{\pi_{k+1}(\cdot|s) - \pi_k(\cdot|s) } 
+ \eta_k \tau_k \rbr{D^{\pi_{k+1}}_{\pi_0}(s) - D^{\pi_k}_{\pi_0}(s)  }  + D^{\pi_{k+1}}_{\pi_k} (s) \leq  -(\tau_k \eta_k + 1) D^{\pi_k}_{\pi_{k+1}} (s) .
\end{align}
On the other hand, choosing $p = \pi^*$ in \eqref{eq_three_point_stochastic} yields 
\begin{align}
\eta_k \inner{Q^{\pi_k, \xi_k}(s, \cdot)}{\pi_{k}(\cdot|s) - \pi^*(\cdot|s) } + & \eta_k \inner{Q^{\pi_k, \xi_k}(s, \cdot)}{\pi_{k+1}(\cdot|s) - \pi_{k}(\cdot|s) } + \eta_k \tau_k \rbr{D^{\pi_{k+1}}_{\pi_0}(s) - D^{\pi^*}_{\pi_0}(s)  }  \nonumber \\
& \leq D^{\pi^*}_{\pi_k}(s) - (\tau_k \eta_k + 1) D^{\pi^*}_{\pi_{k+1}}(s)
- D^{\pi_{k+1}}_{\pi_k}(s) 
. \label{three_point_decomposition_stochastic}
\end{align}

Let us define $\delta_k = Q^{\pi_k, \xi_k} - Q^{\pi_k} \in \RR^{\abs{\cS} \times \abs{\cA}}$, 
then from \eqref{diff_old_pi}, we obtain 
\begin{align}\label{diff_old_opt_stochastic}
\EE_{s \sim \nu^*}  \inner{Q^{\pi_k, \xi_k}(s, \cdot)}{\pi_{k}(\cdot|s) - \pi^*(\cdot|s) } 
= (1-\gamma) \rbr{ f(\pi_k) - f(\pi^*)}
+ \EE_{s \sim \nu^*} \inner{\delta_k(s,\cdot)}{\pi_{k}(\cdot|s) -  \pi^*(\cdot|s)  }.
\end{align}
In addition, to handle the second term in \eqref{three_point_decomposition_stochastic}, we observe the following:  
\begin{align*}
& (1-\gamma) \rbr{V^{\pi_{k+1}} (s) - V^{\pi_k}(s)} +  \EE_{s' \sim d_{s}^{\pi_{k+1}}} \sbr{\tau_k \rbr{D^{\pi_{k+1}}_{\pi_0}(s')  - D^{\pi_{k}}_{\pi_0} (s')  } }\\
\overset{(a)}{=} & \EE_{s' \sim d_{s}^{\pi_{k+1}}} \big[
\inner{Q^{\pi_k, \xi_k}(s', \cdot)}{ \pi_{k+1}(\cdot|s') - \pi_k(\cdot|s')} + \tau_k \rbr{D^{\pi_{k+1}}_{\pi_0}(s')  - D^{\pi_{k}}_{\pi_0} (s')  }
\\
& ~~~~~~ - \inner{\delta_k(s', \cdot)}{ \pi_{k+1}(\cdot|s') - \pi_k(\cdot|s')} 
\big]  \\
\leq & \EE_{s' \sim d_{s}^{\pi_{k+1}}} \big[
\inner{Q^{\pi_k, \xi_k}(s', \cdot)}{ \pi_{k+1}(\cdot|s') - \pi_k(\cdot|s')} + \tau_k \rbr{D^{\pi_{k+1}}_{\pi_0}(s')  - D^{\pi_{k}}_{\pi_0} (s')  }
\\
& ~~~~~~ + \norm{\delta_k(s', \cdot)}_{\infty} \norm{ \pi_{k+1}(\cdot|s') - \pi_k(\cdot|s')}_1
\big]
 \\
\leq & \EE_{s' \sim d_{s}^{\pi_{k+1}}} \big[
\inner{Q^{\pi_k, \xi_k}(s', \cdot)}{ \pi_{k+1}(\cdot|s') - \pi_k(\cdot|s')} + \tau_k \rbr{D^{\pi_{k+1}}_{\pi_0}(s')  - D^{\pi_{k}}_{\pi_0} (s')  }
\\
& ~~~~~~+ \frac{\eta_k}{2} \norm{\delta_k(s, \cdot)}_{\infty}^2 + 
\frac{1}{2\eta_k} \norm{\pi_{k+1}(\cdot|s') - \pi_k(\cdot|s')}_1^2
\big] \\
\overset{(b)}{\leq} & \EE_{s' \sim d_{s}^{\pi_{k+1}}} \big[
\inner{Q^{\pi_k, \xi_k}(s', \cdot)}{ \pi_{k+1}(\cdot|s') - \pi_k(\cdot|s')} + \tau_k \rbr{D^{\pi_{k+1}}_{\pi_0}(s')  - D^{\pi_{k}}_{\pi_0} (s')  }
\\
& ~~~~~~ +  \frac{\eta_k}{2} \norm{\delta_k(s', \cdot)}_{\infty}^2 + 
\frac{1}{\eta_k} D^{\pi_{k+1}}_{\pi_{k}}(s') 
\big] \\
\overset{(c)}{\leq} & (1-\gamma) \big[\inner{Q^{\pi_k, \xi_k}(s, \cdot)}{ \pi_{k+1}(\cdot|s) - \pi_k(\cdot|s)} + \tau_k \rbr{D^{\pi_{k+1}}_{\pi_0}(s)  - D^{\pi_{k}}_{\pi_0} (s)  }  + \frac{1}{\eta_k} D^{\pi_{k+1}}_{\pi_{k}}(s) \big]  
\\
& ~~~~~~ + \frac{\eta_k}{2} \norm{\delta_k}_{\infty}^2,
\end{align*}
where $(a)$ uses  Lemma \ref{lemma_perf_diff}, 
$(b)$ uses Pinsker's inequality, and   $(c)$ uses \eqref{three_point_negative_stochastic} and the fact that 
$d_s^{\pi_{k+1}}(s) \geq 1- \gamma$.
The above relation in turn implies
\begin{align}
\inner{Q^{\pi_k, \xi_k}(s, \cdot)}{ \pi_{k+1}(\cdot|s) - \pi_k(\cdot|s)} & \geq 
\rbr{V^{\pi_{k+1}} (s) - V^{\pi_k}(s)}
- \frac{2 \tau_k}{1-\gamma} \max_{s \in \cS}  \abs{D^{\pi_{k+1}}_{\pi_0}(s)  - D^{\pi_{k}}_{\pi_0} (s)  }  \nonumber \\
& ~~~~  - \frac{\eta_k}{2(1-\gamma) } \norm{\delta_k}_{\infty}^2
-\frac{1}{\eta_k} D^{\pi_{k+1}}_{\pi_{k}}(s).
 \label{stochastic_diff_new_old_clean}
\end{align}
Thus, by combining \eqref{three_point_decomposition_stochastic} and \eqref{stochastic_diff_new_old_clean}, we obtain 
\begin{align*}
& \eta_k   \inner{Q^{\pi_k, \xi_k}(s, \cdot)}{\pi_{k}(\cdot|s) - \pi^*(\cdot|s) } 
+ \eta_k \big[
\rbr{V^{\pi_{k+1}} (s) - V^{\pi_k}(s)}
- \frac{2 \tau_k}{1-\gamma} \max_{s \in \cS}  \abs{D^{\pi_{k+1}}_{\pi_0}(s)  - D^{\pi_{k}}_{\pi_0} (s)  }  \nonumber \\
&  ~~~~~~~~ - \frac{\eta_k}{2(1-\gamma) } \norm{\delta_k}_{\infty}^2
\big] \\
\leq &  D^{\pi^*}_{\pi_k}(s) - (\tau_k \eta_k + 1) D^{\pi^*}_{\pi_{k+1}}(s) + \eta_k \tau_k \max_{s \in \cS} \abs{D^{\pi_{k+1}}_{\pi_0}(s) - D^{\pi^*}_{\pi_0}(s)  } .
\end{align*}
Taking expectation w.r.t. $s \sim \nu^*$ on both sides of the previous relation,  and combining with \eqref{diff_old_opt_stochastic}, then 
\begin{align*}
& (1-\gamma) \sbr{f(\pi_k) - f(\pi^*)} + f(\pi_{k+1}) - f(\pi_k) \\
\leq & 
\frac{1}{\eta_k} \phi(\pi_k, \pi^*) - \rbr{\frac{1}{\eta_k} + \tau_k}  \phi(\pi_{k+1}, \pi^*)
+ \frac{2 \tau_k}{1-\gamma} \max_{s\in \cS} \abs{ D^{\pi_{k+1}}_{\pi_0}(s')  - D^{\pi_{k}}_{\pi_0} (s') }  \\
& ~~~~~~ -   \EE_{s \sim \nu^*}  \inner{\delta_k(s,\cdot)}{\pi_{k}(\cdot|s) -  \pi^*(\cdot|s)  }
+ \frac{\eta_k}{2(1-\gamma) } \norm{\delta_k}_{\infty}^2 + \tau_k \max_{s \in \cS} \abs{D^{\pi_{k+1}}_{\pi_0}(s) - D^{\pi^*}_{\pi_0}(s)  }  .
\end{align*}
Simple rearrangement of the previous relation gives 
\begin{align*}
&  f(\pi_{k+1}) - f(\pi^*)   +   \rbr{\frac{1}{\eta_k} + \tau_k}  \phi(\pi_{k+1}, \pi^*)  \\
\leq & 
\gamma \rbr{f(\pi_k) - f(\pi^*)} + 
\frac{1}{\eta_k} \phi(\pi_k, \pi^*) 
+ \frac{2 \tau_k}{1-\gamma} \max_{s\in \cS} \abs{ D^{\pi_{k+1}}_{\pi_0}(s')  - D^{\pi_{k}}_{\pi_0} (s') }  \\
& ~~~~~~ -  \EE_{s \sim \nu^*}  \inner{\delta_k(s,\cdot)}{\pi_{k}(\cdot|s) -  \pi^*(\cdot|s)  }
+ \frac{\eta_k}{2(1-\gamma) } \norm{\delta_k}_{\infty}^2 + \tau_k \max_{s \in \cS} \abs{D^{\pi_{k+1}}_{\pi_0}(s) - D^{\pi^*}_{\pi_0}(s)  } 
.
\end{align*}
Letting $1+ \eta_k \tau_k = 1/\gamma$ in the above relation, then it holds that  
\begin{align}
&  f(\pi_{k+1}) - f(\pi^*)   +  \frac{1}{\gamma\eta_k}  \phi(\pi_{k+1}, \pi^*)   \nonumber \\
\leq & 
\gamma\rbr{f(\pi_k) - f(\pi^*)} + 
\frac{1}{\eta_k} \phi(\pi_k, \pi^*) 
+ \frac{3}{\gamma \eta_k} \log \abs{\cA}  -  \EE_{s \sim \nu^*}  \inner{\delta_k(s,\cdot)}{\pi_{k}(\cdot|s) -  \pi^*(\cdot|s)  } 
+ \frac{ \eta_k}{2(1-\gamma)} \norm{\delta_k}_{\infty}^2, \nonumber
\end{align}
where we use the fact that $ 0 \leq D^{\pi}_{\pi_0}(s) \leq \log \abs{\cA}$ given $\pi_0$ being the uniform policy.
Now multiplying both sides of previous relation with positive $\alpha_k > 0$, where 
$
\alpha_k \geq \alpha_{k+1} \gamma, ~
\frac{\alpha_k}{\gamma \eta_k} \geq \frac{\alpha_{k+1}}{\eta_{k+1}},
$
and summing up from $t=0$ to $k$,  and taking expectation w.r.t $\cbr{\xi_t}$, we obtain 
\begin{align*}
& \EE \sbr{ \alpha_k \rbr{f(\pi_{k+1}) - f(\pi^*)} + \frac{\alpha_k}{\gamma \eta_k} \phi(\pi_{k+1}, \pi^*) } \\
 \leq &  
\alpha_0 \gamma \rbr{f(\pi_{0}) - f(\pi^*) } + \frac{\alpha_0}{\eta_0} \phi(\pi_0, \pi^*) 
+ 2 \tsum_{t=0}^k \alpha_t \varepsilon_t + 
\frac{2}{1-\gamma}  \tsum_{t=0}^k \alpha_t \eta_t \sigma_t^2
+ \frac{3}{\gamma}   \tsum_{t=0}^k \frac{\alpha_t}{\eta_t} \log \abs{\cA},
\end{align*}
where we use condition \eqref{noise_conditions}, together with $\inner{ \EE_{\xi_k} \delta_k(s,\cdot)}{\pi^*(\cdot|s) -  \pi_k(\cdot|s)  } \leq 2 \norm{\EE_{\xi_k} \delta_k(s,\cdot)}_{\infty} \leq 2 \varepsilon_k$. 
\qed \end{proof}

We now specify the concrete choice of $\cbr{(\eta_k, \tau_k)} $ and the conditions on $\cbr{(\varepsilon_k, \sigma_k)} $ that yield the global linear convergence of the SHPMD method.

\begin{theorem}\label{thrm_convergence_sfpmd}
Take $1 + \eta_k \tau_k = 1/\gamma$ and  $\eta_k = \gamma^{-(k+1)/2}\sqrt{\log \abs{\cA} (1-\gamma)}$ in the SHPMD method.
Suppose 
\begin{align}\label{linear_noise_condition}
\sigma_k = \gamma^{(k+1)/2}, ~ \varepsilon_k =  \gamma^{3(k+1)/4},
\end{align}
then SHPMD produces policy $\pi_k$ satisfying 
\begin{align}\label{eq_convergence_sfpmd}
\EE \sbr{ f(\pi_k) - f(\pi^*) }\leq  \gamma^{k/2 } \frac{32 \sqrt{\log \abs{\cA}} + C }{ (1- \gamma)^{3/2} \gamma}
\coloneqq \cG(k)
,
~\forall k \geq 1.
\end{align}
\end{theorem}

\begin{proof}
Consider $\alpha_t =\gamma^{-(t+1)}$, then given the choice of $\cbr{(\alpha_t, \eta_t)}$, we obtain
\begin{align*}
\frac{\alpha_t}{\alpha_{t+1}} = \gamma, ~
\frac{\alpha_t}{ \eta_t} = \gamma^{-(t+1)/2 } /\sqrt{\log \abs{\cA} (1-\gamma)}, ~
\end{align*}
and thus \eqref{alpha_choice_stoch} in Lemma \ref{stochastic_generic} is satisfied. In addition, direct calculation yields 
\begin{align*}
 \tsum_{t=0}^k \alpha_t \eta_t \sigma_t^2 \leq \frac{2 \sqrt{\log \abs{\cA}}}{(1-\gamma)^{1/2}} \gamma^{-(k+1)/2},~
  \tsum_{t=0}^k \frac{\alpha_t}{\eta_t}  \log \abs{\cA} \leq  \frac{2 \sqrt{\log \abs{\cA}}}{(1-\gamma)^{3/2}} \gamma^{-(k+1)/2},~
  \tsum_{t=0}^k \alpha_t \varepsilon_t \leq  \frac{4}{1-\gamma} \gamma^{-(k+1)/4}.
\end{align*}
Combining the above relation with \eqref{stoch_generic_recursion} in Lemma \ref{stochastic_generic}, we obtain 
\begin{align*}
\gamma^{-(k+1)} \rbr{f(\pi_{k+1}) - f(\pi^*)} 
\leq 
f(\pi_0) - f(\pi^*) 
+ \frac{\gamma^{-1/2}}{(1-\gamma)^{1/2}} \sqrt{\log \abs{\cA}}
+ \frac{8 }{1-\gamma} \gamma^{-(k+1)/4}
+ \frac{10 \sqrt{\log \abs{\cA}}}{(1-\gamma)^{3/2} \gamma}
\gamma^{-(k+1)/2}
\end{align*}
which, after simple rearrangement,  translates into 
\begin{align*}
\EE \sbr{ f(\pi_k) - f(\pi^*) } \leq  \gamma^{k/2} 
\frac{32 \sqrt{\log \abs{\cA}}  + C}{(1-\gamma)^{3/2} \gamma}, ~\forall k \geq 1,
\end{align*}
for $\abs{\cA} \geq 2$. The proof is completed by noting that \eqref{eq_convergence_sfpmd} holds for $\abs{\cA} =1$ trivially. 
\qed \end{proof}

  We then show that with proper choices of $\cbr{(M_k, T_k)} $, the constructed estimate $Q^{\pi_k, \xi_k}$ defined in \eqref{estimator_ind_traj} satisfies condition \eqref{linear_noise_condition},  and consequently 
 SHPMD converges linearly as described in Theorem \ref{thrm_convergence_sfpmd}.
 Accordingly, the method attains an $\cO (\abs{\cS} \abs{\cA} /\epsilon^2)$ sample complexity when finding an $\epsilon$-optimal policy.

\begin{theorem}\label{thrm_sfpmd_ind_sample}
Take $1 + \eta_k \tau_k = 1/\gamma$ and  $\eta_k = \gamma^{-(k+1)/2}\sqrt{\log \abs{\cA}(1-\gamma)}$ in the SHPMD method.
Suppose at each iteration of SHPMD, the number of independent trajectories $M_k$  and the trajectory length $T_k$ satisfy
\begin{align}\label{trajectory_length_number}
T_k \geq \frac{3(k+1)}{4} + \log_{\gamma} \rbr{\frac{1-\gamma}{2C}}, 
M_k \geq \frac{4C^2 \kappa}{(1-\gamma)^2} \gamma^{-(k+1)} \rbr{ \log (\abs{\cS} \abs{\cA}) + 1},
\end{align}
where $\kappa > 0$ is an absolute constant.
Then for any $\epsilon> 0$,   SHPMD finds a policy $\pi_k$ with $\EE \sbr{f(\pi_k) - f(\pi^*)} \leq \epsilon$ in 
$
k = 2 \log_\gamma \big(\frac{\epsilon (1-\gamma)^{3/2} \gamma}{32 \sqrt{\log \abs{\cA}} + C}\big)
$ iterations.
The total number of samples required by SHPMD can be bounded by 
\begin{align}
\label{eq_thrm_sfpmd_ind_sample}
\tilde{ \cO} \rbr{ \frac{ \abs{\cS} \abs{\cA} \log^2 (\abs{\cS} \abs{\cA}) }{(1-\gamma)^7 \epsilon^2}  }.
\end{align}
\end{theorem}

\begin{proof}
Combining the choice of $M_k$ and $T_k$ in \eqref{trajectory_length_number}
with  Proposition 7 in \cite{lan2022policy}, 
one can readily verify that condition \eqref{linear_noise_condition} is satisfied.
From \eqref{eq_convergence_sfpmd}, to find an $\epsilon$-optimal policy, SHPMD needs 
$
k = 2 \log_\gamma \big(\frac{\epsilon (1-\gamma)^{3/2} \gamma}{32 \sqrt{\log \abs{\cA}} + C}\big)
$ iterations. 
Thus the total number of samples can be bounded by 
\begin{align}
\abs{\cS} \abs{\cA}  \tsum_{t=0}^k M_t T_t
& \leq \frac{ 20 C^2 \abs{\cS} \abs{\cA} \log (\abs{\cS} \abs{\cA}) }{(1-\gamma)^2}  \tsum_{t=0}^k  \sbr{\frac{3(t+1)}{4} + \log_{\gamma} \rbr{\frac{1-\gamma}{2C}}}   \gamma^{-(t+1)}  \nonumber \\
& = 
\cO \rbr{
 \frac{  C^2 \abs{\cS} \abs{\cA} \log (\abs{\cS} \abs{\cA}) }{(1-\gamma)^3}
\rbr{ (k+1)  +  \log_{\gamma} \rbr{\frac{1-\gamma}{2C}}}\gamma^{-(k+1)}
} \label{eq_shpmd_sample_raw}
\\
& 
= \cO
\rbr{
\frac{\abs{\cS} \abs{\cA} \log \rbr{\abs{\cS} \abs{\cA}} C^2 (\log \abs{\cA} + C^2)}{(1-\gamma)^7 \epsilon^2}
\log \rbr{
\frac{32 \sqrt{ \log \abs{\cA}} +C}{(1-\gamma)^{3/2} \gamma \epsilon}
}
}. \nonumber
\end{align}
The proof is then completed. 
\qed \end{proof}

As suggested by Theorem \ref{thrm_sfpmd_ind_sample}, SHPMD requires at most $\tilde{\cO}(\frac{\abs{\cS} \abs{\cA}}{(1-\gamma)^7 \epsilon^2})$ samples to find an $\epsilon$-optimal policy, matching the sample complexity of the best existing PG methods in terms of its dependence on the target precision \cite{lan2022policy}.
In the next subsection, we will proceed to establish the local superlinear convergence of SHPMD, from which we can obtain an improved sample complexity for small enough $\epsilon$.

\subsection{Local Superlinear Convergence}

Similar to deterministic HPMD,  the policy $\cbr{\pi_k} $ in SHPMD also exhibits local superlinear convergence to the set of optimal policies.
It is worth pointing out that the superlinear convergence takes effect despite the fact that the noisy first order information $\{Q^{\pi_k, \xi_k} \} $ has linearly decaying noise, as prescribed by \eqref{linear_noise_condition}.
\begin{theorem}[Local Superlinear Convergence]\label{thrm:convergence_to_optimal_sfpmd}
Take $1 + \eta_k \tau_k = 1/\gamma$ and  $\eta_k = \gamma^{-(k+1)/2}\sqrt{\log \abs{\cA}(1-\gamma)}$ in the SHPMD method.
Additionally, let $\cbr{(\sigma_k, \epsilon_k)}$ defined in \eqref{noise_conditions} satisfies \eqref{linear_noise_condition}. 
Then SHPMD satisfies
\begin{align*}
 \mathrm{dist}_{\ell_{1}} (\pi_k ,\Pi^*)  \leq   2  C_\gamma \abs{\cA} \exp \rbr{ -   { \sqrt{\log \abs{\cA} (1-\gamma)} \Delta^*(\cM) \gamma^{-\frac{k}{2} + \frac{1}{2} } }/{4} },
 \text{
 with probability  $1 - \frac{8 \gamma^{k/6}}{1-\gamma}$, 
 }
\end{align*} 
for $k \geq K_1 \coloneqq  \frac{3\underline{K}_1}{2} + 4 \log_\gamma \rbr{ \frac{\Delta^*(\cM) (1-\gamma) \gamma^{1/2}  }{ 8}} $, 
where $\underline{K}_1 = 4 \log_\gamma \rbr{
\frac{\Delta^*(\cM) (1-\gamma)^{3/2} \gamma}{4 \varrho (32 \sqrt{\log \abs{\cA}} + C) }
}$
,  $C_\gamma  = \exp \rbr{ \frac{2C  \sqrt{\log \abs{\cA}} }{(1-\gamma)^{3/2} }} $.
\end{theorem}

\begin{proof}
Recall that  
$
Q^{\pi_k}(s,a) - Q^*(s,a) = \gamma  \tsum_{s' \in \cS} \cP (s'| s,a) \sbr{V^{\pi_k}(s') - V^*(s')}
$
holds.
Combining  this  observation with Assumption \ref{assump_support}   and Theorem \ref{thrm_sfpmd_ind_sample}, then
$
\EE \sbr{ \max_{s,a} \rbr{Q^{\pi_k}(s,a) - Q^*(s,a)}} \leq   
  \frac{\varrho (32  \sqrt{ \log \abs{\cA} } + C)}{(1-\gamma)^{3/2} \gamma} \gamma^{k/2} 
$ for any $k \geq 1$.
Since $Q^{\pi_k}(s,a) - Q^*(s,a) \geq 0$,  we then obtain from Markov inequality that for any $k \geq 1$, 
\begin{align}\label{convergence_q_high_prob}
 \max_{s,a} \rbr{ Q^{\pi_k}(s,a) - Q^*(s,a) } \leq  
   \frac{\varrho (32  \sqrt{ \log \abs{\cA} } + C)}{(1-\gamma)^{3/2} \gamma}  \gamma^{k/4} , ~ 
\text{
 with probability at least  $ 1- \gamma^{k/4}$
  }.
\end{align}


Going forward, let us adopt the same notations as in the proof of Theorem \ref{thrm:convergence_to_optimal}.
In addition,  we denote $\hat{Q}_i^k$ in short for $Q^{\pi_k, \xi_k}(s,i)$, 
and accordingly define $\delta_i^k = \hat{Q}_i^k - Q_i^k$.
Then similar to \eqref{difference_logit},  
by examining the update of SHPMD \eqref{shpmd:update}, it holds that 
for any pair of actions $i, j \in \cA$,  
\begin{align}
\label{difference_logit_sfpmd_recursion}
z_i^{k+1} - z_j^{k+1} = \gamma^{k+1} (z_i^0 - z_j^0) -  \tsum_{t=0}^{k} \gamma^{k+1 - t} \eta_t ( \hat{Q}_i^t - \hat{Q}_j^t) = -  \tsum_{t=0}^{k} \gamma^{k+1 - t} \eta_t ( \hat{Q}_i^t - \hat{Q}_j^t).
\end{align}

Now consider the case where $i \in \cA^*(s)$, $j \notin \cA^*(s)$.
It is clear that  $Q^*(s, i) < Q^*(s,j)$,
and
from \eqref{convergence_q_high_prob},
we conclude that for any $t \geq \underline{K}_1 \coloneqq 4 \log_\gamma \rbr{
\frac{\Delta^*(\cM) (1-\gamma)^{3/2} \gamma}{4 \varrho (32 \log \sqrt{\abs{\cA}} + C) }
}$, 
\begin{align}\label{diff_q_convergence_sfpmd}
\max_{s \in \cS} \rbr{ \max_{i \in \cA^*(s)}  Q_i^t - \min_{j \notin \cA^*(s)} Q_j^t }
 \leq \Delta^*(\cM) / 2 < 0, 
 ~ 
 \text{
  with probability at least $1 - \gamma^{t/4}$.
 }
\end{align}
From Jensen's inequality,  $\EE \sbr{\max_{s\in \cS, i \in \cA} \abs{\delta_i^t}} \leq (\EE \lVert Q^{\pi_t, \xi_t} - Q^{\pi_t}\rVert_\infty^2)^{1/2} \leq \sigma_t $.
Hence  for any $t \geq 0$,
\begin{align}\label{noise_high_prob}
\max_{s\in \cS, i \in \cA}  \abs{\delta_i^t} \leq \gamma^{t/4}, \text{
with probability at least $ 1- \gamma^{t/4}$.
}
\end{align}
Recall that we choose $\eta_k = \gamma^{-(k+1)/2} \sqrt{\log \abs{\cA} (1-\gamma)}$.
Then
by combining \eqref{difference_logit_sfpmd_recursion}, \eqref{diff_q_convergence_sfpmd} and \eqref{noise_high_prob}, 
for
 any $k \geq 3 \underline{K}_1 / 2$, direct calculation yields that with probability at least $p(k) \coloneqq 1 - {8 \gamma^{k/6}}/\rbr{1-\gamma}$,
\begin{align}
z_i^{k+1} - z_j^{k+1}
& = -  \tsum_{t=0}^{\ceil{2k/3} } \gamma^{k+1 - t} \eta_t ( \hat{Q}_i^t - \hat{Q}_j^t)
-   \tsum_{t=\ceil{2k/3} + 1}^{ k } \gamma^{k+1 - t} \eta_t ( Q_i^t - Q_j^t)
-   \tsum_{t=\ceil{2k/3} + 1}^{ k } \gamma^{k+1 - t} \eta_t (\delta_i^t - \delta_j^t)
\nonumber  \\
& = \bigg[- \underbrace{  \tsum_{t=0}^{\ceil{2k/3}} \gamma^{k-\frac{3}{2} t + \frac{1}{2} } ( \hat{Q}_i^t - \hat{Q}_j^t)}_{(A)}
-  \underbrace{  \tsum_{t=\ceil{2k/3} + 1 }^{ k }  \gamma^{k-\frac{3}{2} t + \frac{1}{2} }  ( Q_i^t - Q_j^t)}_{(B)} \nonumber\\
& ~~~~~~ - \underbrace{  \tsum_{t=\ceil{2k/3}+ 1}^{ k }  \gamma^{k-\frac{3}{2} t + \frac{1}{2} }  (\delta_i^t - \delta_j^t)}_{(C)} \bigg]  \sqrt{\log \abs{\cA} (1-\gamma)}
 \nonumber \\
& \geq - \underbrace{ \frac{2C  \sqrt{\log \abs{\cA}} }{(1-\gamma)^{3/2} \gamma}}_{(A')}
+\underbrace{ \gamma^{-\frac{k}{2} + \frac{1}{2} } \frac{ \sqrt{\log \abs{\cA} (1-\gamma)} \Delta^*(\cM)}{2}}_{(B')}
- \underbrace{ \frac{ 2 \gamma^{-k/4}}{ 1- \gamma }  \sqrt{\log \abs{\cA} (1-\gamma)}}_{(C')},
\label{stochastic_logit_diff_opt_nonopt_1}
\end{align}
holds for any $i \in \cA^*(s)$ and $j \notin \cA^*(s)$. 
Here $(A')$ follows from $(A)$ and $\lVert \hat{Q} \rVert_{\infty} \leq C/(1-\gamma)$,
$(B')$ follows from $(B)$ and that \eqref{diff_q_convergence_sfpmd} holds for every $t \geq \ceil{2k/3} \geq \ceil{\underline{K}_1}$, 
 $(C')$ follows from $(C)$ and that \eqref{noise_high_prob} holds for every $t \geq \ceil{2k/3} \geq \ceil{\underline{K}_1}$,
and the definition of $p(k)$ follows from applying the union bound to \eqref{diff_q_convergence_sfpmd} and \eqref{noise_high_prob} from $t = \ceil{2k/3} + 1$ to $k$.

Thus for $k \geq K_1 \coloneqq \max \cbr{3\underline{K}_1/ 2, 4 \log_\gamma \rbr{\Delta^*(\cM) (1-\gamma) \gamma^{1/2}  / 8}}$, we obtain from \eqref{stochastic_logit_diff_opt_nonopt_1} that
\begin{align*}
z_i^{k+1} - z_j^{k+1}
 \geq - \frac{2C  \sqrt{\log \abs{\cA}} }{(1-\gamma)^{3/2}  \gamma}
+ \gamma^{-\frac{k}{2} + \frac{1}{2} } \frac{ \sqrt{\log \abs{\cA} (1-\gamma) } \Delta^*(\cM) }{4},
\text{
 with probability $1 - \frac{8 \gamma^{k/6}}{1-\gamma}$.
}
\end{align*}
 Hence 
$
\pi_{k+1}(j|s) \leq C_\gamma \exp \rbr{ -   \gamma^{-\frac{k}{2} + \frac{1}{2} }  \sqrt{\log \abs{\cA} (1-\gamma) } \Delta^*(\cM)  / 4 } 
$,
where $C_\gamma  = \exp \rbr{ {2C  \sqrt{\log \abs{\cA}} }/{(1-\gamma)^{3/2} }} $.
From Lemma \ref{lemma:optimal_policy_set}, we conclude 
\begin{align*}
 \mathrm{dist}_{\ell_{1}} (\pi_k ,\Pi^*)  \leq   2  C_\gamma \abs{\cA} \exp \rbr{ -   { \sqrt{\log \abs{\cA} (1-\gamma)} \Delta^*(\cM)} \gamma^{-\frac{k}{2} + \frac{1}{2} } / 4}, ~~ \forall k \geq K_1.
 \end{align*} 
The proof is then completed. 
\qed \end{proof}

Similar to Corollary \ref{corollary:value_convergence}, 
by exploiting Theorem \ref{thrm:convergence_to_optimal_sfpmd}, 
we can establish the local superlinear convergence of the optimality gap. 

\begin{corollary}\label{corollary:value_convergence_sfpmd}
With same settings as in Theorem \ref{thrm:convergence_to_optimal_sfpmd},  then for any $k \geq K_1$,
SHPMD satisfies
\begin{align}
\label{local_value_superlinear_sfpmd}
 f(\pi_k) - f(\pi^*) \leq \frac{ 2 C \abs{\cA}  C_\gamma}{(1-\gamma)^2}  \exp \rbr{ -   { \sqrt{\log \abs{\cA} (1-\gamma)} \Delta^*(\cM) \gamma^{-\frac{k}{2} + \frac{1}{2} } }/{4}  }, 
 \text{
 with probability $1 - \frac{8 \gamma^{k/6}}{1-\gamma}$.
 }
\end{align}
\end{corollary}

\begin{proof}
The proof of Corollary \ref{corollary:value_convergence_sfpmd} follows from Theorem \ref{thrm:convergence_to_optimal_sfpmd}
 and the exact same lines as in Corollary \ref{corollary:value_convergence}. 
\qed \end{proof}

It might be worth noting here that the established local superlinear convergence in Corollary \ref{corollary:value_convergence_sfpmd} holds in high probability.
Converting \eqref{local_value_superlinear_sfpmd} into an expectation bound yields an expected optimality gap that still converges linearly, consistent with Theorem \ref{thrm_convergence_sfpmd}, and prior literature on linearly converging stochastic PG methods (e.g., \cite{lan2022policy}).
Corollary \ref{corollary:value_convergence_sfpmd} can be viewed as a refined probabilistic characterization of the optimality gap, which states that with a  probability stated in \eqref{local_value_superlinear_sfpmd}, the optimality gap diminishes superlinearly.


By exploiting the superlinear convergence established in Corollary \ref{corollary:value_convergence_sfpmd}, an immediate consequence is that we can improve the sample complexity of the SHPMD method obtained in Theorem \ref{thrm_sfpmd_ind_sample}, when searching for a policy with a small optimality gap.

\begin{theorem}\label{thrm_improve_sample}
Let $\cbr{(\tau_k,\eta_k, M_k, T_k)} $ be chosen as in Theorem \ref{thrm_sfpmd_ind_sample}.
Consider any $\epsilon_0 \leq \cG(K_1)$, where $\cG(\cdot)$ is defined as in \eqref{eq_convergence_sfpmd}, and $K_1$ is defined as in Theorem \ref{thrm:convergence_to_optimal_sfpmd}.
Then for any $\epsilon \in (0, \epsilon_0)$, 
with probability at least $ 1 - {3 \epsilon_0^{1/3}}/(1-\gamma)^{1/6}$, 
SHPMD finds an $\epsilon$-optimal policy $\pi_k$ in 
\begin{align}
k = 2 \log_\gamma 
\rbr{
\frac{32 \sqrt{\log \abs{\cA}} + C}{(1-\gamma)^{3/2} \gamma \epsilon_0 }
}
+ 
2
\log_{1/\gamma}
\rbr{
\frac{4}{\sqrt{\log \abs{\cA} (1-\gamma) \gamma} \Delta^*(\cM)} 
\log \rbr{
\frac{2 C \abs{\cA} C_\gamma}{(1-\gamma)^2 \epsilon}
}
}
\label{eq_shpmd_acc_iter}
\end{align}
iterations. 
In addition, the total number of samples of SHPMD for finding $\pi_k$ can be bounded by 
\begin{align}
\tilde{ \cO} \rbr{ \frac{ \abs{\cS} \abs{\cA} \log^2 (\abs{\cS} \abs{\cA}) }{(1-\gamma)^7 [ (1-\gamma)^2 \Delta^*(\cM) \epsilon_0]^2 }  }.
\label{eq_shpmd_acc_sample}
\end{align}
\end{theorem}

\begin{proof}
 Let us denote the first part of the right hand side in \eqref{eq_shpmd_acc_iter} as $\mathcal{I}(\epsilon_0)$,
and the second part as $\mathcal{I}'(\epsilon)$.
Given the choice of $\epsilon_0 \leq \cG(K_1)$,
for any $\epsilon < \epsilon_0$, 
by combining the choice of $k$ specified in \eqref{eq_shpmd_acc_iter} and Theorem \ref{thrm:convergence_to_optimal_sfpmd},
one can verify that 
\begin{align*}
f(\pi_k) - f(\pi^*) \leq \epsilon, 
\text{
with probability $1 - {8 \gamma^{k/6}}/\rbr{1-\gamma} \overset{(a)}{\geq} 1 - {3 \epsilon_0^{1/3}}/(1-\gamma)^{1/6}$,
}
\end{align*}
where $(a)$ follows from decomposition $k = \cI(\epsilon_0) + \cI'(\epsilon)$ and direct calculations.
In addition,
 the total number of samples can be bounded by 
\begin{align*}
\abs{\cS} \abs{\cA}  \tsum_{t=0}^k M_t T_t
& \overset{(b)}{=} 
\cO \rbr{
 \frac{  C^2 \abs{\cS} \abs{\cA} \log (\abs{\cS} \abs{\cA}) }{(1-\gamma)^3}
\rbr{ (k+1)  +  \log_{\gamma} \rbr{\frac{1-\gamma}{2C}}}\gamma^{-(k+1)}
} \\
&  = 
\cO
\rbr{
 \frac{  C^2 \abs{\cS} \abs{\cA} \log (\abs{\cS} \abs{\cA}) }{(1-\gamma)^3}
\rbr{ \cI(\epsilon_0) + \cI'(\epsilon) 
}
 \gamma^{- \cI(\epsilon_0) - \cI'(\epsilon) }
} \\
& \overset{(c)}{=} 
\cO \rbr{
 \frac{  C^2 \abs{\cS} \abs{\cA} \log (\abs{\cS} \abs{\cA}) ( \log \abs{\cA} + C^2) }{(1-\gamma)^8 \Delta^*(\cM)^2 \epsilon_0^2 }
 \log \rbr{
 \frac{32 \sqrt{\log \abs{\cA}} + C}{(1-\gamma)^{3/2} \gamma \epsilon_0}
 }
 \log^2 \rbr{
 \frac{2 C \abs{\cA} C_\gamma}{(1-\gamma)^2 \epsilon}
 }
},
\end{align*}
where $(b)$ follows from \eqref{eq_shpmd_sample_raw}, and $(c)$ follows from the definition of $\cI(\epsilon_0)$, $\cI'(\epsilon)$, and direct calculations.
The proof is then completed by combining the above relation and the definition of $C_\gamma$ in Theorem \ref{thrm:convergence_to_optimal_sfpmd}.
\qed \end{proof}

In view of Theorem \ref{thrm_improve_sample}, SHPMD attains a better than $\tilde{\cO}(\abs{\cS} \abs{\cA}/ \epsilon^2)$ sample complexity for small enough target precision $\epsilon$.
Specifically, for any fixed $\epsilon_0 \leq \cG(K_1)$, and any $\epsilon < \epsilon_0$, 
a sample complexity of $\tilde{\cO}(\abs{\cS} \abs{\cA}/ \epsilon_0^2)$ holds with probability $ 1- \cO(\epsilon_0^{1/3}) $.
This sample complexity strictly improves upon \eqref{eq_thrm_sfpmd_ind_sample} when 
$\epsilon \leq (1-\gamma)^{2} \Delta^*(\cM) \epsilon_0$.
Notably, 
within such a precision region, the sample complexity is independent of the target precision $\epsilon$ up to a logarithmic factor, and the
 success probability can be boosted when $\epsilon_0$ approaches $0$.
To the best of our knowledge, this appears to be the first result among PG methods with a better than $\tilde{\cO}(\abs{\cS} \abs{\cA}/ \epsilon^2)$ sample complexity holding in high probability, when searching for a close-to-optimal policy.


\subsection{Last-iterate Convergence of the Policy}

In this subsection, we proceed to show that with slightly smaller stepsizes $\cbr{\eta_k}$ than the one specified in Theorem \ref{thrm_convergence_sfpmd} and \ref{thrm_sfpmd_ind_sample}, SHPMD exhibits the last-iterate convergence of the policy almost surely. 

\begin{theorem}\label{thrm:prob_lb_sfpmd}
Let the parameters in SHPMD $\cbr{(\eta_k, \tau_k)}$ and the noisy estimate $\{Q^{\pi_k, \xi_k}\}$ satisfy  
\begin{align}\label{param_stochastic_last_iter}
1+\eta_k \tau_k = 1/\gamma, ~ \sigma_k = \gamma^{(k+1)/2}, ~ \eta_k \sigma_k = \gamma^{\beta (k+1)} \sqrt{\log \abs{\cA} (1-\gamma)} , ~ \epsilon_k =   \gamma^{3(k+1)/4 }, 
\end{align}
where $ 0 < \beta < 1/2$.
Then 
$
\lim_{k \to \infty}  \pi_k = \pi^*_U
$
almost surely, where $\pi^*_U$ is defined as in \eqref{eq_def_limit_policy}. 
\end{theorem}

\begin{proof}
Note that \eqref{param_stochastic_last_iter} implies $\eta_k = \gamma^{-(\frac{1}{2} - \beta) (k+1)} \sqrt{\log \abs{\cA} (1-\gamma)}$.
It is clear that condition \eqref{alpha_choice_stoch} 
holds  by choosing $\alpha_k = \gamma^{-(k+1)}$.
Hence one can apply  Lemma \ref{stochastic_generic} and obtain, after direct calculations, that  
\begin{align}\label{shpmd_policy_convergence_opt_gap}
\EE \sbr{ f(\pi_k) - f(\pi^*) } \leq  \gamma^{(\frac{1}{2} - \beta) k} 
\frac{32 \sqrt{\log \abs{\cA}}  + C}{(1-\gamma)^{3/2} \gamma (1-2\beta) }, ~\forall k \geq 1.
\end{align}
Consider any pair of actions $i \in \cA^*(s)$, $j \notin  \cA^*(s)$. 
Applying Markov's inequality to the above relation, then for $t \geq \underline{K}_1 = \frac{4}{1-2\beta} \log_\gamma \rbr{\frac{\Delta^*(\cM) (1-\gamma)^{3/2} \gamma (1-2\beta)  }{4\varrho (32 \log \abs{\cA} + C)} }$, 
\begin{align}\label{diff_q_convergence_sfpmd_policy}
\max_{s \in \cS} \rbr{ \max_{i \in \cA^*(s)}  Q_i^t - \min_{j \notin \cA^*(s)} Q_j^t }
 \leq \Delta^*(\cM) / 2 < 0, 
 ~ 
 \text{
  with probability at least $1 - \gamma^{t/4}$.
 }
\end{align}
In addition, given the specification of $\cbr{\sigma_k}$, it also holds that for any $t \geq 0$, 
\begin{align}\label{noise_high_prob_policy}
\max_{s\in \cS, i \in \cA}  \abs{\delta_i^t} \leq \gamma^{(\frac{1}{2} - \frac{\beta}{2}) t }, \text{
with probability at least $ 1- \gamma^{\beta t / 2}$.
}
\end{align}

For any $z \in (0,1)$, and $k \geq \underline{K}_1 / z$. 
We make the following observations. 
First, given the fact that $\lVert \hat{Q}^{\pi_k, \xi_k} \rVert_\infty \leq C / (1-\gamma)$, and making use of the definition of $\cbr{\eta_k}$, then 
\begin{align}\label{shpmd_early_history_logit_diff}
 \abr{ \tsum_{t=0}^{\ceil{zk} } \gamma^{k+1 - t} \eta_t ( \hat{Q}_i^t - \hat{Q}_j^t)} 
 \leq 
 \frac{2 C \sqrt{\log \abs{\cA}}}{(1-\gamma)^{3/2} \gamma} \gamma^{[1 - (\frac{3}{2} - \beta) z] k  }.
\end{align}
In addition, since $k \geq \underline{K}_1 / z$, one can then apply \eqref{diff_q_convergence_sfpmd_policy} and obtain 
\begin{align}\label{shpmd_recent_history_logit_diff_opt_nonopt}
-   \tsum_{t=\ceil{zk} + 1}^{ k } \gamma^{k+1 - t} \eta_t ( Q_i^t - Q_j^t)
\geq  \frac{ \sqrt{\log \abs{\cA} (1-\gamma)}  \Delta^*(\cM)}{2}  \gamma^{-(\frac{1}{2} - \beta) k + \frac{1}{2} },
\text{
with probability $ 1 - \frac{4 \gamma^{zk / 4}}{1-\gamma} $. 
}
\end{align}
In view of \eqref{noise_high_prob_policy}, it can also be directly verified that
\begin{align}\label{shpmd_recent_history_noise_accumulation}
\abr{
 \tsum_{t=\ceil{zk} + 1}^{ k } \gamma^{k+1 - t} \eta_t (\delta_i^t - \delta_j^t)
 }
 \leq 
 \frac{2  \sqrt{\log \abs{\cA}}  \gamma^{\beta k / 2}}{(1-\gamma)^{1/2} (1-\beta)}, 
 \text{
 with probability $ 1 - \frac{ 2 \gamma^{\beta z k/ 2}}{(1-\gamma) \beta }$.
 }
\end{align}

Hence by combining \eqref{shpmd_early_history_logit_diff}, \eqref{shpmd_recent_history_logit_diff_opt_nonopt}, and \eqref{shpmd_recent_history_noise_accumulation} with a union bound,  and letting $k \geq \overline{K}_1 \coloneqq \underline{K}_1 / z$,
then with probability $p(k) \coloneqq 1 - \frac{4 \gamma^{zk / 4}}{1-\gamma} - \frac{\gamma^{\beta z k}}{(1-\gamma) \beta }$, 
it holds that
for any $i \in \cA^*(s), j \notin \cA^*(s)$, 
\begin{align}
& z_i^{k+1} - z_j^{k+1} \nonumber \\ 
 = &  -  \tsum_{t=0}^{\ceil{zk} } \gamma^{k+1 - t} \eta_t ( \hat{Q}_i^t - \hat{Q}_j^t)
-   \tsum_{t=\ceil{zk} + 1}^{ k } \gamma^{k+1 - t} \eta_t ( Q_i^t - Q_j^t)
-   \tsum_{t=\ceil{zk} + 1}^{ k } \gamma^{k+1 - t} \eta_t (\delta_i^t - \delta_j^t) \nonumber \\ 
  \geq  & 
-  \frac{2 C \sqrt{\log \abs{\cA}}}{(1-\gamma)^{3/2} \gamma} \gamma^{[1 - (\frac{3}{2} - \beta) z] k  }
+  \frac{ \sqrt{\log \abs{\cA} (1-\gamma)}  \Delta^*(\cM)}{2}  \gamma^{-(\frac{1}{2} - \beta) k + \frac{1}{2} } 
-    \frac{2  \sqrt{\log \abs{\cA}}  \gamma^{\beta k / 2}}{(1-\gamma)^{1/2} (1-\beta)}.
\label{shpmd_logit_diff_opt_nonopt}
\end{align}
Choosing $z \in (0, \frac{2}{3 - 2\beta})$ in \eqref{shpmd_logit_diff_opt_nonopt}, then
for $k \geq \overline{K}_1$, 
 with probability $p(k)$, we obtain that for any $ j \notin \cA^*((s)$, 
\begin{align}\label{shpmd_policy_to_opt_set}
\pi_{k+1}(j|s) \leq C_\gamma \exp \rbr{ -   \gamma^{-( \frac{1}{2} -\beta)k  + \frac{1}{2} } \frac{ \sqrt{\log \abs{\cA} (1-\gamma) } \Delta^*(\cM)  }{4} } , 
~ 
C_\gamma = \exp \rbr{\frac{2 (C+2) \sqrt{\log \abs{\cA}}}{(1-\gamma)^{3/2} \gamma} }.
\end{align}
 Whenever \eqref{shpmd_policy_to_opt_set} holds,  applying Assumption \ref{assump_support} and similar arguments as in Corollary \ref{corollary:value_convergence_sfpmd},  then
  \begin{align}\label{shpmd_policy_superlinear_q_convergence}
 \max_{s \in \cS, a \in \cA}   Q^{\pi_k}(s,a) - Q^*(s,a)  \leq \frac{ 2 \varrho C \abs{\cA}  C_\gamma}{(1-\gamma)^2}  \exp \rbr{ -   \frac{ \sqrt{\log \abs{\cA} (1-\gamma)} \Delta^*(\cM) \gamma^{- (\frac{1}{2} - \beta) k + \frac{1}{2} } }{4}  }.
 \end{align}
 
 Now let  $k \geq \max \cbr{ \overline{K}_1, \tilde{K}_1} / z$,
 where $ \tilde{K}_1 = \min \cbr{t \geq 0: \gamma^{-(\frac{1}{2} - \beta) t} \geq \frac{ 6  t \log(1/\gamma)}{\sqrt{\log \abs{\cA} (1-\gamma) \gamma } \Delta^*(\cM)}}$. 
 Consider any pair of actions $i, j \in \cA^*(s)$.
In this case, \eqref{shpmd_early_history_logit_diff} and \eqref{shpmd_recent_history_noise_accumulation} still hold. 
 Instead of \eqref{shpmd_recent_history_logit_diff_opt_nonopt},  we observe
 \begin{align*}
\abr{   \tsum_{t=\ceil{zk} + 1}^{ k } \gamma^{k+1 - t} \eta_t ( Q_i^t - Q_j^t)} 
& \leq 
    \tsum_{t=\ceil{zk} + 1}^{ k } \gamma^{k+1 - t} \eta_t \rbr{ Q_i^t - Q_i^*  + Q_j^t - Q_j^*} \\
    & \overset{(a)}{\leq}
    \tsum_{t=\ceil{zk} + 1}^{ k } \gamma^{k+1 - t}  \eta_t
    \frac{ 4 \varrho C \abs{\cA}  C_\gamma}{(1-\gamma)^2}  \exp \rbr{ -   \frac{ \sqrt{\log \abs{\cA} (1-\gamma)} \Delta^*(\cM) \gamma^{- (\frac{1}{2} - \beta) k + \frac{1}{2} } }{4}  } \\
    & \overset{(b)}{\leq} 
    \frac{4 \varrho C \sqrt{\log \abs{\cA}} \abs{\cA} C_\gamma}{(1-\gamma)^{3/2}} \gamma^k k, ~ 
 \end{align*}
    with probability $p'(k) \coloneqq 1 - \tsum_{t = \ceil{zk} + 1}^k (1-p(t)) = 1 - \frac{16 \gamma^{z^2 k /4}}{z (1-\gamma)^2} - \frac{4 \gamma^{\beta z^2 k / 2 }}{\beta^2 z (1-\gamma)^2}$, 
 where $(a)$ follows from \eqref{shpmd_policy_superlinear_q_convergence} and $k \geq \overline{K}_1 / z$;
  $(b)$ follows from the definition of $\cbr{\eta_t}$, $\tilde{K}_1$, and $k \geq \tilde{K}_1 / z$;
 and the definition of $p'(k)$ follows from applying the union bound to \eqref{shpmd_policy_superlinear_q_convergence} from $t = \ceil{zk} + 1$ to $k$, together with the definition of $p(k)$.
Hence by combining the above relation with  \eqref{shpmd_early_history_logit_diff} and \eqref{shpmd_recent_history_noise_accumulation}, we obtain that 
 with probability $p'(k)$, for any $i, j \in \cA^*(s)$, 
\begin{align}\label{shpmd_logit_dff_opt_opt}
  \abr{ z_i^{k+1} - z_j^{k+1} } 
\leq 
 \frac{2 C \sqrt{\log \abs{\cA}}}{(1-\gamma)^{3/2} \gamma} \gamma^{[1 - (\frac{3}{2} - \beta) z] k  }
 +  \frac{4 \varrho C \sqrt{\log \abs{\cA}} \abs{\cA} C_\gamma}{(1-\gamma)^{3/2}} \gamma^k k
 +   \frac{2  \sqrt{\log \abs{\cA}}  \gamma^{\beta k / 2}}{(1-\gamma)^{1/2} (1-\beta)}.
\end{align}
Let us denote $\cE_k$ as the event where \eqref{shpmd_logit_dff_opt_opt}  and \eqref{shpmd_policy_to_opt_set} hold,
then given the definition of $p'(k)$,  applying Borel–Cantelli lemma shows that with probability $1$,   $\{ \cE_k^\complement \}$ 
occur finitely many times.
Consequently, we can take $k \to \infty$ in \eqref{shpmd_logit_dff_opt_opt} and \eqref{shpmd_policy_to_opt_set}, yielding
\begin{align*}
\lim_{k \to \infty} \pi_{k+1}(j|s) = 0, \forall j \notin \cA^*(s); 
~ \lim_{k \to \infty} \frac{\pi_{k+1}(i|s) }{\pi_{k+1}(j|s) } = 1, ~ \forall i, j \in \cA^*(s), 
\text{
with probability $1$.
}
\end{align*}
Applying Lemma \ref{lemma_prob_ub_and_lb} to the above relation concludes the proof.
\qed \end{proof}

It should be noted that although Theorem \ref{thrm:prob_lb_sfpmd} is stated in an asymptotic fashion, by combining \eqref{shpmd_policy_to_opt_set}, \eqref{shpmd_logit_dff_opt_opt}, the definition of $p'(k)$, and Lemma \ref{lemma_prob_ub_and_lb}, 
one can also establish the non-asymptotic convergence of $\cbr{\pi_k}$ to the limiting policy $\pi^*_U$.

Finally, we show that for the particular choice of $\cbr{(\eta_k, \tau_k)}$ in Theorem \ref{thrm:prob_lb_sfpmd}, the sample complexity of SHPMD with the last-iterate policy convergence can be arbitrarily close to $\cO(\abs{\cS} \abs{\cA} / \epsilon^2)$ by taking $\beta \to 0$.

\begin{theorem}\label{thrm_sfpmd_ind_sample_last_iterate}
Let $\cbr{(\eta_k, \tau_k)}$ in SHPMD be chosen as in \eqref{param_stochastic_last_iter},
where $\beta \in (0,1/2)$.
Suppose at each iteration, the number of independent trajectories $M_k$  and the trajectory length $T_k$ satisfy \eqref{trajectory_length_number}.
Then for any $\epsilon> 0$,   SHPMD finds a policy $\pi_k$ with $\EE \sbr{f(\pi_k) - f(\pi^*)} \leq \epsilon$ in 
$
k = \frac{2}{ 1-2\beta }\log_\gamma \rbr{\frac{\epsilon (1-\gamma)^{3/2} \gamma (1-2\beta) }{32 \sqrt{\log \abs{\cA}} + C}} 
$
 iterations.
In addition, the total number of samples required by SHPMD can be bounded by 
\begin{align*}
\tilde{ \cO} \rbr{
\sbr{
 \frac{ \abs{\cS} \abs{\cA} \log^2 (\abs{\cS} \abs{\cA}) }{(1-\gamma)^7 (1-2\beta)^3 \epsilon^2}
 }^{\frac{1}{1-2\beta} }
   }.
\end{align*}
\end{theorem}

\begin{proof}
Following the same line as in the proof of Theorem \ref{thrm_sfpmd_ind_sample},
with the choice of $\cbr{(T_k, M_k)}$ specified as \eqref{trajectory_length_number}, 
 the condition on $\cbr{(\sigma_k, \epsilon_k)}$ in \eqref{param_stochastic_last_iter} is satisfied. 
 From \eqref{shpmd_policy_convergence_opt_gap}, to find an $\epsilon$-optimal policy, SHPMD needs 
$
k = \frac{2}{ 1-2\beta }\log_\gamma \rbr{\frac{\epsilon (1-\gamma)^{3/2} \gamma (1-2\beta) }{32 \sqrt{\log \abs{\cA}} + C}} 
$ iterations. 
 The total number of samples is given as
\begin{align*}
\abs{\cS} \abs{\cA}  \tsum_{t=0}^k M_t T_t
&  \overset{(a)}{=} 
\cO \rbr{
 \frac{  C^2 \abs{\cS} \abs{\cA} \log (\abs{\cS} \abs{\cA}) }{(1-\gamma)^3}
\rbr{ (k+1)  +  \log_{\gamma} \rbr{\frac{1-\gamma}{2C}}}\gamma^{-(k+1)}
}
\\
& 
= \cO
\rbr{
\sbr{
\frac{\abs{\cS} \abs{\cA} \log \rbr{\abs{\cS} \abs{\cA}} C^2  (\log \abs{\cA} + C^2)  }{(1-\gamma)^7 (1-2\beta)^3 \epsilon^2}
}^{\frac{1}{1-2\beta}}
\log \rbr{
\frac{32 \sqrt{ \log \abs{\cA}} +C}{(1-\gamma)^{3/2} \gamma (1-2 \beta) \epsilon}
}
} ,
\end{align*}
where $(a)$ uses  \eqref{eq_shpmd_sample_raw}.
The proof is then completed.
\qed \end{proof}




\section{Concluding Remarks}\label{sec:discussion}

In this paper, we propose the homotopic policy mirror descent (HPMD) method for solving discounted, infinite horizon MDPs with finite state and action spaces, and study its convergence properties.
By first focusing on the Kullback-Leibler divergence,  we establish the global linear convergence, and the local superlinear convergence for both the optimality gap, and the distance to the set of optimal policies, in an assumption-free manner.
The phase transition from linear to superlinear convergence occurs 
 within $\cO(\log(1/\Delta^*))$ iterations,  where $\Delta^*$ is defined via a gap quantity associated with the optimal state-action value function.
More importantly, 
we establish a non-asymptotic characterization on the last-iterate convergence of the policy, where the limiting policy corresponds to the optimal policy with the maximal entropy for every state.
We then show that the local acceleration and last-iterate policy convergence of HPMD hold for a general class of decomposable Bregman divergences. As a byproduct of the analysis, we also discover the finite-time exact convergence of HPMD with some common Bregman divergences, including the $p$-th power of $\ell_p$-norm and the negative Tsallis entropy.
Finally, for the stochastic HPMD method,
by exploiting the local superlinear convergence, we further establish a sample complexity that is strictly better than $\tilde{\cO} \rbr{\abs{\cS} \abs{\cA} / \epsilon^2}$ when searching for a policy with small optimality gap.

\bibliographystyle{plain}
\bibliography{references}

\end{document}